\documentclass[11pt]{article}

\usepackage[utf8]{inputenc}
\usepackage[T1]{fontenc}
\usepackage{amsmath,amssymb}
\usepackage{array}   
\usepackage{booktabs}
\usepackage{graphicx}
\usepackage{xcolor}
\usepackage[hidelinks]{hyperref}
\usepackage{natbib}
\usepackage{amsthm}
\usepackage{tikz}
\usetikzlibrary{arrows.meta}

\theoremstyle{definition}
\newtheorem{result}{Result}

\newif\ifanon
\anonfalse

\ifanon
  \newcommand{\versiondoi}{\withheld}
  \newcommand{\priorversiondoi}{\withheld}
  \newcommand{\conceptdoi}{\withheld}
  \newcommand{\repourl}{\withheld}
  \newcommand{\dataseturl}{\withheld}
  \newcommand{\repopath}{\withheld}
  \newcommand{\datasetpath}{\withheld}
  \newcommand{\harnessissue}{an open issue}
  \newcommand{\authorname}{\withheld}
  \newcommand{\pdfauthorstring}{Anonymous authors}
\else
  \newcommand{\versiondoi}{%
    \href{https://doi.org/10.5281/zenodo.21939143}{10.5281/zenodo.21939143}}
  \newcommand{\priorversiondoi}{%
    \href{https://doi.org/10.5281/zenodo.21708923}{10.5281/zenodo.21708923}}
  \newcommand{\conceptdoi}{%
    \href{https://doi.org/10.5281/zenodo.21708922}{10.5281/zenodo.21708922}}
  \newcommand{\repourl}{\url{https://github.com/amoghsingh130/flipeval}}
  \newcommand{\dataseturl}{%
    \url{https://huggingface.co/datasets/AmoghSingh123/flipeval-artifacts}}
  \newcommand{\repopath}{\texttt{github.com/\allowbreak amoghsingh130/\allowbreak flipeval}}
  \newcommand{\datasetpath}{\texttt{huggingface.co/\allowbreak datasets/\allowbreak AmoghSingh123/\allowbreak flipeval-artifacts}}
  \newcommand{\harnessissue}{issue \#3831}
  \newcommand{\authorname}{Amogh Singh}
  \newcommand{\pdfauthorstring}{Amogh Singh}
\fi

\title{Certifying Compressed Language Models:\\ An Audit and a Statistical Toolkit}
\ifanon
  \author{Anonymous authors\\ Paper under double-blind review}
\else
  \author{Amogh Singh\\ Georgia Institute of Technology}
\fi
\date{August 2026}

\hypersetup{
  pdftitle={Certifying Compressed Language Models: An Audit and a Statistical Toolkit},
  pdfauthor={\pdfauthorstring},
}

\newcommand{\AuditFrozenCandidates}{17}
\newcommand{\AuditIneligible}{1}
\newcommand{\AuditEligible}{16}
\newcommand{\AuditAssessable}{11}
\newcommand{\AuditNotAssessable}{5}
\newcommand{\AuditNotAssessableInsufficient}{4}
\newcommand{\AuditNotAssessableOutsideFramework}{1}

\newcommand{\AuditBelowThresholdAtMedian}{1}
\newcommand{\AuditAboveThroughout}{10}
\newcommand{\AuditChangesWithinIQR}{1}
\newcommand{\AuditBelowThroughout}{0}

\newcommand{\AuditPerItemTaskMatched}{0}
\newcommand{\AuditPerItemOtherTaskOnly}{3}
\newcommand{\AuditPerItemNone}{13}
\newcommand{\AuditPerItemOtherTaskClaims}{R08, R15, R16}

\newcommand{\AuditXtabQualTotal}{10}
\newcommand{\AuditXtabRetroTotal}{6}

\newcommand{\AuditMarginPP}{2}
\newcommand{\AuditIneligibleClaim}{R10}
\newcommand{\AuditOutsideFrameworkClaim}{R04}

\newcommand{\AuditSensitiveClaim}{R01}
\newcommand{\AuditSensitiveN}{1{,}838}
\newcommand{\AuditSensitiveNReq}{2{,}010}

\newcommand{\AuditSensitiveCellsBelow}{345}
\newcommand{\AuditSensitiveCellsTotal}{792}
\newcommand{\AuditSensitiveCellsPct}{43.6}

\newcommand{\AuditVal}[2]{\csname AuditData#1#2\endcsname}
\expandafter\def\csname AuditDataR01cls\endcsname{changes classification within IQR}
\expandafter\def\csname AuditDataR01dstar\endcsname{0.118915}
\expandafter\def\csname AuditDataR01n\endcsname{1{,}838}
\expandafter\def\csname AuditDataR01nreqMed\endcsname{2{,}010}
\expandafter\def\csname AuditDataR01nreqQOne\endcsname{1{,}364}
\expandafter\def\csname AuditDataR01nreqQThree\endcsname{4{,}328}
\expandafter\def\csname AuditDataR01tier\endcsname{family+bits}
\expandafter\def\csname AuditDataR02blocker\endcsname{no reported or imputable n, no baseline, no numeric delta}
\expandafter\def\csname AuditDataR02kind\endcsname{insufficient reporting}
\expandafter\def\csname AuditDataR02n\endcsname{---}
\expandafter\def\csname AuditDataR02tier\endcsname{family+bits}
\expandafter\def\csname AuditDataR03cls\endcsname{above throughout}
\expandafter\def\csname AuditDataR03dstar\endcsname{---}
\expandafter\def\csname AuditDataR03n\endcsname{18{,}300}
\expandafter\def\csname AuditDataR03nreqMed\endcsname{866}
\expandafter\def\csname AuditDataR03nreqQOne\endcsname{619}
\expandafter\def\csname AuditDataR03nreqQThree\endcsname{1{,}237}
\expandafter\def\csname AuditDataR03tier\endcsname{family+bits}
\expandafter\def\csname AuditDataR04blocker\endcsname{qualifying quote asserts negligible loss on COCO CIDEr, a generation metric with no per-item correct/incorrect state; V1/V2 are flip-model quantities that do not apply to it}
\expandafter\def\csname AuditDataR04kind\endcsname{metric-incompatible}
\expandafter\def\csname AuditDataR04n\endcsname{1{,}319}
\expandafter\def\csname AuditDataR04tier\endcsname{family+bits+benchmark}
\expandafter\def\csname AuditDataR05cls\endcsname{above throughout}
\expandafter\def\csname AuditDataR05dstar\endcsname{0.908491}
\expandafter\def\csname AuditDataR05n\endcsname{14{,}042}
\expandafter\def\csname AuditDataR05nreqMed\endcsname{2{,}255}
\expandafter\def\csname AuditDataR05nreqQOne\endcsname{1{,}525}
\expandafter\def\csname AuditDataR05nreqQThree\endcsname{3{,}865}
\expandafter\def\csname AuditDataR05tier\endcsname{bits+benchmark}
\expandafter\def\csname AuditDataR06cls\endcsname{above throughout}
\expandafter\def\csname AuditDataR06dstar\endcsname{---}
\expandafter\def\csname AuditDataR06n\endcsname{18{,}904}
\expandafter\def\csname AuditDataR06nreqMed\endcsname{1{,}841}
\expandafter\def\csname AuditDataR06nreqQOne\endcsname{932}
\expandafter\def\csname AuditDataR06nreqQThree\endcsname{6{,}046}
\expandafter\def\csname AuditDataR06tier\endcsname{bits}
\expandafter\def\csname AuditDataR07cls\endcsname{above throughout}
\expandafter\def\csname AuditDataR07dstar\endcsname{0.802904}
\expandafter\def\csname AuditDataR07n\endcsname{12{,}410}
\expandafter\def\csname AuditDataR07nreqMed\endcsname{1{,}841}
\expandafter\def\csname AuditDataR07nreqQOne\endcsname{932}
\expandafter\def\csname AuditDataR07nreqQThree\endcsname{6{,}046}
\expandafter\def\csname AuditDataR07tier\endcsname{bits}
\expandafter\def\csname AuditDataR08cls\endcsname{above throughout}
\expandafter\def\csname AuditDataR08dstar\endcsname{---}
\expandafter\def\csname AuditDataR08n\endcsname{42{,}701}
\expandafter\def\csname AuditDataR08nreqMed\endcsname{742}
\expandafter\def\csname AuditDataR08nreqQOne\endcsname{371}
\expandafter\def\csname AuditDataR08nreqQThree\endcsname{1{,}052}
\expandafter\def\csname AuditDataR08tier\endcsname{family+bits}
\expandafter\def\csname AuditDataR09cls\endcsname{above throughout}
\expandafter\def\csname AuditDataR09dstar\endcsname{---}
\expandafter\def\csname AuditDataR09n\endcsname{42{,}701}
\expandafter\def\csname AuditDataR09nreqMed\endcsname{742}
\expandafter\def\csname AuditDataR09nreqQOne\endcsname{433}
\expandafter\def\csname AuditDataR09nreqQThree\endcsname{1{,}113}
\expandafter\def\csname AuditDataR09tier\endcsname{family+bits}
\expandafter\def\csname AuditDataR10blocker\endcsname{quoted claim appears in neither prose nor a table caption (\S{}3.1): the recorded exact\_quote appears nowhere in the source; 98.6\% is a table cell}
\expandafter\def\csname AuditDataR10kind\endcsname{ineligible}
\expandafter\def\csname AuditDataR10n\endcsname{28{,}659}
\expandafter\def\csname AuditDataR10tier\endcsname{family+bits}
\expandafter\def\csname AuditDataR11blocker\endcsname{no reported or imputable n, no baseline, no numeric delta}
\expandafter\def\csname AuditDataR11kind\endcsname{insufficient reporting}
\expandafter\def\csname AuditDataR11n\endcsname{---}
\expandafter\def\csname AuditDataR11tier\endcsname{bits}
\expandafter\def\csname AuditDataR12cls\endcsname{above throughout}
\expandafter\def\csname AuditDataR12dstar\endcsname{0.908491}
\expandafter\def\csname AuditDataR12n\endcsname{14{,}042}
\expandafter\def\csname AuditDataR12nreqMed\endcsname{742}
\expandafter\def\csname AuditDataR12nreqQOne\endcsname{433}
\expandafter\def\csname AuditDataR12nreqQThree\endcsname{1{,}113}
\expandafter\def\csname AuditDataR12tier\endcsname{family+bits}
\expandafter\def\csname AuditDataR13blocker\endcsname{no on-page baseline accuracy and no computable delta}
\expandafter\def\csname AuditDataR13kind\endcsname{insufficient reporting}
\expandafter\def\csname AuditDataR13n\endcsname{250}
\expandafter\def\csname AuditDataR13tier\endcsname{family+bits}
\expandafter\def\csname AuditDataR14blocker\endcsname{no baseline accuracy stated (Figure 8 chart only)}
\expandafter\def\csname AuditDataR14kind\endcsname{insufficient reporting}
\expandafter\def\csname AuditDataR14n\endcsname{728}
\expandafter\def\csname AuditDataR14tier\endcsname{family+bits}
\expandafter\def\csname AuditDataR15cls\endcsname{above throughout}
\expandafter\def\csname AuditDataR15dstar\endcsname{---}
\expandafter\def\csname AuditDataR15n\endcsname{42{,}701}
\expandafter\def\csname AuditDataR15nreqMed\endcsname{866}
\expandafter\def\csname AuditDataR15nreqQOne\endcsname{619}
\expandafter\def\csname AuditDataR15nreqQThree\endcsname{1{,}237}
\expandafter\def\csname AuditDataR15tier\endcsname{family+bits}
\expandafter\def\csname AuditDataR16cls\endcsname{above throughout}
\expandafter\def\csname AuditDataR16dstar\endcsname{---}
\expandafter\def\csname AuditDataR16n\endcsname{42{,}701}
\expandafter\def\csname AuditDataR16nreqMed\endcsname{742}
\expandafter\def\csname AuditDataR16nreqQOne\endcsname{371}
\expandafter\def\csname AuditDataR16nreqQThree\endcsname{1{,}052}
\expandafter\def\csname AuditDataR16tier\endcsname{family+bits}
\expandafter\def\csname AuditDataR17cls\endcsname{above throughout}
\expandafter\def\csname AuditDataR17dstar\endcsname{---}
\expandafter\def\csname AuditDataR17n\endcsname{28{,}659}
\expandafter\def\csname AuditDataR17nreqMed\endcsname{2{,}081}
\expandafter\def\csname AuditDataR17nreqQOne\endcsname{1{,}332}
\expandafter\def\csname AuditDataR17nreqQThree\endcsname{3{,}644}
\expandafter\def\csname AuditDataR17tier\endcsname{bits+benchmark}

\begin{document}
\maketitle

\begin{abstract}
A fraction of a point of benchmark accuracy is the usual evidence that a
compressed model is equivalent to its original. That quantity is least
informative when two models are most alike: a net delta is what survives
cancellation between opposing per-item changes, and cancellation is most
complete in the regime equivalence claims occupy.

Across an atlas of 1{,}707 paired model-by-task cells mined from public
per-item evaluation dumps (1.3B--405B), churn runs roughly five times the net
accuracy delta, and cells scoring identically to their baseline still disagree
on individual items.

In a preregistered audit of \AuditFrozenCandidates{} equivalence claims from
three registered frames (method papers, model cards, vendor documentation),
\AuditEligible{} are eligible. None states a prospective numerical equivalence
margin, and none releases \emph{task-matched} per-item outputs, though
\AuditPerItemOtherTaskOnly{} release outputs for other tasks only;
\AuditNotAssessable{} report too little to assess numerically, so a reader
cannot check them at any sample size. We audit evidential sufficiency, not
truth: no claim is called false.

We supply the missing instrument: paired equivalence testing at a declared
margin, with certification tables giving the items an evaluation needs,
computed from disagreement observed under compression, not from
independent-binomial variance. A controlled experiment pairs GPTQ and AWQ on
byte-identical calibration samples across five seeds. Under the frozen
eight-cell decision rule H3 is supported: changing the calibration draw was
sufficient to reverse the observed method ordering in 5 of 8 confirmatory
cells.

{\sloppy
The reporting standard we propose is five lines: declare a margin, run the
paired test, report churn beside net delta, cite the sample size you met,
release per-item outputs. It applies to any comparison between two models alike
enough to be worth comparing. All per-item outputs, protocols and code are
released.\par}
\end{abstract}

\section{Introduction}
\label{sec:intro}

%

Compressed language models are frequently released with a sentence of the form
``negligible degradation'' or ``$99.x\%$ recovery'', supported by a difference of
a fraction of a point in aggregate accuracy on a fixed benchmark. That sentence
is an equivalence claim. Equivalence claims have a statistical form: a declared
margin, a test, and a sample size sufficient to reject the composite null that
the difference exceeds the margin. In the sources we audit, that form is almost
never present.

The missing test is the smaller half of the problem. A net accuracy delta is the
residue left after per-item changes in opposite directions cancel, so the
evidence is weakest exactly where it is offered: the more alike two models are,
the more completely their disagreements cancel.
Figure~\ref{fig:cancellation} shows one cell of our controlled experiment doing
this. Two 4-bit quantizations of the same model, GPTQ and AWQ, scored on the
same $1{,}000$ GSM8K items, land $0.58$ points apart. Underneath that gap,
$9.1\%$ of items go from correct to wrong and $8.5\%$ from wrong to correct, so
the reported difference is what is left of two large opposing quantities.
Certifying equivalence at a declared $\pm 2$-point margin at that disagreement
rate needs $2{,}730$ items, nearly three times the number run; that is a
planning requirement, and it says the evaluation cannot support the claim, not
that the two methods differ. The sign of the gap changes with the calibration
draw alone. The cell is the most extreme of the eight registered cells and is
shown because it is legible; all eight appear in panel~D.

%
%
\definecolor{fharm}{RGB}{140,45,20}
\definecolor{fben}{RGB}{125,178,219}
\definecolor{fneutral}{RGB}{110,110,110}
\definecolor{frule}{RGB}{60,60,60}

\begin{figure}[!t]
\centering
\begin{tikzpicture}[
  x=1cm, y=1cm, line width=0.5pt,
  font=\scriptsize,
  panel/.style={draw=frule!35, rounded corners=1pt, line width=0.4pt},
  ttl/.style={font=\scriptsize\bfseries, anchor=west},
  lbl/.style={font=\scriptsize, anchor=west},
  num/.style={font=\scriptsize, anchor=west},
]
\node[panel, minimum width=6.0cm, minimum height=3.75cm, anchor=south west] at (0.0,6.15) {};
\node[ttl] at (0.18,9.6) {A\quad The aggregate view};
\node[lbl, font=\scriptsize\itshape, text width=5.5cm] at (0.18,9.120000000000001) {Two 4-bit methods, one model, one item set.};
\node[lbl] at (0.18,8.03) {GPTQ};
\draw[fill=frule!12, draw=frule!35] (1.32,7.87) rectangle (4.82,8.19);
\draw[fill=fneutral, draw=none] (1.32,7.87) rectangle (3.9198000000000004,8.19);
\node[num, anchor=west] at (4.9,8.03) {74.28\%};
\node[lbl] at (0.18,7.41) {AWQ};
\draw[fill=frule!12, draw=frule!35] (1.32,7.25) rectangle (4.82,7.57);
\draw[fill=fneutral, draw=none] (1.32,7.25) rectangle (3.8995000000000006,7.57);
\node[num, anchor=west] at (4.9,7.41) {73.70\%};
\draw[frule!55, line width=0.4pt] (1.32,6.87) -- (1.32,6.75) -- (4.82,6.75) -- (4.82,6.87);
\node[font=\scriptsize, anchor=north] at (3.0700000000000003,6.73) {0 to 100\% accuracy, $n=1{,}000$ per seed};
\node[font=\scriptsize\bfseries, anchor=west] at (0.18,8.530000000000001) {Gap: 0.58 pp};
\node[panel, minimum width=6.0cm, minimum height=3.75cm, anchor=south west] at (6.6,6.15) {};
\node[ttl] at (6.779999999999999,9.6) {B\quad The same items, paired};
\node[lbl, font=\scriptsize\itshape, text width=5.5cm] at (6.779999999999999,9.120000000000001) {Every item, going from GPTQ to AWQ.};
\draw[fill=fharm, draw=none] (7.0,8.4) rectangle (7.47424,8.82);
\draw[fill=fben, draw=none] (7.47424,8.4) rectangle (7.91832,8.82);
\draw[fill=frule!8, draw=none] (7.91832,8.4) rectangle (12.2,8.82);
\draw[draw=frule!45] (7.0,8.4) rectangle (12.2,8.82);
\node[font=\scriptsize, anchor=west] at (7.99832,8.610000000000001) {82.34\% unchanged};
\node[lbl, anchor=west] at (7.0,8.120000000000001) {\textcolor{fharm}{$\blacksquare$} 9.12\% correct $\to$ wrong};
\node[lbl, anchor=west] at (7.0,7.74) {\textcolor{fben}{$\blacksquare$} 8.54\% wrong $\to$ correct};
\node[lbl, anchor=west] at (7.0,7.34) {net delta $= 8.54 - 9.12 = -0.58$ pp};
\node[lbl, anchor=west, font=\scriptsize\bfseries] at (7.0,6.960000000000001) {churn $= 8.54 + 9.12 = 17.66$\%};
\node[lbl, anchor=west] at (7.0,6.6000000000000005) {28.66\% of answers change in all};
\node[panel, minimum width=6.0cm, minimum height=3.75cm, anchor=south west] at (0.0,1.55) {};
\node[ttl] at (0.18,5.0) {C\quad Could this evaluation certify?};
\node[lbl, font=\scriptsize\itshape, text width=5.5cm] at (0.18,4.52) {Planning requirement at a declared $\pm2$ pp margin.};
\node[lbl] at (0.18,3.76) {items run};
\draw[fill=fneutral, draw=none] (2.28,3.5999999999999996) rectangle (3.3715750915750915,3.9199999999999995);
\node[num] at (3.4515750915750916,3.76) {1{,}000};
\node[lbl] at (0.18,3.1399999999999997) {items required};
\draw[fill=fharm, draw=none] (2.28,2.9799999999999995) rectangle (5.26,3.2999999999999994);
\node[num] at (5.34,3.1399999999999997) {2{,}730};
\node[lbl, text width=5.55cm, anchor=north west] at (0.18,2.77) {Observed disagreement 17.66\%. A planning requirement, not evidence the methods differ.};
\node[panel, minimum width=6.0cm, minimum height=3.75cm, anchor=south west] at (6.6,1.55) {};
\node[ttl] at (6.779999999999999,5.0) {D\quad The calibration draw alone};
\node[lbl, font=\scriptsize\itshape, text width=5.5cm] at (6.779999999999999,4.52) {GPTQ minus AWQ, per calibration seed.};
\draw[frule!70, line width=0.5pt] (6.949999999999999,3.5) -- (11.7,3.5);
\node[font=\scriptsize, anchor=east] at (6.93,3.5) {0};
\draw[fill=fben, draw=none] (7.249999999999999,3.5) rectangle (7.6499999999999995,3.997391304347826);
\node[font=\scriptsize, anchor=south] at (7.449999999999999,4.037391304347826) {2.2};
\node[font=\scriptsize, anchor=north] at (7.449999999999999,2.55) {s0};
\draw[fill=fben, draw=none] (7.97,3.5) rectangle (8.37,3.8617391304347826);
\node[font=\scriptsize, anchor=south] at (8.17,3.9017391304347826) {1.6};
\node[font=\scriptsize, anchor=north] at (8.17,2.55) {s1};
\draw[fill=fharm, draw=none] (8.69,3.5) rectangle (9.089999999999998,2.98);
\node[font=\scriptsize, anchor=north] at (8.889999999999999,2.94) {-2.3};
\node[font=\scriptsize, anchor=north] at (8.889999999999999,2.55) {s2};
\draw[fill=fben, draw=none] (9.41,3.5) rectangle (9.809999999999999,3.5452173913043477);
\node[font=\scriptsize, anchor=south] at (9.61,3.5852173913043477) {0.2};
\node[font=\scriptsize, anchor=north] at (9.61,2.55) {s3};
\draw[fill=fben, draw=none] (10.129999999999999,3.5) rectangle (10.529999999999998,3.771304347826087);
\node[font=\scriptsize, anchor=south] at (10.329999999999998,3.811304347826087) {1.2};
\node[font=\scriptsize, anchor=north] at (10.329999999999998,2.55) {s4};
\node[font=\scriptsize, anchor=west] at (6.8999999999999995,1.9500000000000002) {all cells:};
\node[font=\scriptsize, text=fharm] at (8.18,1.9500000000000002) {$\bullet$};
\node[font=\scriptsize, text=frule!55] at (8.52,1.9500000000000002) {$\circ$};
\node[font=\scriptsize, text=frule!55] at (8.86,1.9500000000000002) {$\circ$};
\node[font=\scriptsize, text=fharm] at (9.2,1.9500000000000002) {$\bullet$};
\node[font=\scriptsize, text=frule!55] at (9.54,1.9500000000000002) {$\circ$};
\node[font=\scriptsize, text=fharm] at (9.879999999999999,1.9500000000000002) {$\bullet$};
\draw[frule!60, line width=0.4pt] (9.879999999999999,1.9500000000000002) circle (0.16);
\node[font=\scriptsize, text=fharm] at (10.219999999999999,1.9500000000000002) {$\bullet$};
\node[font=\scriptsize, text=fharm] at (10.56,1.9500000000000002) {$\bullet$};
\node[font=\scriptsize, anchor=west] at (10.76,1.9500000000000002) {\ \ $\bullet$ reversed};
\draw[frule!25, line width=0.4pt] (0.0,1.25) -- (12.6,1.25);
\node[anchor=north west, inner sep=0, text width=12.6cm, font=\scriptsize] at (0.0,1.1300000000000001) {\textbf{This is not one cell's problem.} Across 1{,}707 paired model-by-task cells mined from public per-item evaluation dumps, median per-item churn is about five times the median net accuracy delta. 145 of those cells post an aggregate accuracy identical to their baseline, and 128 of the 145 still change which items they get right. Ratios are undefined for a zero net delta, so those cells are counted here and excluded from the ratio.};
\end{tikzpicture}
\caption{\textbf{Near-equal aggregate accuracy does not certify interchangeable behavior.} One registered cell: Qwen2.5-7B on GSM8K, 4-bit GPTQ against 4-bit AWQ, paired on byte-identical calibration samples across five seeds. (A)~and~(B)~The aggregate gap of 0.58~pp is the small difference between two large opposing quantities summing to 17.66\%. (C)~The item count needed to certify equivalence at a declared $\pm2$~pp margin is a planning requirement computed at an assumed true difference of zero: it says the evaluation cannot support the claim, not that the methods differ. (D)~The sign changes with the calibration draw alone, and across the eight registered cells the winner reverses in 5 of 8. \textbf{Scope.} This cell is an illustrative example chosen because it is legible, and it is the most extreme of the eight: its churn-to-net-delta ratio is $30.4\times$ against a median of $12.7\times$ across the eight. All eight are shown in~(D) for that reason. The atlas figures in the panel above are observational and describe the public evaluation record rather than a census of compression.}
\label{fig:cancellation}
\end{figure}

One cell settles nothing about the field, so we measured the public record. An
atlas of $1{,}707$ paired model-by-task cells, mined from published per-item
evaluation dumps spanning 1.3B to 405B parameters, puts per-item churn at
roughly five times the net delta, and a substantial minority of its cells post
an accuracy identical to their baseline while still disagreeing on individual
items
(\S\ref{sec:atlas:identical}). Its two strata sit a generation apart in method
and give nearly the same ratio. The atlas is observational: it describes the
evidence the field circulates and supports no causal claim about quantization in
general.

Could a reader check a published equivalence claim against any of this? We
enumerate \AuditFrozenCandidates{} such claims from three
registered sampling frames: method papers, official quantized model cards, and
inference-stack vendor documentation. \AuditIneligible{} of them,
\AuditIneligibleClaim{}, is ruled ineligible against the registered inclusion
rule, leaving \AuditEligible{}. Not one of the \AuditEligible{} declares a
numerical tolerance in advance, and not one releases \emph{task-matched}
per-item outputs, those covering the tasks its claim rests on;
\AuditPerItemOtherTaskOnly{} release outputs for other tasks only. So no
equivalence claim in the sample can be checked against a standard its author set
or rerun by a reader, and a further \AuditNotAssessable{} cannot be assessed
numerically at all from what they report. The remaining \AuditAssessable{} are
classified against an approximate planning threshold at a uniform
\AuditMarginPP{}\,pp margin, and that classification is reported together with
its sensitivity to the imputed disagreement rate (\S\ref{sec:audit:taxonomy},
\S\ref{sec:audit:results}). No claim is described as false: the audited property
is the evidential sufficiency of the reported evaluation, not the truth of the
underlying equivalence, and these are prevalences in this audited sample of
\AuditEligible{} eligible sources, not the literature at large.

The mechanism predicts something sharper, and a preregistered experiment tests
it. If cancellation grows with similarity, the least informative comparison is
not compressed against original but \emph{compressed against compressed}: two
methods at the same bit width, far closer to each other than either is to its
baseline. Pairing GPTQ and AWQ on byte-identical calibration samples across five
seeds and eight model-by-benchmark cells, we observe churn at a median of
$12.7\times$ the net delta, and under the frozen eight-cell decision rule H3 is
supported: within the controlled pipeline, changing the calibration sample was
sufficient to reverse the observed GPTQ--AWQ ranking in \textbf{5 of 8} cells,
at gaps that would ordinarily be reported as equivalence
(\S\ref{sec:minigrid:verdict}). The evidence problem is worst at the point where
a practitioner actually chooses.

Equivalence is a certification problem, so we compute certification tables
giving the number of items an evaluation needs to certify a compressed model
within $\pm m$ points of its baseline, from the per-item disagreement rates
observed under compression, not from independent-binomial variance. The paired
correction is large, and the requirement is set by churn rather than by task
difficulty (\S\ref{sec:certification}).
Per-item flips as a diagnostic for compressed models are due to
\citet{dutta2024flips}, and calibration-\emph{data} effects on quantization
quality are due to \citet{williamsaletras2024}; we claim neither. Our
contributions are (i) the certification tables and the paired-design correction
they quantify (\S\ref{sec:certification}); (ii) the atlas of the public record
and its measured cancellation ratio (\S\ref{sec:atlas}); (iii) the preregistered
audit and its verdict artifact (\S\ref{sec:audit}); and (iv) the seed-paired
experiment (\S\ref{sec:minigrid}). Relative to the closest existing work
\citep{llmaccuracystats2026}, which performs one-sided McNemar
\emph{detection}, ours is the shift from detection to certification: a declared
margin, TOST, and required-$n$ tables.

All of this assumes the evaluation itself holds still, and a small exploratory
study indicates it does not: on one fixed FP16 model, changing the
answer-extraction filter moved reported GSM8K accuracy from $0.232$ to $0.566$
without changing one token of model output. That is a scoring decision, not a
modelling one, and it is larger than every compression effect reported here; it
is also one model on one task (\S\ref{sec:sensitivity}).
Protocols were frozen before the analyses they
govern and deviations are dated amendments (\S\ref{sec:prereg}), the analyst
decisions that moved the headline are reported with the direction they moved it,
and every artifact is archived at \versiondoi{} (\S\ref{sec:artifacts}). The
limits of the exercise are in \S\ref{sec:limitations}.

Those findings imply a reporting standard, and that standard is what this paper
is for. It is five lines long, and short enough to state before the evidence for
it. A compression claim should \textbf{declare a margin},
because ``equivalent'' without $\pm m$ is not testable
(\S\ref{sec:certification}); \textbf{run the paired test} at that margin rather
than reporting a failure to detect a difference (\S\ref{sec:certification});
\textbf{report churn beside net delta}, because they are different quantities
and the second does not summarise the first (\S\ref{sec:atlas});
\textbf{cite the sample size it met} against the count its benchmark family
requires (\S\ref{sec:certification}); and \textbf{release per-item outputs},
which is what converts an assertion into something a third party can check
(\S\ref{sec:audit}). The rest of the paper is the case that all five are needed:
\S\ref{sec:certification} supplies the instrument the margin, the test and the
sample size require; \S\ref{sec:atlas} and \S\ref{sec:minigrid} measure what
reporting churn beside net delta is worth; and \S\ref{sec:audit} measures how
rarely any of the five is met today. \S\ref{sec:conclusion} returns to them with
the evidence for each in hand.
Figure~\ref{fig:standard} shows the five lines applied to one registered cell
beside the aggregate report of the same run, which is what the standard costs
and what it adds.

\definecolor{fharm}{RGB}{140,45,20}
\definecolor{fben}{RGB}{125,178,219}
\definecolor{fneutral}{RGB}{110,110,110}
\definecolor{frule}{RGB}{60,60,60}

\begin{figure}[!tp]
\centering
\begin{tikzpicture}[
  x=1cm, y=1cm, line width=0.5pt,
  font=\scriptsize,
  ttl/.style={font=\scriptsize\bfseries, anchor=north west},
  lbl/.style={font=\scriptsize, anchor=north west},
  lab/.style={font=\scriptsize, anchor=north west,
    execute at begin node={\hyphenpenalty=10000\relax}},
  gone/.style={font=\scriptsize, text=fneutral, anchor=north west},
]
\node[anchor=north west, inner sep=0, text width=12.6cm, font=\scriptsize] at (0,12.07) {Qwen2.5-7B on GSM8K, 4-bit GPTQ against 4-bit AWQ, $1{,}000$ paired items, five calibration seeds. One evaluation, reported twice. Column~A is the whole of the report; column~B is the same run, same items, same numbers, under the five-line standard.};
\node[ttl, text width=3.06cm, inner sep=0] at (2.86,10.85) {A\quad As reported today};
\node[ttl, text width=6.24cm, inner sep=0] at (6.3,10.85) {B\quad The same evaluation, reported to the standard};
\draw[frule!45, line width=0.5pt] (0,10.09) -- (12.6,10.09);
\node[lab, text width=2.52cm, inner sep=0] at (0.04,9.91) {\textbf{1}\quad declare a margin};
\node[gone, text width=3.06cm, inner sep=0] at (2.86,9.91) {\textemdash\ none declared};
\node[lbl, text width=6.24cm, inner sep=0] at (6.3,9.91) {$\pm2$ pp, declared in advance};
\draw[frule!18, line width=0.3pt] (0,9.07) -- (12.6,9.07);
\node[lab, text width=2.52cm, inner sep=0] at (0.04,8.98) {\textbf{2}\quad run the paired equivalence test at that margin};
\node[gone, text width=3.06cm, inner sep=0] at (2.86,8.98) {\textemdash\ no equivalence test is reported};
\draw[fill=frule!8, draw=none] (7.065,8.205) rectangle (10.935,8.505);
\draw[frule!70, line width=0.5pt, dash pattern=on 1.4pt off 1.0pt] (7.065,8.105) -- (7.065,8.605);
\draw[frule!70, line width=0.5pt, dash pattern=on 1.4pt off 1.0pt] (10.935,8.105) -- (10.935,8.605);
\node[font=\scriptsize, anchor=south] at (9.000,8.625) {declared equivalence region, $\pm2$ pp};
\draw[frule!40, line width=0.4pt] (6.400,8.185) -- (11.600,8.185);
\node[font=\scriptsize, anchor=north] at (7.065,8.145) {$-2$};
\node[font=\scriptsize, anchor=north] at (9.000,8.145) {$0$};
\node[font=\scriptsize, anchor=north] at (10.935,8.145) {$+2$};
\draw[frule!85, line width=0.9pt] (6.833,8.355) -- (10.296,8.355);
\draw[frule!85, line width=0.9pt] (6.833,8.265) -- (6.833,8.445);
\draw[frule!85, line width=0.9pt] (10.296,8.265) -- (10.296,8.445);
\draw[fill=fharm, draw=fharm] (8.439,8.355) circle (0.075);
\node[font=\scriptsize, anchor=west] at (11.700,8.355) {pp};
\node[lbl, text width=6.24cm, inner sep=0] at (6.3,7.52) {$\Delta = -0.58$ pp (AWQ $-$ GPTQ), 95\% paired bootstrap CI $[-2.24, +1.34]$ pp. No TOST verdict is recorded for this cell, and this is not the 90\% interval TOST would read.};
\draw[frule!18, line width=0.3pt] (0,6.01) -- (12.6,6.01);
\node[lab, text width=2.52cm, inner sep=0] at (0.04,5.92) {\textbf{3}\quad report churn beside net delta};
\node[gone, text width=3.06cm, inner sep=0] at (2.86,5.92) {net delta $-0.58$ pp \\ (GPTQ 74.28\%, AWQ 73.70\%)};
\node[lbl, text width=6.24cm, inner sep=0] at (6.3,5.92) {net delta $-0.58$ pp \emph{and} churn 17.66\%: 9.12\% correct $\to$ wrong, 8.54\% wrong $\to$ correct. The net is what is left of the two.};
\draw[frule!18, line width=0.3pt] (0,4.41) -- (12.6,4.41);
\node[lab, text width=2.52cm, inner sep=0] at (0.04,4.32) {\textbf{4}\quad cite the sample size met against the size required};
\node[gone, text width=3.06cm, inner sep=0] at (2.86,4.32) {$n = 1{,}000$; no requirement cited};
\node[lbl, text width=6.24cm, inner sep=0] at (6.3,4.32) {$1{,}000$ items evaluated against $2{,}730$ required at $\pm2$ pp. That is a planning requirement at an assumed true difference of zero: it says this evaluation cannot support the claim, and it is \emph{not} evidence that the two methods differ.};
\draw[frule!18, line width=0.3pt] (0,2.48) -- (12.6,2.48);
\node[lab, text width=2.52cm, inner sep=0] at (0.04,2.39) {\textbf{5}\quad release per-item outputs};
\node[gone, text width=3.06cm, inner sep=0] at (2.86,2.39) {\textemdash\ not released};
\node[lbl, text width=6.24cm, inner sep=0] at (6.3,2.39) {released: 88 per-item cell JSONL files, one per cell of the controlled experiment};
\draw[frule!25, rounded corners=1pt, line width=0.4pt] (2.72,10.05) rectangle (6.0600000000000005,1.52);
\draw[frule!45, line width=0.5pt] (0,1.48) -- (12.6,1.48);
\node[anchor=north west, inner sep=0, text width=12.6cm, font=\scriptsize] at (0,1.34) {\textbf{Nothing in column~B is a new experiment.} Every quantity there was already produced by the run column~A summarises; what changes is what is reported. Column~A is complete as drawn: there is no further evidence behind it. Across the 16 eligible sources of our audit, 0 declare a prospective numerical margin and 0 release task-matched per-item outputs.};
\end{tikzpicture}
\caption{\textbf{The same evaluation, before and after the reporting standard.} Both columns describe one registered cell: Qwen2.5-7B on GSM8K, 4-bit GPTQ against 4-bit AWQ, $1{,}000$ paired items. \textbf{A}~is the aggregate report (two accuracies and a net delta of $0.58$~pp), and it is complete: there is no further evidence behind it. \textbf{B}~is the same run reported under the five lines of \S\ref{sec:conclusion}, and it needs no additional experiment. Row~2 draws the declared $\pm2$~pp margin together with the observed paired delta and the committed 95\% percentile-bootstrap interval for it; \emph{no TOST verdict exists for this cell}, and that interval is not the 90\% two-sided interval TOST at one-sided $\alpha=.05$ requires (\S\ref{sec:cert:method}), so it settles the equivalence question in neither direction. Row~4's $2{,}730$ is a \emph{planning} requirement computed at an assumed true difference of zero: it says this evaluation cannot support an equivalence claim at $\pm2$~pp, and it is not evidence that the two methods differ. Figure~\ref{fig:cancellation} decomposes the same cell; every value here is read from the artifacts recorded in \texttt{paper/figures/fig2\_values.json}.}
\label{fig:standard}
\end{figure}

%
%
%

\section{Related work and positioning}
\label{sec:related}


\citet{dutta2024flips} show that compressed models can match baseline aggregate
accuracy while flipping many individual answers, and propose flips and KL
divergence as complementary metrics. \textbf{The flips metric is theirs, and we
do not claim it.} What we add is that decomposition measured across the public
record at scale (\S\ref{sec:atlas}), the observed flip-rate distribution used as
the \emph{variance model} for equivalence certification
(\S\ref{sec:certification}), and the audit that follows from it
(\S\ref{sec:audit}). Concurrent work reaches the same premise independently:
\citet{rababah2026illusion}, posted two days before our preregistration froze
and unknown to us until the prior-art sweep of 2026-07-24, define
\emph{correctness agreement}, the joint-correct cell of the same $2\times2$
table our accuracy-state churn is built from and interconvertible with it given
the marginal accuracies, over a quantization family disjoint from ours. We read
this as external corroboration of the premise, not competition on it, and we
make no priority claim over it: no committed artifact in this repository
predates its posting. What it does not supply is the machinery that turns the
shared premise into a decision. Two further concurrent studies measure per-item
change on adjacent objects, on model versions and on community checkpoints
(Appendix~\ref{app:related:concurrent}).

\citet{williamsaletras2024} establish that the calibration \emph{data} affects
quantized model quality, finding substantial variation in downstream task
performance across calibration sets. \textbf{Calibration-data effects are
theirs.} Our question is orthogonal and finer-grained: holding the calibration
corpus fixed, does the calibration \emph{sample seed} change the \emph{ranking}
of two methods (\S\ref{sec:minigrid})? The sweep found no prior work that varies
a calibration seed with the corpus held fixed and asks whether the resulting
instability reorders two methods. The nearest antecedents vary the calibration
set itself and disagree with each other about how much that matters:
\citet{williamsaletras2024} find the effect substantial, and
\citet{paglieri2024outliers} find it diminishing on modern models. That
disagreement is the strongest available objection to this paper, since a field
insensitive to the calibration set would seem unlikely to reorder on the seed.
Appendix~\ref{app:reconcile} answers it at length: both results hold at once,
because neither design measures the \emph{gap between} two methods, which is the
quantity a practitioner choosing between them depends on, and our own eight
cells show a seed-induced range at least as large as that gap in 7 of 8.

The closest existing work is \citet{llmaccuracystats2026}, which brings
one-sided McNemar \emph{detection} to LLM accuracy comparisons and ships it in
the evaluation harness the field already uses. We agree with its diagnosis and
differ in the question asked. Detection asks whether there is evidence of a
difference; a non-significant result is not equivalence, and at the sample sizes
in \S\ref{sec:audit} it is frequently just an absence of resolution. Our deltas
are TOST certification at a \emph{declared} margin, required-$n$ tables computed
from empirical churn instead of independent-binomial variance with the
$2.3$--$14.7\times$ correction that implies, and the audit of published claims
that is the empirical case the distinction matters. \citet{helcig2026slq} occupy
the phrase this paper is about, defining losslessness for quantized LLMs and
building a method reaching it; their question is the complement of ours and they
compute no equivalence test, power or required $n$
(Appendix~\ref{app:related:lossless}). The constructive-audit genre is well
established \citep{dodge2019showyourwork,marie2021mtaudit} and has begun to
adopt preregistration directly, which is the precedent for \S\ref{sec:prereg}.
We follow it in framing and in tone: the finding is about what the field
reports, not about whether individual authors are right, and we audit the
equivalence \emph{language} these papers use and the evidence offered for it
rather than the methods themselves (Appendices~\ref{app:related:genre}
and~\ref{app:related:families}).

\section{Paired certification: what an equivalence claim has to show}

\label{sec:certification}

\subsection{Detection is not certification}

The statistical question behind ``near-lossless'' is not the one usually
answered. A McNemar test, including the one-sided variant developed by the
closest existing work on LLM accuracy statistics
\citep{llmaccuracystats2026}, asks whether there is evidence that the
compressed model differs from its baseline. Failing to find such evidence is not
evidence of equivalence; with a small enough evaluation, nothing is detectable.
Our registration commits to this in advance: ``we will not interpret failure to
reject a difference as equivalence''.
Certification asks the other question: \emph{is the difference provably smaller
than a margin I declare in advance?} The standard instrument is TOST: two
one-sided tests at level $\alpha$ each, rejecting the composite null
$|\Delta| \geq m$ in favour of equivalence within $\pm m$. TOST converts an
equivalence claim into something with a sample-size requirement, and this
section computes that requirement empirically.

\subsection{Method}
\label{sec:cert:method}

For a discordance rate $p_d$ and margin $m$, under TOST at one-sided
$\alpha=.05$ (equivalent to requiring a 90\% two-sided confidence interval to
fall inside $\pm m$) with 80\% power,
\begin{align}
\mathrm{sd}_{\mathrm{paired}}      &= \sqrt{p_d}, &
\mathrm{sd}_{\mathrm{independent}} &= \sqrt{2p(1-p)}, \label{eq:sds}\\[2pt]
n_{\mathrm{req}} &= \left\lceil \left(\frac{(z_{1-\alpha} + z_{1-\beta})\,\mathrm{sd}}{m}\right)^{\!2} \right\rceil ,
\label{eq:nreq}
\end{align}
with $z_{1-\alpha} = 1.6449$ one-sided and $z_{1-\beta} = 0.8416$.
The paired standard deviation is the whole point: it is set by $p_d$, the rate
at which the two models disagree item by item, and not by how hard the benchmark
is. The independent form depends instead on $p$, the family's median baseline
accuracy. Appendix~\ref{app:cert:method} derives both, justifies the one-sided
$z$, and states how the discordance quartiles are read off the atlas
(\S\ref{sec:atlas}).

Equation~\eqref{eq:nreq} is the standard TOST planning sample size under an
assumed \emph{true} paired accuracy difference of zero. It answers ``how many
items would this benchmark need for an equivalence test to have 80\% power,
given how much models of this kind disagree item-by-item''. It does \emph{not}
answer ``how many items would certify the delta this source observed''. Under a
true difference $\delta$ the quantity TOST must resolve is the separation
$m - |\delta|$ instead of $m$, so the requirement grows without bound as
$|\delta|$ approaches the margin. Every $n_{\mathrm{req}}$ reported in this
paper is therefore a \emph{lower bound} on what a non-zero true difference would
demand, and these are design tables, not retrospective certification of any
particular reported result.

\begin{table}[!t]
\centering
\small
\caption{Items required to certify equivalence within $\pm 2$\,pp under TOST at
one-sided $\alpha=.05$ (a 90\% two-sided interval) with 80\% power, assuming a
true difference of zero, by benchmark family, at the 25th/50th/75th
percentiles of the discordance rates the atlas observes for that family. The
naive column is the same requirement computed by treating the two runs as
independent samples. Eleven benchmark families plus the pooled row.}
\label{tab:certification}
\setlength{\tabcolsep}{2pt}
\small
\begin{tabular}{lr>{\centering\arraybackslash}p{2.6cm}>{\centering\arraybackslash}p{2.65cm}rr}
\toprule
Benchmark & Atlas cells & Discordance p25/med/p75 & Required $n$ p25/\textbf{med}/p75 & Naive & Advantage \\
\midrule
musr          &  24 & 0.015 / 0.034 / 0.044 & 232 / \textbf{519} / 681 & 7{,}617 & $14.7\times$ \\
gpqa          &  24 & 0.035 / 0.048 / 0.067 & 545 / \textbf{749} / 1{,}035 & 7{,}233 & $9.7\times$ \\
bbh           & 192 & 0.024 / 0.044 / 0.060 & 371 / \textbf{681} / 928 & 6{,}492 & $9.5\times$ \\
mmlu\_pro     &   5 & 0.048 / 0.053 / 0.059 & 739 / \textbf{827} / 913      & 7{,}695 & $9.3\times$ \\
hellaswag     &  23 & 0.031 / 0.045 / 0.099 & 477 / \textbf{695} / 1{,}538 & 4{,}905 & $7.1\times$ \\
arc\_challenge&  17 & 0.072 / 0.079 / 0.099 & 1{,}112 / \textbf{1{,}218} / 1{,}536 & 7{,}363 & $6.0\times$ \\
ifeval        &   8 & 0.048 / 0.052 / 0.072 & 736 / \textbf{800} / 1{,}108 & 4{,}211 & $5.3\times$ \\
mmlu          & 1{,}311 & 0.093 / 0.140 / 0.259 & 1{,}432 / \textbf{2{,}164} / 4{,}005 & 7{,}727 & $3.6\times$ \\
winogrande    &  23 & 0.067 / 0.092 / 0.221 & 1{,}031 / \textbf{1{,}416} / 3{,}422 & 5{,}600 & $4.0\times$ \\
math          &  56 & 0.107 / 0.141 / 0.169 & 1{,}661 / \textbf{2{,}186} / 2{,}610 & 5{,}222 & $2.4\times$ \\
gsm8k         &  24 & 0.040 / 0.077 / 0.198 & 619 / \textbf{1{,}184} / 3{,}068 & 2{,}671 & $2.3\times$ \\
\midrule
\textbf{ALL (pooled)} & \textbf{1{,}707} & 0.064 / 0.120 / 0.225 & 994 / \textbf{1{,}855} / 3{,}478 & 7{,}722 & $\mathbf{4.2\times}$ \\
\bottomrule
\end{tabular}
\end{table}

\subsection{Reading the table}

Pick the benchmark family and the margin you intend to certify, and read the
item count under optimistic, typical or pessimistic compression behaviour for
that family. \emph{MMLU, 2\,pp margin, typical discordance $\Rightarrow$
evaluate at least 2{,}164 items.} Ignoring pairing would have demanded
$7{,}727$, or $3.6\times$ the compute for the same conclusion. Tightening the
margin to 1\,pp raises the requirement to $8{,}656$ items and relaxing it to
3\,pp lowers it to $962$: the margin is the practitioner's declaration and its
cost is quadratic, so parity within 1\,pp is four times as expensive to certify
as parity within 2\,pp. Across families the paired design's advantage ranges
from $2.3\times$ (GSM8K) to $14.7\times$ (MuSR) and sits at $4.2\times$ pooled
over 1{,}707 cells, so a practitioner using it reaches the same equivalence
conclusion on roughly a quarter of the evaluation budget.

\subsection{The requirement is driven by churn, not by difficulty}

The ordering in Table~\ref{tab:certification} is not the intuitive one, and this
is the section's main conceptual point.

MMLU needs about \textbf{2{,}164} items at a 2\,pp margin. GPQA needs
\textbf{749}. GPQA is by any ordinary account the harder benchmark, and its
median baseline accuracy in the atlas is 0.373 against MMLU's 0.492, on a task
designed to resist exactly the models that saturate MMLU. Difficulty is not what
sets the requirement. What sets it is how much the compressed model's per-item
correctness \emph{churns} relative to the baseline: MMLU's typical discordance
is 0.140 against GPQA's 0.048, so MMLU's paired variance is nearly three times
larger and its required sample size scales with it. A practitioner therefore
cannot reason about evaluation size from intuitions about task hardness,
headroom or score level. Two benchmarks with similar apparent difficulty can
differ by a factor of three in the evidence they require, and the only way to
know which is which is to measure churn. That is what the atlas is for, and it
is why these tables are empirical and not analytic.

\subsection{Which benchmarks this fits, and how to size an evaluation}
\label{sec:cert:applicability}

The framework asks one thing of a benchmark: that the harness scoring it assign
each item a correct or incorrect state. Multiple-choice suites supply that
(MMLU, MMLU-Pro, ARC-Challenge, HellaSwag, WinoGrande, GPQA, MuSR), as do
rule-scored instruction following (IFEval), exact-match reasoning suites (BBH,
GSM8K, MATH), and code benchmarks judged pass or fail on a single sample
(HumanEval and MBPP at $k=1$). Summarisation and translation scored by ROUGE,
BLEU or COMET do not, and neither do LLM-as-a-judge ratings, pass@$k$ for
$k>1$, or perplexity-style measures such as WikiText-2: each returns a graded
per-item score, or no per-item score at all. Those need the continuous
adaptation sketched in \S\ref{sec:limitations:continuous} and
Appendix~\ref{app:continuous}, which this paper states as mathematics and does
not validate.

Fitting the framework is not the same as having a number for it. Eleven
benchmark families carry measured discordance rows in
Table~\ref{tab:certification}, and for those a practitioner can look up a
required $n$ today. A binary benchmark with no row, HumanEval and MBPP among
them, is covered by the same arithmetic, but nothing here supplies its
discordance rate: that has to be measured on the practitioner's own model pair
first, which is the boundary against extrapolation drawn in
\S\ref{sec:cert:caveats}.

To size an evaluation:
\begin{enumerate}
\item Declare the margin $m$ before running anything. It states the difference
you are prepared to call immaterial, in accuracy points, and its cost is
quadratic.
\item Find the benchmark family's row in Table~\ref{tab:certification}. If it
has no row, measure discordance on your own pair and use
Equations~\eqref{eq:sds} and~\eqref{eq:nreq} directly.
\item Read the required $n$ at median discordance for the typical case, and at
p75 for a conservative one. The p25 column describes optimistic compression
behaviour and is not a planning figure.
\item Compare that against the number of items you can actually afford to run.
\item If you fall short, evaluate more items, or widen $m$ and report which $m$
you used, or report the paired difference and its interval without calling it
certified. Reporting a non-significant difference as equivalence is not one of
the options.
\end{enumerate}

Figure~\ref{fig:requiredn} is Table~\ref{tab:certification} drawn for step~3:
the families ordered by what they require, with the p25--p75 band each
requirement sits in.

\definecolor{fharm}{RGB}{140,45,20}
\definecolor{fben}{RGB}{125,178,219}
\definecolor{fneutral}{RGB}{110,110,110}
\definecolor{frule}{RGB}{60,60,60}

\begin{figure}[!tp]
\centering
\begin{tikzpicture}[
  x=1cm, y=1cm, line width=0.5pt,
  font=\scriptsize,
  hd/.style={font=\scriptsize\bfseries},
  lbl/.style={font=\scriptsize, anchor=west},
  num/.style={font=\scriptsize, anchor=east},
]
\node[hd, anchor=north] at (5.91,8.49) {items required to certify equivalence within $\pm2$ pp};
\node[font=\scriptsize\itshape, anchor=north] at (5.91,8.21) {p25 \rule[0.10em]{0.45cm}{0.30ex} p75 band, $\bullet$ median, log scale};
\node[hd, anchor=west] at (0.04,7.75) {benchmark family};
\node[hd, anchor=east] at (10.18,7.75) {med.};
\node[hd, anchor=east] at (11.13,7.75) {churn};
\node[hd, anchor=east] at (12.58,7.75) {base acc};
\draw[frule!45, line width=0.5pt] (0,7.51) -- (12.6,7.51);
\node[lbl] at (0.04,7.258) {MuSR};
\draw[fill=frule!20, draw=none] (2.907,7.192) rectangle (4.696,7.323);
\draw[frule!60, line width=0.4pt] (2.907,7.173) -- (2.907,7.343);
\draw[frule!60, line width=0.4pt] (4.696,7.173) -- (4.696,7.343);
\draw[fill=fharm, draw=fharm] (4.244,7.258) circle (0.070);
\node[num] at (10.18,7.258) {519};
\node[num] at (11.13,7.258) {0.034};
\node[num] at (12.58,7.258) {0.560};
\node[lbl] at (0.04,6.863) {BBH};
\draw[fill=frule!20, draw=none] (3.687,6.798) rectangle (5.210,6.928);
\draw[frule!60, line width=0.4pt] (3.687,6.778) -- (3.687,6.948);
\draw[frule!60, line width=0.4pt] (5.210,6.778) -- (5.210,6.948);
\draw[fill=fharm, draw=fharm] (4.696,6.863) circle (0.070);
\node[num] at (10.18,6.863) {681};
\node[num] at (11.13,6.863) {0.044};
\node[num] at (12.58,6.863) {0.700};
\node[lbl] at (0.04,6.468) {HellaSwag};
\draw[fill=frule!20, draw=none] (4.104,6.403) rectangle (6.049,6.533);
\draw[frule!60, line width=0.4pt] (4.104,6.383) -- (4.104,6.553);
\draw[frule!60, line width=0.4pt] (6.049,6.383) -- (6.049,6.553);
\draw[fill=fharm, draw=fharm] (4.730,6.468) circle (0.070);
\node[num] at (10.18,6.468) {695};
\node[num] at (11.13,6.468) {0.045};
\node[num] at (12.58,6.468) {0.802};
\node[lbl] at (0.04,6.073) {GPQA};
\draw[fill=frule!20, draw=none] (4.326,6.008) rectangle (5.391,6.138);
\draw[frule!60, line width=0.4pt] (4.326,5.988) -- (4.326,6.158);
\draw[frule!60, line width=0.4pt] (5.391,5.988) -- (5.391,6.158);
\draw[fill=fharm, draw=fharm] (4.854,6.073) circle (0.070);
\node[num] at (10.18,6.073) {749};
\node[num] at (11.13,6.073) {0.048};
\node[num] at (12.58,6.073) {0.373};
\node[lbl] at (0.04,5.678) {IFEval};
\draw[fill=frule!20, draw=none] (4.825,5.613) rectangle (5.505,5.743);
\draw[frule!60, line width=0.4pt] (4.825,5.593) -- (4.825,5.763);
\draw[frule!60, line width=0.4pt] (5.505,5.593) -- (5.505,5.763);
\draw[fill=fharm, draw=fharm] (4.963,5.678) circle (0.070);
\node[num] at (10.18,5.678) {800};
\node[num] at (11.13,5.678) {0.052};
\node[num] at (12.58,5.678) {0.837};
\node[lbl] at (0.04,5.283) {MMLU-Pro};
\draw[fill=frule!20, draw=none] (4.832,5.218) rectangle (5.183,5.348);
\draw[frule!60, line width=0.4pt] (4.832,5.198) -- (4.832,5.368);
\draw[frule!60, line width=0.4pt] (5.183,5.198) -- (5.183,5.368);
\draw[fill=fharm, draw=fharm] (5.019,5.283) circle (0.070);
\node[num] at (10.18,5.283) {827};
\node[num] at (11.13,5.283) {0.053};
\node[num] at (12.58,5.283) {0.533};
\node[lbl] at (0.04,4.888) {GSM8K};
\draw[fill=frule!20, draw=none] (4.537,4.823) rectangle (7.197,4.953);
\draw[frule!60, line width=0.4pt] (4.537,4.803) -- (4.537,4.973);
\draw[frule!60, line width=0.4pt] (7.197,4.803) -- (7.197,4.973);
\draw[fill=fharm, draw=fharm] (5.615,4.888) circle (0.070);
\node[num] at (10.18,4.888) {1{,}184};
\node[num] at (11.13,4.888) {0.077};
\node[num] at (12.58,4.888) {0.096};
\node[lbl] at (0.04,4.493) {ARC-Challenge};
\draw[fill=frule!20, draw=none] (5.511,4.428) rectangle (6.047,4.558);
\draw[frule!60, line width=0.4pt] (5.511,4.408) -- (5.511,4.578);
\draw[frule!60, line width=0.4pt] (6.047,4.408) -- (6.047,4.578);
\draw[fill=fharm, draw=fharm] (5.662,4.493) circle (0.070);
\node[num] at (10.18,4.493) {1{,}218};
\node[num] at (11.13,4.493) {0.079};
\node[num] at (12.58,4.493) {0.609};
\node[lbl] at (0.04,4.098) {WinoGrande};
\draw[fill=frule!20, draw=none] (5.385,4.033) rectangle (7.378,4.163);
\draw[frule!60, line width=0.4pt] (5.385,4.013) -- (5.385,4.183);
\draw[frule!60, line width=0.4pt] (7.378,4.013) -- (7.378,4.183);
\draw[fill=fharm, draw=fharm] (5.912,4.098) circle (0.070);
\node[num] at (10.18,4.098) {1{,}416};
\node[num] at (11.13,4.098) {0.092};
\node[num] at (12.58,4.098) {0.762};
\node[lbl] at (0.04,3.703) {MMLU};
\draw[fill=frule!20, draw=none] (5.931,3.638) rectangle (7.640,3.768);
\draw[frule!60, line width=0.4pt] (5.931,3.618) -- (5.931,3.788);
\draw[frule!60, line width=0.4pt] (7.640,3.618) -- (7.640,3.788);
\draw[fill=fharm, draw=fharm] (6.617,3.703) circle (0.070);
\node[num] at (10.18,3.703) {2{,}164};
\node[num] at (11.13,3.703) {0.140};
\node[num] at (12.58,3.703) {0.492};
\node[lbl] at (0.04,3.308) {MATH};
\draw[fill=frule!20, draw=none] (6.177,3.243) rectangle (6.928,3.373);
\draw[frule!60, line width=0.4pt] (6.177,3.223) -- (6.177,3.393);
\draw[frule!60, line width=0.4pt] (6.928,3.223) -- (6.928,3.393);
\draw[fill=fharm, draw=fharm] (6.634,3.308) circle (0.070);
\node[num] at (10.18,3.308) {2{,}186};
\node[num] at (11.13,3.308) {0.141};
\node[num] at (12.58,3.308) {0.215};
\draw[frule!30, line width=0.4pt] (0,3.110) -- (12.6,3.110);
\node[lbl] at (0.04,2.913) {\textbf{ALL (pooled)}};
\draw[fill=frule!20, draw=none] (5.324,2.848) rectangle (7.405,2.978);
\draw[frule!60, line width=0.4pt] (5.324,2.828) -- (5.324,2.998);
\draw[frule!60, line width=0.4pt] (7.405,2.828) -- (7.405,2.998);
\draw[fill=fharm, draw=fharm] (6.361,2.913) circle (0.070);
\node[num] at (10.18,2.913) {\textbf{1{,}855}};
\node[num] at (11.13,2.913) {0.120};
\node[num] at (12.58,2.913) {0.515};
\draw[frule!16, line width=0.3pt] (2.660,2.715) -- (2.660,7.515);
\draw[frule!16, line width=0.3pt] (4.182,2.715) -- (4.182,7.515);
\draw[frule!16, line width=0.3pt] (5.334,2.715) -- (5.334,7.515);
\draw[frule!16, line width=0.3pt] (6.486,2.715) -- (6.486,7.515);
\draw[frule!16, line width=0.3pt] (8.008,2.715) -- (8.008,7.515);
\draw[frule!16, line width=0.3pt] (9.160,2.715) -- (9.160,7.515);
\draw[frule!30, line width=0.4pt] (0,2.715) -- (12.6,2.715);
\node[lbl] at (0.04,2.278) {pooled, by margin};
\draw[frule!55, line width=0.4pt, {Latex[length=1.4mm]}-{Latex[length=1.4mm]}] (5.015,2.278) -- (8.664,2.278);
\draw[fill=fneutral, draw=fneutral] (8.664,2.278) circle (0.055);
\node[font=\scriptsize, anchor=south] at (8.664,2.348) {$\pm1$};
\draw[fill=fneutral, draw=fneutral] (6.361,2.278) circle (0.055);
\node[font=\scriptsize, anchor=south] at (6.361,2.348) {$\pm2$};
\draw[fill=fneutral, draw=fneutral] (5.015,2.278) circle (0.055);
\node[font=\scriptsize, anchor=south] at (5.015,2.348) {$\pm3$};
\node[num] at (12.58,2.278) {7{,}420 / 1{,}855 / 825};
\draw[frule!45, line width=0.5pt] (0,2.040) -- (12.6,2.040);
\node[font=\scriptsize, anchor=north] at (2.660,2.000) {200};
\node[font=\scriptsize, anchor=north] at (4.182,2.000) {500};
\node[font=\scriptsize, anchor=north] at (5.334,2.000) {1{,}000};
\node[font=\scriptsize, anchor=north] at (6.486,2.000) {2{,}000};
\node[font=\scriptsize, anchor=north] at (8.008,2.000) {5{,}000};
\node[font=\scriptsize, anchor=north] at (9.160,2.000) {10{,}000};
\node[font=\scriptsize, anchor=north east] at (12.6,2.000) {items};
\node[anchor=north west, inner sep=0, text width=12.6cm, font=\scriptsize] at (0,1.620) {\textbf{The ordering is set by churn, not by difficulty.} MMLU needs $2{,}164$ items at $\pm2$ pp and GPQA needs $749$, although GPQA is the harder benchmark: its median baseline accuracy in the atlas is $0.373$ against MMLU's $0.492$. What separates them is the churn column, $0.140$ against $0.048$. Read the row, not the leaderboard.};
\end{tikzpicture}
\caption{\textbf{How many items an equivalence claim needs, by benchmark family.} The same data as Table~\ref{tab:certification}, drawn for reference at the registered $\pm2$~pp margin: TOST at one-sided $\alpha=.05$ with 80\% power, at an assumed true difference of zero. Each row is one family; the band spans the requirement at the 25th and 75th percentiles of the discordance the atlas observes for that family and the disc is the median. Families are ordered by that median, and the two right-hand columns are what the ordering does and does not follow: it tracks median churn and not median baseline accuracy. \textbf{Other margins.} The requirement scales as $1/m^2$, so on this log axis a change of margin translates the whole chart without reshaping it: across all twelve rows the $\pm1$~pp requirement is $4.00$--$4.00\times$ the $\pm2$~pp one and the $\pm3$~pp requirement is $0.44$--$0.45\times$ it. The ruler row marks the pooled median at all three margins on this same axis ($825$, $1{,}855$, $7{,}420$ items). \textbf{Scope.} These are planning sizes, and a lower bound under any non-zero true difference (\S\ref{sec:cert:method}); meeting one makes the equivalence test informative, it does not certify anything on its own. Families with fewer than four analysable atlas cells are absent, and two of the rows shown rest on thin evidence (\S\ref{sec:cert:caveats}).}
\label{fig:requiredn}
\end{figure}

\subsection{Scope and caveats}
\label{sec:cert:caveats}

These caveats travel with the table, and any reuse of it should carry them.
Families with fewer than four analysable cells are omitted, because a quartile
over three points is not an empirical distribution; this drops nothing from the
registered set, and all surviving families appear in
Table~\ref{tab:certification}. Two of the surviving rows rest on thin evidence
and are indicative only, and family aggregation and the excluded feasibility
probes are covered in Appendices~\ref{app:cert:method}
and~\ref{app:cert:caveats-detail}. Finally, these are certification sample
sizes, not detection sample sizes: they certify equivalence within $\pm m$, the
$n$ required to \emph{detect} a difference at the same margin is larger, and
reporting one as the other is the error this section exists to prevent.

This section \emph{does not support}: extrapolation to benchmark families
absent from the atlas; extrapolation to compression methods absent from the
atlas population (\S\ref{sec:atlas:caveats}); or any claim that a model meeting
these counts is equivalent, because meeting the count makes the test
\emph{informative}, and the test still has to be run and to pass.

\subsection{Preregistration compliance, and the choice that moved the headline}
\label{sec:prereg}
\label{sec:prereg:choices}

Each artifact this paper depends on was committed before the analysis it
governs could run, and each frozen file carries a \emph{Dated Amendments}
section, so deviations are appended and never edited in, each recording whether
results had been inspected before the decision.
Appendix~\ref{app:prereg-detail} carries the bookkeeping: the freeze timeline
against its commits, the content hashes a reader can check the inputs against,
what each freeze fixes in advance, the disclosed points of pre-registration data
contact, the inspection discipline observed during grid execution, the H3 rule
as registered, and the seven interpretive rulings in full.

%
%
One ruling moved the headline, and the order in which its numbers arrived is
itself the disclosure. The frozen §4 names the registered
\AuditMarginPP{}\,pp margin first and adds ``(and at the claim's own margin when
it states one)''; we took that as two live readings and chose the
claim-specific one, which raised the count of claims below the planning
threshold from $K = 1$ of 12 to $K = 5$, and the metric-compatibility ruling on
R04 then gave the $K = 4$ that was published for ten days. That reading was
wrong: no audited source states a margin, so the conditional never fires, and
what the implementation had substituted was each source's own largest reported
delta, which is a measured outcome and not a decision rule its authors adopted.
Amendment~2 (signed 2026-07-31) withdrew it and returned the analysis to the
uniform registered margin, where the rev-3 recomputation gives
\textbf{\AuditBelowThresholdAtMedian{} of \AuditAssessable{} assessable claims}
below the threshold, the same claim the very first pass had flagged. The full
reasoning, the other six rulings, and the readings as they ship in the released
CSV are in Appendix~\ref{app:prereg:choices}.

\subsection{Atlas validation and population correction}
\label{sec:prereg-spotcheck}

The protocol required an independent spot-check before any atlas number could be
quoted externally. It re-derived ten stratified cells from a fresh
reimplementation of the registered definitions instead of a rerun of our own
code, and all 262 compared fields reconciled exactly: the arithmetic was right,
and the \emph{population} was not, because two defects had silently dropped
cells non-randomly. The repair executed a rule that was already binding and
under-implemented, it enlarged the analysable population by 44\%, it left the
audit's headline verdicts unmoved, and it was made \emph{after} results had been
inspected, which we disclose instead of presenting it as a pre-specified step.
Both revisions are published and the delta between them is reported
(Appendix~\ref{app:prereg:rev2delta}); a field whose near-lossless claims cannot
be rechecked, because per-item outputs are never released, is one whose
corrections cannot be seen either.

\section{The atlas: cancellation at scale, and what certification costs}

\label{sec:atlas}

\subsection{Construction, coverage, and the analysis population}
\label{sec:atlas:construction}

The atlas mines paired per-item records from public evaluation dumps at zero GPU
cost. Its protocol was frozen on 2026-07-15, before any flip statistic was
computed beyond two feasibility probes that the registration discloses by name,
and the pair list itself was frozen as a machine-readable manifest, with dataset
URLs, run timestamps and task lists, before any flip statistic existed.
Two sources are in scope: \textbf{S1}, the Open LLM Leaderboard v1 archive of
community quantizations of 2023-era base models, and \textbf{S2}, the Neural
Magic / Red~Hat per-item dumps for quantized Llama-3.1 at 8B, 70B and 405B
(Appendix~\ref{app:atlas:sources}). An item enters the paired analysis
only if its full-prompt hash is identical across the pair, and a cell is
excluded if fewer than 99\% of joinable items pass that check, because
prompt-hash identity, not harness version, is the operative control
(Appendix~\ref{app:atlas:pairing}).

\label{sec:atlas:coverage}%
The enumeration yields 2{,}055 pair-task cells, of which 1{,}807 are analysed
and 248 are excluded or skipped, most often because no results file exists for
that task in any recorded run (Appendix~\ref{app:atlas:exclusions} breaks the
exclusions down by reason). One of those reasons is a finding in itself. An
empty join intersection means the two sides of the pair were evaluated on
\emph{different item sets}: the leaderboard's baseline run and its quantized run
do not share the items whose scores are being compared. The resulting difference
in reported accuracy is not a paired quantity at all, no per-item analysis of it
is possible, and this is invisible to anyone reading only the aggregate
leaderboard numbers.

An earlier revision overstated the exclusion counts behind that observation,
reporting 643 float-scored and 132 empty-join exclusions and attributing both to
upstream reporting practice. A spot-check established that most of those cells
were readable and that our own parser could not see them; the figures above are
what survives once that defect is fixed (\S\ref{sec:prereg-spotcheck}).

Every statistic in the remainder of this section, in the certification tables of
\S\ref{sec:certification}, and in the discordance imputation of
\S\ref{sec:audit:rules} uses a single population: the \textbf{1{,}707} cells that
are neither excluded nor part of a disclosed feasibility probe
(S1 = 1{,}398, S2 = 309), spanning base models from 1.3B to 405B parameters. The
99 probe cells span the two pairs whose results were known before the
registration was written; the registration requires them in the atlas but not in
any headline aggregate, and they are tiny hand-built sanity pairs that would
distort any quartile (Appendix~\ref{app:atlas-detail}). The 1{,}807 figure above
is pipeline accounting and is never used as an analysis denominator.

\subsection{Net delta versus per-item churn}
\label{sec:atlas:netgross}

Across the atlas, per-item accuracy-state churn runs several times the net
accuracy delta. The ratio of median accuracy-state churn to median absolute
accuracy change is \textbf{5.40} pooled over the 1{,}707 cells
($0.120000 / 0.022222$), and 5.22 and 5.19 within S1 and S2. All three divide
unrounded medians; Table~\ref{tab:atlas-strata} prints the stratum medians to
enough places that a reader can reproduce the stratum figures from it. These
are ratios of medians, not medians of cellwise ratios, and the two
summarise different things. The median among finite cellwise ratios is
\textbf{3.85}. Assigning $+\infty$ to the 128 zero-delta/nonzero-churn cells
raises the all-cell median to 4.20. The finite ratios are the 1{,}562 cells
whose net delta is non-zero; a cellwise ratio is undefined for the remaining 145
(\S\ref{sec:atlas:identical}), and the exclusion is not evenly spread, removing
6.8\% of S1 and 16.2\% of S2. The pooled figure describes the typical churn
against the typical net delta; the cellwise median describes the typical cell.
Neither is the headline on its own.
The net delta a model card reports is the residue left after
harmful and beneficial flips cancel; the churn is the quantity that describes how
much of the model's behaviour actually changed. They are different quantities and
both should be reported. The consequence for inference is direct: churn is the
variance term in Equations~\eqref{eq:sds}--\eqref{eq:nreq}, so a compressed model
whose net delta looks reassuringly small can still be sitting on the noisiest
possible evidence base for the claim that it is unchanged.

The ratio is preserved across a generational change in method. The two strata
differ by nearly a factor of three in how much behaviour they disturb, with
median accuracy-state churn falling by a factor of 2.86 from S1 to S2, and by
almost exactly the same factor, 2.85, in the net delta they report
(Table~\ref{tab:atlas-strata}), so the understatement ratio barely moves: 5.22
in S1 and 5.19 in S2. Two generations of
compression method, one of them three times gentler than the other, hide the
disturbance they do cause by the same multiple. Better methods have made the
difference smaller \emph{without making the evidence for equivalence any more
sufficient}, because the quantity that determines how much evidence is required
scales down in lockstep with the quantity being claimed small.

\begin{table}[!t]
\centering
\caption{The two atlas strata, on the 1{,}707-cell analysis population. S1 is
the Open LLM Leaderboard v1 archive of community quantizations of 2023-era base
models, TheBloke-era GPTQ and similar; S2 is the Neural Magic / Red~Hat
W4A16, INT8 and FP8 releases from 8B to 405B. Percentages are within-stratum
shares of cells. The two medians are given to six places because the ratios
quoted in the text divide them unrounded; dividing the printed cells reproduces
5.22 and 5.19.}
\label{tab:atlas-strata}
\begin{tabular}{lrr}
\toprule
 & S1 & S2 \\
\midrule
Cells                                & 1{,}398        & 309 \\
Median accuracy-state churn          & 0.137452       & 0.048000 \\
Median $|$net accuracy delta$|$      & 0.026316       & 0.009242 \\
TOST-equivalent at 2\,pp             & 68 (4.9\%)     & 53 (17.2\%) \\
Exact McNemar $p < 0.05$             & 371 (26.5\%)   & 19 (6.1\%) \\
\bottomrule
\end{tabular}
\end{table}

Table~\ref{tab:atlas-strata} contains the empirical core of the paper's
motivation. In S1, 4.9\% of cells can be certified equivalent at the registered
2\,pp margin and 26.5\% show a detectable difference, which leaves
967 cells (69.2\%) of the public record in a \textbf{gray zone}:
neither certifiable as equivalent at the margin the field implicitly uses, nor
detectably degraded, at the sample sizes actually evaluated. In S2 the shares
move to 17.2\% certifiable and 6.1\% detectably different, and
240 cells (77.7\%) remain in the same gray zone.%
The S1$\to$S2 contrast is a genuine improvement in compression method and
evaluation practice, and should be read as one: a modern vendor quantization of
an 8B--405B instruction-tuned model perturbs per-item behaviour roughly a third
as much as a 2023 community quantization of a 7B base model, and its evaluations
are correspondingly more often certifiable. What does \emph{not} improve is the
evidential situation. The modal cell in \emph{both} strata is one where the
evaluation cannot answer the question the release note answers, for the reason
above: certifying a smaller margin requires \emph{more} items, not fewer.

\subsection{Identical scores, different answers}
\label{sec:atlas:identical}

The clearest single demonstration that aggregate accuracy is not a summary of
behaviour is the subset of cells where the aggregate does not move at all. Of
the 1{,}707 analysable cells, \textbf{145 (8.49\%) post an exactly identical
accuracy} to their baseline. Not similar: identical, to machine precision.
Among those 145 cells, the median accuracy-state churn is 0.0720, the
mean is 0.0887, and the maximum is 0.3434; 128 of the 145 have nonzero
churn, with a median of 0.0922 among that subset. About one in twelve compressed
model evaluations in the public record reports a score identical to its
baseline, and half of those still disagree with the baseline on more than 7\% of
individual items.

The mechanism is concrete in the most extreme such cell
(Table~\ref{tab:identical-extreme}, Appendix~\ref{app:atlas:extreme}): a 2023-era
community GPTQ of a 7B base model on one MMLU subject, where exactly 17.17\% of
items broke and exactly 17.17\% healed. The rates being equal, the net delta is
exactly zero and the exact McNemar test returns $p = 1.0$, correctly: there is no
evidence of a directional difference. Meanwhile a third of the answers changed. A
model card would truthfully write ``no change in accuracy'' for a cell where one
item in three behaves differently. The caveats belong in the same breath: at
$n = 198$ symmetric flip counts are easier to hit by chance, and this is an S1
cell from the noisier stratum. It illustrates the mechanism, not a typical
magnitude; the finding to carry away is the 145 cells and their 7.20\% median
churn, not this cell's 34\%.

\subsection{Population caveats}
\label{sec:atlas:caveats}

These caveats are registered, and they are load-bearing. S1 is community
quantizations of 2023-era models conditioned on leaderboard coverage: a pair
exists only if someone chose to submit both the base model and its quantization
to the Open LLM Leaderboard, and that choice was not random with respect to
anything. S2 is one vendor's releases, evaluated by that vendor.
\textbf{This is the public record of compression evaluation, not a census of
quantization.} It is the right population for what this paper asks (what the
circulating evidence looks like, and how much would suffice) and the wrong one
for how quantization behaves on average, which would need a designed sample
rather than the record left by the field's own reporting choices; the controlled
experiment of \S\ref{sec:minigrid} exists precisely because the atlas cannot
answer causal questions. Two further limits: S1's archive carries no
declared license, recorded as a limitation in the datasheet
(\S\ref{sec:artifacts}), and per-cell statistics inherit the item counts of
whoever ran the original evaluation, which is why many cells are small, an
inheritance that is itself part of the finding.

The findings above are claimed at the population each was measured on and no
wider. This section does not support any statement about quantization methods in
general, about models or methods absent from the two sources, or about causal
attribution of churn to any design choice. Nothing here is evidence for or
against H3.

\section{What published equivalence claims actually report}

\label{sec:audit}

\subsection{What was audited, and what was not}

The audit protocol was frozen on 2026-07-15, before any per-claim power
computation was run, and the claim list itself was frozen separately before any
verdict was computed.\footnote{%
Registration frozen at commit \texttt{b74fd58}; claim table frozen at commit
\texttt{715a7ce} with its sha256 recorded in the verdicts document.}
Sources were enumerated exhaustively within three fixed frames: method papers
(F1), official quantized model cards (F2), and inference-stack vendor blogs and
documentation (F3). A source enters the pool if it asserts, in prose or a table
caption, that a compressed model's benchmark quality is
equivalent-or-negligibly-different from its uncompressed baseline, under a
trigger vocabulary fixed in advance. Every claim meeting the criterion is
audited; there is no discretionary sub-selection.

The frozen table contains \AuditFrozenCandidates{} candidate claims: 7 method
papers, 7 official model cards or blogs, and 3 vendor documents. A full-text
review of every source, conducted after the first verdicts were computed, found
that one candidate's recorded quotation appears nowhere in its source, having
been composed from a table cell. The registered rule requires the assertion to
appear in prose or a table caption, so \AuditIneligibleClaim{} is excluded by
applying that rule as registered, leaving \textbf{\AuditEligible{}} eligible
sources; every count in this section is over those \AuditEligible{}, and the
exclusion lowers no count reported here
(Appendix~\ref{app:audit:eligibility}).

Each claim's fields were extracted by two blinded automated passes, separate
language-model agent sessions with the second given no access to the first
pass's output, followed by human reconciliation of every disagreement before the
verdict stage (\S\ref{sec:prereg}). Separate sessions prevent the second pass
from copying the first; they do \emph{not} make the two passes statistically
independent, because both share a common model prior, and nothing here is
inter-rater agreement. Appendix~\ref{app:extraction} specifies the procedure and
reports what can and cannot be measured after the fact.

\textbf{No audited claim is described as false, and none of our findings implies
that any audited model is in fact degraded.} The audited property is the
\emph{evidential sufficiency of the reported evaluation}: whether the evaluation
offered in support of an equivalence claim was large enough to have detected the
difference the claim pronounces negligible. A claim can be perfectly correct and
still be unsupported by the evaluation offered for it, and several of the claims
below are very probably correct. Nor is any proportion reported here a
prevalence estimate for the field. What is missing from the audited sources is a
reporting standard: this section measures the gap such a standard would close,
and \S\ref{sec:certification} supplies the instrument.

\subsection{What the sources declare}
\label{sec:audit:taxonomy}

%
%
Across the \AuditEligible{} eligible sources we distinguish two independent
properties: what form the source's evidence takes, and whether the reported
information suffices to assess it numerically at all.

\begin{table}[!t]
\centering
\caption{What the 16 eligible sources declare. Columns are the form of the
evidence offered; rows are whether the source reports enough for a numerical
assessment under the registered paired-outcome framework. \emph{Prospective
numerical decision margin} means a tolerance stated independently of the
observed result: a threshold that could have been written down before the
evaluation ran. That column is empty.}
\label{tab:audit-taxonomy}
\setlength{\tabcolsep}{4pt}
\begin{tabular}{l>{\centering\arraybackslash}p{2.2cm}>{\centering\arraybackslash}p{2.4cm}>{\centering\arraybackslash}p{2.2cm}}
\toprule
 & Prospective numerical decision margin & Retrospective numerical description of a result & Qualitative language only, no number \\
\midrule
Numerically assessable     & 0 & 5 & 7 \\
Not numerically assessable & 0 & 1 & 3 \\
\midrule
\textbf{Total}             & \textbf{0} & \textbf{6} & \textbf{10} \\
\bottomrule
\end{tabular}
\end{table}

\textbf{No audited source declares a prospective numerical equivalence margin}
(Table~\ref{tab:audit-taxonomy}). \AuditXtabQualTotal{} of the
\AuditEligible{} make the claim in purely qualitative terms (``negligible'',
``matches'', ``practically no accuracy decrease''), with no number attached to
the assertion at all. The remaining \AuditXtabRetroTotal{} cite a number, and in
every case it is a \emph{measured outcome}: a recovery percentage or an achieved
benchmark score, computed after the evaluation rather than set before it. This
is established over complete source text, not the quoted sentence: every source
was archived and searched in full, including tables, captions, footnotes and
appendices, for a registered vocabulary of tolerance language
(Appendix~\ref{app:audit:sweep}).

The consequence for this section is structural. An equivalence claim with no
declared margin cannot be evaluated against its own standard, because it has
set none; it can only be evaluated against a standard supplied from outside.
Everything below therefore uses the margin registered in advance
(\S\ref{sec:audit:rules}), and no quantity derived from a source's own reported
results is described anywhere in this paper as that source's margin.

\subsection{Where the claim is written}
\label{sec:audit:locus}

A second question turns out to matter as much as what form an assertion takes:
\emph{where in the document it is written}, and whether it is written in words at
all. Across six model cards from one publisher with the underlying evidence held
constant, the claim appears in prose in three, as a bare juxtaposition of two
scores in two, and only as a table cell in one
(Table~\ref{tab:audit-locus}, Appendix~\ref{app:audit:locus}). The registered
inclusion rule is keyed to a property that varies independently of the evidence,
so the claims do not become weaker as the locus moves from sentence to cell:
they stop being written down. The consequence is that
\AuditFrozenCandidates{} is a floor on the population and not a census of it,
with the shortfall concentrated in exactly the sources that state their claims
least explicitly. This bounds what the counts here can mean without weakening
them, since every reported quantity is computed over the sources that were
captured, and it means the true rate of unbounded equivalence claims in this
literature is understated here rather than overstated.

The locus review is author re-verification against the archived sources, not
independent verification by a second coder; the project has no second human, and
Appendix~\ref{app:extraction} states what that does and does not license. Every
archived source, its content hash, and the retrieval script needed to reproduce
these locations are released with the paper, so the classification is checkable
by any reader without re-fetching anything.

\subsection{Availability of per-item outputs}
\label{sec:audit:v3}

The most actionable finding requires no statistics at all.

%
\begin{quote}
\textbf{\AuditPerItemTaskMatched{} of the \AuditEligible{} eligible sources
release \emph{task-matched} per-item outputs}, meaning the item-level results,
for the tasks the equivalence claim is actually about, that a third party would
need to rerun the paired comparison. The tally is \AuditPerItemTaskMatched{}
\emph{yes}, \AuditPerItemOtherTaskOnly{} \emph{partial}
(\AuditPerItemOtherTaskClaims{}), \AuditPerItemNone{} \emph{no}.
\end{quote}

\AuditPerItemOtherTaskClaims{} come closest: they release per-item outputs for
other suites, but not for the tasks their audited equivalence claim is about, so
the released artifacts and the asserted claim do not intersect
(Appendix~\ref{app:audit:v3detail}). This finding is more actionable than any
power calculation because the two failures have different repair costs.
Underpowering is fixable by evaluating more items, and \S\ref{sec:certification}
says how many. Irreproducibility is not fixable downstream at all: with no
per-item outputs, nobody outside the releasing organisation can run the paired
test at \emph{any} sample size, cannot compute churn, and cannot check the
arithmetic. Per-item outputs for a benchmark of a few thousand items are a file
of a few megabytes, and the gap between that cost and its consequence is the
single clearest reporting-standards recommendation this paper makes.

\subsection{Verdict rules}
\label{sec:audit:rules}

Because no source declares a margin, \textbf{the applicable margin throughout is
the uniform \AuditMarginPP{}\,pp registered in advance}; a source's reported
deltas are outcomes of its evaluation, not a decision rule it adopted, and are
never used as one.

For each claim, at its reported or rule-imputed sample size, the frozen protocol
computes three quantities: \emph{V1}, the minimum detectable difference under
the paired-flip model; \emph{V2}, the number of items TOST would require at the
applicable margin; and \emph{V3}, whether per-item outputs were released. A
claim is recorded as \emph{below the approximate planning threshold} iff its
reported $n$ falls below the V2 requirement.
Appendix~\ref{app:audit:verdictrules} gives the estimators and the one-sided
$z$ convention, under which the requirement is that a 90\% two-sided interval
fall inside the margin.
That requirement answers a question asked \emph{before} an evaluation is run,
namely how many items it would need, so falling below it is a statement about
how the evaluation was sized, not a diagnosis applied to its result.

The discordance rate $p_d$ is reported by no source, so it is imputed from the
atlas (\S\ref{sec:atlas}) by matching the nearest (method family, bit width,
benchmark) cell, most-specific tier first, taking the \emph{median} over the
first non-empty tier because per-cell discordance is right-skewed
(Appendix~\ref{app:audit:imputation}). Because $p_d$ is imputed and never
reported, each verdict is additionally recomputed at the first and third
quartiles of the same atlas cells that supplied its median.
This quartile sweep was added after the point-imputation results had been seen,
and is reported as post-hoc sensitivity, not as a registered analysis.
Because $n_{\mathrm{req}}$ is increasing in $p_d$, the two quartiles bracket the
whole interval: a claim above threshold at $Q_3$ is above it throughout, and one
below threshold at $Q_1$ is below it throughout. The interval statement is
established at its endpoints, not inferred from the scatter of observed cells.

\subsection{Results}
\label{sec:audit:results}

Of the \AuditEligible{} eligible claims, \AuditNotAssessable{} are \textbf{not
numerically assessable}: their reporting does not support a verdict under the
registered framework, leaving \AuditAssessable{} assessable claims. Of those
\AuditAssessable{}, \AuditBelowThresholdAtMedian{} falls below the
approximate planning threshold at the registered \AuditMarginPP{}\,pp margin
under the median discordance imputation, and that one claim's classification
does not survive the sensitivity analysis.

\begin{table}[!t]
\centering
\caption{Sensitivity classification of the 11 numerically assessable claims at
the registered 2\,pp margin. Each claim's required sample size is recomputed at
the first and third quartiles of the atlas cells that supplied its median
discordance; because required $n$ increases monotonically in the discordance
rate, the quartiles bracket the whole interval. Per-claim values are in
Appendix~\ref{app:audit-table}.}
\label{tab:audit-sensitivity}
\begin{tabular}{lr}
\toprule
Classification across the atlas-IQR interval & Claims \\
\midrule
Above the planning threshold throughout          & \AuditAboveThroughout{} \\
Changes classification within the interval       & \AuditChangesWithinIQR{} \\
Below the planning threshold throughout          & \AuditBelowThroughout{} \\
\midrule
\textbf{Total assessable}                        & \textbf{\AuditAssessable{}} \\
\bottomrule
\end{tabular}
\end{table}

The single flagged claim is \AuditSensitiveClaim{}, the GPTQ paper, which
reports $n = \AuditSensitiveN{}$ against a requirement of
$\AuditSensitiveNReq{}$ items at an imputed discordance rate of $0.130$. Its
classification reverses at a rate of $0.1189$, and
$\AuditSensitiveCellsBelow{}$ of the $\AuditSensitiveCellsTotal{}$ atlas cells
supplying the imputation lie below that point
(Appendix~\ref{app:audit:r01}). This is a sensitivity-dependent planning flag,
not a verdict: the $\AuditSensitiveCellsPct{}\%$ is a descriptive share of
reference cells, not a probability that the claim is underpowered, and those
cells share models, benchmarks and infrastructure and are not independent
observations.

No assessable claim falls below the threshold throughout the interval.
Table~\ref{tab:audit-sensitivity} is therefore best read as a statement about
resolution, not about error: at the margin a reporting standard would plausibly
impose, ten of these eleven evaluations were sized to carry the claim made on
them, and the eleventh cannot be classified without committing to a discordance
rate its authors never reported.

The same conclusion appears in the detection direction, where the
independent-binomial columns are uniformly worse by roughly a factor of two
(Table~\ref{tab:audit-mdd}, Appendix~\ref{app:audit:fullrobustness}); an audit at
a \AuditMarginPP{}\,pp margin is a weak test and should be understood as one.
Second, \textbf{R04 and R14 carry no verdict}; their numbers are computable and
are retained in the released CSV for transparency, but they are excluded from
the assessable set (\S\ref{sec:audit:indeterminate}), and the appendix table
italicises them for that reason. Finally, \AuditSensitiveClaim{} is unstable in
two independent directions that must not be read as one finding: it is
\emph{margin-sensitive}, flipping across the registered
1\,pp\,$\to$\,3\,pp sweep, so it turns on where the yardstick is set, and
separately \emph{imputation-sensitive}, flipping across the interquartile range
of the atlas cells supplying its discordance rate, so it also turns on a
quantity its source never reported. No other assessable claim is sensitive in
either direction (Appendix~\ref{app:audit:robustness}).

\subsection{Claims that cannot be assessed}
\label{sec:audit:indeterminate}

\AuditNotAssessable{} of the \AuditEligible{} eligible claims cannot be assessed
numerically, in two kinds, each recorded with exactly one primary blocker.
\AuditNotAssessableInsufficient{} are cases of \emph{insufficient reporting},
where a registered input is genuinely absent: no sample size, no baseline, or
headline equivalence evidence existing only as a chart image with no extractable
numbers. The remaining \AuditNotAssessableOutsideFramework{},
\AuditOutsideFrameworkClaim{}, is \emph{outside the registered calculation}: it
reports enough, but about a quantity our registered \emph{binary paired-outcome}
model cannot score (\S\ref{sec:audit:r04}). The per-claim blockers are in
Appendix~\ref{app:audit:indeterminate}. Every non-assessable claim retains
whatever components its available inputs support, reported as supplementary
transparency only and \textbf{never verdict-bearing}.

R14 deserves a word, because its numbers look assessable and are not. Its
imputed sample size of $728$ falls just short of the $742$ the 2\,pp calculation
would require, close enough that a reader might suspect it was set aside for
being inconvenient. It is set aside because the missing baseline is a registered
input: without it there is no paired comparison to size, and the $742$ is a
requirement for a calculation that cannot be performed. The proximity is a
coincidence with no bearing on the count, and R14 does not enter the threshold
tally in either direction.

This category is itself a result. Two of the four insufficiently-reported claims
are among the most-cited results in the field, and their headline equivalence
evidence is a chart image with no extractable numbers. That a mechanical,
pre-registered audit cannot evaluate them is precisely the reporting-standards
problem this paper exists to address: the claim may well be true, but it has been
placed beyond the reach of checking.

\subsection{R04: outside the registered calculation}
\label{sec:audit:r04}

R04 is recorded because excluding it removed what was, at the time, the audit's
largest single number. The AWQ paper's qualifying sentence, the one that meets
the frozen §3 inclusion trigger, asserts negligible loss on \emph{COCO
CIDEr}, a generation metric that assigns a graded score to a caption and has no
per-item correct/incorrect state. V1 and V2 are flip-model quantities defined on
$d_i \in \{-1,0,+1\}$, and there is no discordance rate to impute because there
is no per-item accuracy state to be discordant about. This is a limitation of
\emph{our registered calculation}, not of paired analysis: a graded metric
supports paired resampling on the same items perfectly well, and a certificate
for it would be a straightforward extension. It is simply not the model we
registered.

Excluding it was not free. The first extraction pass had scored R04 on GSM8K,
the source's own accuracy benchmark, and reported it as the audit's largest
shortfall; that reading was overruled because computing a TOST requirement on
GSM8K audits a sentence the source wrote about a different benchmark, in
different and non-trigger language. The computation is retained in the released
CSV as a labelled transparency column, and \textbf{it is not claimed anywhere in
this paper as an audit result} (Appendix~\ref{app:audit:r04detail}). An audit's
currency is unimpeachability, and it is not spent on its own largest number.

The count of claims below the planning threshold moved four times before it
settled, and the order in which the numbers arrived is itself part of the
disclosure. A first pass applying the registered \AuditMarginPP{}\,pp margin
uniformly returned $K = 1$ of 12. Re-reading the frozen label produced $K = 5$;
the R04 ruling above moved one claim out of the determinate set, giving the
$K = 4$ that was reported for ten days. Amendment~2 then returned the analysis
to the uniform registered margin, where the rev-3 recomputation gives
\textbf{\AuditBelowThresholdAtMedian{} of \AuditAssessable{}} assessable claims
below the threshold, the same claim the very first pass had flagged. The
claim-derived margins and the shortfall range computed under the superseded
reading are \textbf{withdrawn, not recomputed}, and are non-verdict-bearing
wherever they still appear as history
(Appendices~\ref{app:audit:history} and~\ref{app:prereg:choices}).

The findings above hold at the reported sample sizes and under the frozen
protocol as amended, and they are claims about evidence, not about models. This section \textbf{does not establish}: that any audited model is
degraded; that any audited claim is false; that the claims' authors reached a
wrong conclusion; or that any claim is underpowered in a sense that survives the
sensitivity analysis, and none is.
Nor does it establish a prevalence: \AuditEligible{} eligible sources within
three frozen frames are not a sample from which the field's reporting practice
can be estimated, and no proportion reported here should be read as one. Every
numerical result here also rests on a discordance rate that no source reported
and that is imputed from the atlas, which is why each verdict is recomputed
across the interquartile range of the cells supplying it and why the one flagged
claim is reported as sensitivity-dependent rather than as a finding. Where a
classification is called robust, that means only that it holds throughout the
interquartile range of the atlas cells supplying its discordance rate, not
across every plausible model of discordance. It also does not
establish anything about sources outside the three frozen frames, and it
operates at claim level and not at claim\,$\times$\,benchmark granularity,
because that is the granularity of the frozen claim table
(Appendix~\ref{app:prereg:choices}, interpretive choice~4).

\section{Cancellation is worse where practitioners choose}
\label{sec:minigrid}

\subsection{Design}

The registered confirmatory design compares 4-bit GPTQ against 4-bit AWQ at
five calibration seeds $\{0,1,2,3,4\}$ over eight model-by-benchmark cells:
$\{$Qwen2.5-1.5B-Instruct, Qwen2.5-7B-Instruct, Llama-3.2-3B-Instruct,
Llama-3.1-8B-Instruct$\}\times\{$MMLU, GSM8K$\}$, with MMLU as the registered
likelihood-based benchmark and GSM8K as the registered generative one.

The pairing is the design's load-bearing element. For seed $s$ the calibration
set is 128 samples of exactly 2{,}048 tokens drawn from C4 by a fixed,
seed-determined procedure, and \textbf{GPTQ seed $s$ and AWQ seed $s$ receive
the identical ordered calibration samples}; selected document indices and token
hashes are persisted (Appendix~\ref{app:escalation}). A ranking difference at
seed $s$ is therefore attributable to method-by-calibration interaction and not
to the two methods having seen different data.

Per cell, with $d_s = \mathrm{acc}_{\mathrm{GPTQ},s} - \mathrm{acc}_{\mathrm{AWQ},s}$:
a \emph{winner flip} occurs when two seeds give $d_s$ and $d_t$ of opposite sign
with both nonzero, exact ties counting as neither and reported separately;
$\mathrm{gap}$ is the absolute difference of the two five-seed mean accuracies;
$\mathrm{range}_m$ is the spread of method $m$ across seeds; and the
\emph{range/gap} criterion holds when
$\max(\mathrm{range}_{\mathrm{GPTQ}}, \mathrm{range}_{\mathrm{AWQ}}) \geq
\mathrm{gap}$. Seed-level SD and item-level SE are reported as separate variance
components rather than collapsed, as the registration requires.

The eight cells were built in two stages, and the distinction between the two
matters exactly once. Four cells ran first; the 7B and 8B cells sat behind a
mechanical escalation screen frozen before either stage ran
(Appendix~\ref{app:escalation}). \textbf{The screen decides only which cells to
build and states no H3 outcome; the registered
Supported/Disconfirmed/Inconclusive rule is defined over all eight cells and was
applied once, after all eight existed.} The confirmatory rule was therefore
never applied to a cell set chosen after its own result was known, and no
reduced-cell variant was constructed at any point.

\subsection{The eight-cell verdict}
\label{sec:minigrid:verdict}

Applied once, mechanically, to all eight cells:
\textbf{H3 is supported.} The winner reverses across seeds in \textbf{5 of the 8}
cells against a threshold of 3, and the range/gap criterion holds in
7 of the 8 against a threshold of 4. The supported limb of the frozen
rule is a disjunction, and \emph{both disjuncts are satisfied independently}:
either one alone would have returned the same verdict. The disconfirming limb is
a conjunction, and both of its conjuncts fail. It requires winner flips in at
most 1 of 8, and there are 5, and it requires
$\max\mathrm{range} < 0.5\,\mathrm{gap}$ in at least 6 of 8, which holds in 1.
The two limbs cannot both hold and do not, so the frozen text classifies the
outcome with no interpretation required.

No exact accuracy tie occurs anywhere in the confirmatory set: 0 of 8
cells contain one, over 0 of 40 (model, task, seed) triples. The
registered tie convention, under which a tie is neither a flip nor a non-flip
and can neither create nor erase a flip between two non-tied seeds, therefore has no
effect on any cell's classification, and the denominator for both criteria
remains all eight cells.

What the verdict says is bounded by what was registered. Over the eight cells,
the choice between GPTQ and AWQ at 4 bits is not stable against
calibration-seed randomness alone. It licenses no statement about 3-bit
behaviour, ARC-Challenge, HellaSwag, calibration-\emph{dataset} effects, or any
cell outside the registered set; the FP16 baseline gate governs the baseline
only and says nothing about quantized accuracy.

Table~\ref{tab:h3-eightcell} gives the two rule inputs per cell, and
Table~\ref{tab:h3-ds} in Appendix~\ref{app:minigrid:ds} gives the per-seed
differences $d_s$ behind the winner-flip column, so every flip determination is
checkable by inspection of sign.

\begin{table}[!t]
\centering
\small
\caption{The eight registered confirmatory cells. $\mathrm{gap}$ is the absolute
difference of the two five-seed mean accuracies; $\mathrm{range}_m$ is the
seed-wise spread of method $m$; the range/gap criterion holds when
$\max(\mathrm{range}_{\mathrm{GPTQ}},\mathrm{range}_{\mathrm{AWQ}}) \geq
\mathrm{gap}$. Values reproduce the signed decision record.}
\label{tab:h3-eightcell}
\setlength{\tabcolsep}{3pt}
\small
\begin{tabular}{llrrrll}
\toprule
Model & Task & $\mathrm{gap}$ & $\mathrm{range}_{\mathrm{GPTQ}}$ & $\mathrm{range}_{\mathrm{AWQ}}$ & Flip & Range/gap \\
\midrule
Qwen2.5-1.5B & MMLU  & 0.012292 & 0.040521 & 0.018445 & \textbf{TRUE}  & \textbf{TRUE} \\
Qwen2.5-1.5B & GSM8K & 0.096800 & 0.014000 & 0.033000 & FALSE          & FALSE \\
Llama-3.2-3B & MMLU  & 0.030922 & 0.063809 & 0.029341 & FALSE          & \textbf{TRUE} \\
Llama-3.2-3B & GSM8K & 0.017800 & 0.011000 & 0.034000 & \textbf{TRUE}  & \textbf{TRUE} \\
Qwen2.5-7B   & MMLU  & 0.006267 & 0.015667 & 0.012391 & FALSE          & \textbf{TRUE} \\
Qwen2.5-7B   & GSM8K & 0.005800 & 0.048000 & 0.035000 & \textbf{TRUE}  & \textbf{TRUE} \\
Llama-3.1-8B & MMLU  & 0.017163 & 0.040023 & 0.025495 & \textbf{TRUE}  & \textbf{TRUE} \\
Llama-3.1-8B & GSM8K & 0.013200 & 0.033000 & 0.026000 & \textbf{TRUE}  & \textbf{TRUE} \\
\midrule
\multicolumn{7}{l}{Winner flip in 5 of 8 (threshold $\geq 3$: met)} \\
\multicolumn{7}{l}{Range/gap in 7 of 8 (threshold $\geq 4$: met)} \\
\multicolumn{7}{l}{$\max\mathrm{range} < 0.5\,\mathrm{gap}$ in 1 of 8 (disconfirm threshold $\geq 6$: not met)} \\
\multicolumn{7}{l}{Exact ties: 0 of 40} \\
\bottomrule
\end{tabular}
\end{table}

\subsection{Supporting analyses}
\label{sec:minigrid:supporting}

Three further registered analyses were run once per cell in the same job that
produced the verdict, and none of them modifies it: the verdict is what the
frozen rule returned over winner flips and range/gap, and a weak bootstrap rate
or a poorly resolved cell is a limitation to report, not a second application of
the rule. Table~\ref{tab:h3-supporting} in
Appendix~\ref{app:minigrid:supporting} carries all three.

Two of them matter for how the verdict should be read. First, on MMLU the
calibration seed moves accuracy more than the sample of items does in every
cell, with the larger seed-level SD running $1.6$ to $5.9\times$ its own
item-level SE, whereas on GSM8K the two components are of the same order
throughout. Second, the bootstrap and the winner-flip criterion measure
different things and a reader who assumes otherwise will read them as
self-contradictory: a winner flip asks whether two \emph{individual seeds}
disagree on sign, which is H3's registered question, while the bootstrap
rank-flip rate asks whether the \emph{five-seed mean} ranking survives
resampling seed labels and items together. Winner flips occur in 5 of 8 cells,
the bootstrap rate is below $0.05$ in 6 of 8, and \textbf{3 cells have both}. In
those three, some pair of individual seeds disagrees about which method wins
while the five-seed average survives resampling in more than 95\% of
replicates. The practical reading is that \emph{a single calibration run can
hand you the wrong winner}, and averaging over the registered five seeds is
substantially more stable than any one of them.

The limitation on that reading, stated plainly. With five seeds the bootstrap
resamples seed labels from a sample of five, so it has limited resolution on the
seed-level distribution. What the low rates bound is the stability of
\emph{this} observed mean under resampling; they do not establish that five
seeds suffice in general, and no result here licenses five as a sufficient
number. Two GSM8K cells are unstable on both measures, and there even the
five-seed mean ranking does not survive resampling.

\subsection{Cancellation is more complete between two methods than against FP16}
\label{sec:minigrid:churnratio}

\begin{result}\label{res:churnratio}
\emph{Descriptive; not a registered hypothesis and not an inferential test.}
The ratio of per-item churn to aggregate net delta is larger in the
controlled method-against-method contrast than in the observational
quantized-against-FP16 contrast. The ordering is stable across the two
aggregations, which is a robustness check on the choice of summary
(Table~\ref{tab:churn-aggregations}).
Comparing like with like, the
median of the per-cell ratios is \textbf{3.85} across the atlas
(\S\ref{sec:atlas:netgross}) and \textbf{12.7} across the eight controlled
cells, where the per-cell ratios run \textbf{$3.0\times$ to $30.5\times$};
comparing medians instead, the atlas gives \textbf{5.40} against
\textbf{12.1}.
\end{result}

\begin{table}[!t]
\centering
\caption{The churn-to-net-delta ratio under both aggregations, in both regimes.
The two aggregations answer different questions, so the paper reports both
rather than choosing: a ratio of medians describes the typical churn against the
typical net delta, and a median of per-cell ratios describes the typical cell.
The controlled regime exceeds the observational one either way. All figures here
are descriptive summaries of the two populations; the comparison is not a
registered hypothesis and no inferential test is performed on it. The last row is
why the two columns are not exactly parallel: a cellwise ratio is undefined
where the net delta is exactly zero, which never happens in the eight controlled
cells and happens in 145 atlas cells, 6.8\% of S1 and 16.2\% of S2
(\S\ref{sec:atlas:identical}).}
\label{tab:churn-aggregations}
\begin{tabular}{lrr}
\toprule
 & Atlas (observational) & Controlled \\
\midrule
Ratio of medians                    & 5.40            & 12.1 \\
Median of per-cell ratios           & 3.85            & 12.7 \\
Cells entering the per-cell median  & 1{,}562 of 1{,}707 & 8 of 8 \\
\bottomrule
\end{tabular}
\end{table}

The extreme cell makes the size of the discrepancy concrete. On
Qwen2.5-7B/GSM8K the two methods differ by 0.58\,pp in aggregate
accuracy, a gap any reader would call equivalence, while 17.7\% of
items change correctness state between them and 28.7\% of answers
change outright. The mechanism is cancellation, and it follows from what the two
contrasts are. Two quantization methods at the same bit width are far closer to
each other in aggregate than either is to its FP16 baseline. Harmful and
beneficial flips are correspondingly better balanced, cancellation in the
aggregate is more complete, and the surviving net delta therefore hides a
proportionally larger amount of per-item movement.

We observe two regimes; we do not establish a curve. These are two
measured points, one observational contrast at 1{,}707 cells and one controlled
contrast at 8, and nothing here licenses reading the ratio as a monotone
function of aggregate closeness. What the two points support is the direction of
the effect and the fact that the regime in which equivalence claims are actually
made is the less favourable of the two.

\subsection{Which arm of the experiment carries the result}
\label{sec:minigrid:resolution}

\textbf{This quantity was not registered.} It was requested on 2026-07-26,
\emph{after} the eight-cell verdict had been computed and signed, in response to
prior art identified the same day,%
\ \citet{paglieri2024outliers}, which reports calibration effects diminishing in
modern LLMs (Appendix~\ref{app:reconcile}). It is descriptive, it tests no
hypothesis, and it does not modify the verdict. It is recorded with its
provenance, not folded silently into the registered results.

Using the machinery of \S\ref{sec:certification} itself, with paired
$\mathrm{sd} = \sqrt{p_d}$ and $p_d$ the per-cell GPTQ-against-AWQ
accuracy-state churn, the two tasks separate sharply
(Table~\ref{tab:h3-resolution}, Appendix~\ref{app:minigrid:resolution}).
\textbf{MMLU carries the result}: all four MMLU cells satisfy the range/gap
criterion, their seed-induced ranges run $5.5$ to $17.5$ paired standard errors,
and the instability is far larger than the measurement noise.
\textbf{GSM8K is under-resolved at $n = 1{,}000$}: there the ranges run roughly
two to three-and-a-half SE, and in three of the four GSM8K cells the \emph{mean
gap itself} sits at or below $1.25$ SE, so the quantity the range/gap criterion
compares against is, in those cells, not resolved by the benchmark at the $n$
used. Any per-cell reading of GSM8K should say so, and \S\ref{sec:limitations}
does.

The split is also the answer to an obvious objection, that a certification-first
framework demanding thousands of items will simply decline to conclude anything.
The fourth GSM8K cell is the sharp case: Qwen2.5-1.5B posts a GPTQ--AWQ gap of
\textbf{9.68\,pp}, which lands at $5.65$ paired SE, resolved on the same
benchmark, at the same $n$, under the same test, where its three siblings are
not. Same items, same instrument, opposite verdicts, separated by effect size
and not by any threshold somebody chose. A framework that returned
``insufficient evidence'' everywhere could not produce that pattern, and neither
could one tuned to a convenient cut-off.

None of this weakens the verdict. The range/gap criterion holds in 7 of 8 cells
and winner flips occur in 5 of 8, both computed from the registered per-seed
accuracies by the frozen rule, and neither is a significance test. What it means
is that the GSM8K cells carry materially less resolving power than the MMLU
cells, which is a statement about where the evidence concentrates, not a
reweighting of the rule.

\subsection{Resolution of the confirmatory cells against the certification tables}
\label{sec:minigrid:selfaudit}

The certification table of \S\ref{sec:certification} requires, for GSM8K at a
2\,pp margin, about \textbf{1{,}184} items at median discordance and
\textbf{3{,}068} at the p75 discordance. The confirmatory GSM8K cells ran at
$n = 1{,}000$, already below the median requirement, and the discordance
actually observed in them is $p_d = 0.1766$, $0.1808$, $0.2086$ and $0.2940$,
at or above the table's p75, where the requirement is roughly three times the
$n$ used. The certification table therefore predicted this shortfall in advance,
and the same instrument applies to this experiment as to the audited claims.
One caveat bounds the comparison. The table's discordance percentiles describe a
quantized-against-FP16 contrast, while $p_d$ here is GPTQ-against-AWQ. The
required-$n$ column is a function of discordance whatever the pair, so reading
the observed $p_d$ against the table's brackets is legitimate; the percentile
labels are not strictly like-for-like and are used only to locate the magnitude.

\section{Exploratory scoring-pipeline sensitivity}
\label{sec:sensitivity}

The certification apparatus of \S\ref{sec:certification} treats the compressed
model as the source of per-item churn and the evaluation as a fixed instrument.
This section asks the complementary question and finds the instrument is not
fixed: \emph{how much does evaluation-harness configuration move accuracy and
per-item answers on a single, unchanged model, relative to how much quantization
moves them under a fixed configuration?}
It is a small, preregistered, \emph{exploratory} study on one model and the bridge
item subsets, frozen before any of its own results existed. The motivation is
not hypothetical: this campaign produced two live configuration defects inside
eight days on the pinned \texttt{lm\_eval}~0.4.12, one of which moved reported
GSM8K accuracy $0.232 \to 0.566$ with not one token of model output changed
(Appendix~\ref{app:harness-detail}).

The headline statistic, fixed before any run, is
$R_{\mathrm{cond}} = C_{\mathrm{cond}}/\bar{Q}$: churn between the reference
configuration and a condition on the fixed FP16 model, over the mean churn of
the ten Qwen2.5-1.5B quantized variants against that model's FP16 cell on the
same items, giving $\bar{Q}(\text{MMLU}) = 0.199$ and
$\bar{Q}(\text{GSM8K}) = 0.287$. So $R$ asks directly how a
configuration change compares with swapping in a quantized model. The protocol
forbids printing the ratio without both inputs, and both appear in every row of
Table~\ref{tab:sensitivity}.

Across the five conditions the ratio runs $R \in [0.836,\,1.585]$, so on a
fixed model a configuration change moves per-item correctness by as much as, or
more than, swapping in a quantized variant does. The condition-by-condition
table, the directional splits, and the MMLU cell whose net delta is positive
while its churn is on the quantization scale are in
Appendix~\ref{app:sensitivity:conditions}.

\paragraph{Scope.}
On a fixed FP16 model and matched items, harness configuration moves per-item
correctness by an amount comparable to quantization, with $R \in [0.836,\,1.585]$
across the five conditions and the scoring-filter choice ($R=1.585$) exceeding
the quantization denominator outright, so a compression comparison run under an
unstated or mismatched configuration can be dominated by the configuration.
These caveats travel with the numbers. This is \textbf{one model}
(Qwen2.5-1.5B-Instruct); the Llama-3.2-3B arm is admitted only after its seed-0
canary passes, which has not run, so no Llama $R$ is reported. The items are the
bridge subsets ($n=400$ MMLU, $n=200$ GSM8K), chosen so the study
touches no confirmatory item definition. And the study is
\textbf{exploratory}: it adjusts no gate, no escalation rule, and no
confirmatory analysis, and states no H3 outcome. It contextualises how a
compression score should be read; it decides nothing about the compression
methods themselves.

\section{Artifacts}
\label{sec:artifacts}

Six artifacts are released, the first being the one this paper's own audit found
nobody publishes: the per-item outputs, 88 cell JSONL files spanning every cell
of the controlled experiment.
Section~\ref{sec:audit:v3} reports that \AuditPerItemTaskMatched{} of the
\AuditEligible{} eligible sources release \emph{task-matched} per-item outputs,
and Section~\ref{sec:conclusion} asks the field to close that gap. The other
five are the flip atlas, the \texttt{flipeval} package (Apache-2.0; everything
else CC-BY-4.0), the frozen 17-claim audit table and its verdict CSV, the
certification tables, and the reproduction package.

The audit reads 17 sources in full, and \textbf{their full-text captures are in
no release}: a redistribution review conducted on 2026-08-02 found four of them
carrying no grant permitting a third party to republish their text, and the
seven method papers under arXiv's default licence, which authorises arXiv rather
than us, so the working archive is kept private. What ships in its place is each
source's URL and pinned version identifier, a SHA-256 per captured file, the
manifest and a retrieval script (Appendix~\ref{app:artifacts:released}).

%
%
The archived release is canonical: \textbf{\versiondoi}, which resolves to the
frozen v1.2.0 state this paper describes and not to the latest one. The package
is at \repourl, tag \texttt{v1.2.0}, and the data is mirrored at \dataseturl.
\textbf{v1.0.0 (\priorversiondoi) is historical}: it predates the rev-3 audit
verdicts reported here, carries the superseded rev-2 file, and stays published
rather than withdrawn so that the correction remains visible.
Appendix~\ref{app:artifacts-detail} describes every artifact in full, with the
datasheet, licensing findings, metadata and maintenance statement.

\section{Limitations}
\label{sec:limitations}

The atlas is a record, not a sample. S1 is community quantizations of 2023-era
models, conditioned on leaderboard coverage, and S2 is one vendor's releases
evaluated by that vendor, so statistics computed over the atlas describe the
evidence the field circulates and not the behaviour of quantization in general.
Its pipeline had two bugs found and fixed during the run, and its numbers were
held provisional pending an independent spot-check; that check has since been
completed twice, with 262 compared fields reconciled on the first pass, which
surfaced the population defect that produced rev-2, and 14 of 14 cells at 126 of
126 fields on the targeted second (\S\ref{sec:prereg-spotcheck}), so the figures
reported here are rev-2 and are no longer provisional, and the delta between the
two revisions is reported (Appendix~\ref{app:prereg:rev2delta}). The audit then
leans on that record: no audited source reports a per-item disagreement rate, so
the audit imputes one from the median of the nearest non-empty atlas tier, and
the tiers actually reached run from the most specific, matching model family,
bit width and benchmark together, down to bit width alone. A claim matched at a
coarse tier rests on a wider and less similar set of cells, so the match tier
and the number of atlas cells behind it are released per claim, the
independent-binomial robustness column bounds the effect of the imputation in
the conservative direction, and Appendix~\ref{app:audit:imputation} names the
weakest such match in the table explicitly. In the certification tables,
per-subject and per-subtask cells are collapsed into families, so a family's
quartile band mixes subject-level with model-level variation; this makes the
p25--p75 columns conservative, and two families rest on fewer than 12 cells and
are indicative only. Audit verdicts are computed per claim, at the pooled $n$
the frozen table records, because operating at a claim\,$\times$\,benchmark
granularity would require inventing rows the freeze does not contain. The three
frames fix the population: sources outside them are not audited, and claims
discovered after the table freeze would form a separately reported post-freeze
stratum. Every proportion reported here describes this audited sample of
\AuditEligible{} eligible sources and is not an estimate of prevalence in the
compression literature, which the three frames were never drawn to represent,
and the audit measures evidential sufficiency only, saying nothing about whether
an audited model is in fact equivalent.

The controlled experiment covers all 8 registered confirmatory cells (1.5B to
8B parameters, two benchmarks, 4 bits, two methods, five seeds) and nothing
beyond them, so the verdict licenses no statement about 3-bit behaviour,
ARC-Challenge, HellaSwag, calibration-\emph{dataset} effects, or any model or
benchmark outside the registered set. The atlas reaches 405B but is
observational; the controlled grid is causal but small, and eight cells is a
small number of cells however cleanly they were obtained. The experiment's
two-level bootstrap (\S\ref{sec:minigrid:supporting}) resamples seed labels from a sample
of five, so it has limited resolution on the seed-level distribution it draws
from, and where its rank-flip rate is low what that bounds is the stability of
the \emph{observed} five-seed mean under resampling. The three cells that
combine a winner flip with a rank-flip rate below $0.05$ should be read as
``one calibration run can mislead, and five were more stable here'', not as a
recommendation of five; a design aimed at the seed-level distribution itself
would need more seeds than the registration committed to. The four GSM8K
confirmatory cells ran at $n = 1{,}000$, where a post-hoc resolution analysis
puts the seed-induced range at roughly two to three-and-a-half paired standard
errors and the mean gap itself at or below $1.25$\,SE in three of the four cells
(\S\ref{sec:minigrid:resolution}), and seed-level SD and item-level SE are of
the same order there, where on MMLU the seed term dominates. This paper's own
certification table asked for about $1{,}184$ GSM8K items at median discordance
and $3{,}068$ at p75, and the discordance observed in these cells sits at or
above p75, so the shortfall was predicted by \S\ref{sec:certification} before it
was measured (\S\ref{sec:minigrid:selfaudit}); the verdict is unaffected, being
a count over the frozen rule and not a significance test, but the MMLU cells
carry the evidence and any per-cell reading of GSM8K should say so. Finally, all
benchmarks are English and predominantly multiple-choice or short-answer, with
per-item correctness defined by the original harness, and float-scored
generation tasks are outside the flip model entirely: 33 atlas cells were
excluded for exactly this reason, so nothing here applies to quality claims
about open-ended generation, including the CIDEr-style metric that put
\AuditOutsideFrameworkClaim{} outside our registered binary paired-outcome
calculation (\S\ref{sec:audit:r04}). That is a limit of the flip model we
registered, not of paired analysis, which handles graded metrics perfectly well.

\subsection{Continuous and graded metrics, and what certifying them would take}
\label{sec:limitations:continuous}

The binary outcome model is the paper's binding scope condition, and it was the
necessary first step rather than a convenience. With per-item correctness the
paired difference $d_i$ takes values in $\{-1,0,+1\}$, so its variance under a
true difference of zero is exactly the discordance rate $p_d$, a quantity two
per-item dumps determine by counting; that identity is what makes
Table~\ref{tab:certification} an empirical table instead of a modelling
assumption, and it is also the only outcome type the public record and the
audited sources supply. TOST itself is not so restricted. It applies to the mean
paired difference of any bounded per-item score, with $p_d$ replaced by
$\sigma_d^2$, the variance of the per-item score difference, and the binary
formula of \S\ref{sec:cert:method} is the case $\sigma_d^2 = p_d$. What does not
come with that generalisation is everything empirical: $\sigma_d$ has to be
measured per metric and per family, and this paper measures it for none, since
the atlas records accuracy-state churn and the float-scored cells it met were
excluded rather than scored. Churn also stops being a count of flips and becomes
a dispersion of score changes, and the margin has to be declared in the metric's
own units, where two CIDEr points and two accuracy points are unrelated
quantities. Appendix~\ref{app:continuous} states the generalisation, works
through pass@$k$, n-gram and description-overlap metrics and judge scales, and
returns to \AuditOutsideFrameworkClaim{} to say what scoring it would have
required. None of that extension is validated here, and no required-$n$ reported
in this paper transfers to a metric outside the binary model.

%
%
%

\section{Conclusion: a reporting standard for compression claims}
\label{sec:conclusion}

%
Figure~\ref{fig:cancellation} is one evaluation of one model, and it is the
argument in miniature. Two 4-bit quantizations of Qwen2.5-7B that sit $0.58$
points apart on GSM8K disagree on $17.7\%$ of the items; certifying equivalence
at a declared $\pm 2$-point margin needs $2{,}730$ items and the evaluation ran
$1{,}000$; and which of the two comes out ahead changes with the calibration
draw. None of that is visible in the two aggregate scores. The cell is the most
extreme of the eight registered cells, but the mechanism is not peculiar to it:
across the $1{,}707$ cells of the public record, median churn is about five
times the median net accuracy delta (\S\ref{sec:atlas}). Aggregate accuracy is
not a summary of behaviour, and a net delta is not a summary of what a
compression changed.

The claim ``near-lossless'' is cheap to write and expensive to support. This
paper measured the gap, and what the measurement mostly found was absence, not
error. Across \AuditEligible{} eligible sources, none declares a prospective
numerical equivalence margin and none releases \emph{task-matched} per-item
outputs, though \AuditPerItemOtherTaskOnly{} release outputs for other tasks
only, and \AuditNotAssessable{} cannot be assessed at all from what they report
(\S\ref{sec:audit}). These are counts in this audited sample of
\AuditEligible{} sources, not estimates for the literature. The apparatus that
closes the gap is equivalence certification at a declared margin, with
sample-size tables computed from observed churn rather than assumed variance
(\S\ref{sec:certification}); the same apparatus is being extended to
anytime-valid sequential certification, a component that is registered but has
not been run and for which no results are reported here.

The controlled experiment says where this standard actually bites, and it is not
where the five lines suggest. Each of them is written as though the comparison
at issue were a compressed model against its original. The cell in
Figure~\ref{fig:cancellation} is not that comparison: H3 puts two
\emph{compressed} models side by side, GPTQ against AWQ at the same bit width,
which is the choice a practitioner makes after deciding to compress at all. The
evidence problem is strictly worse there, at a median per-cell churn of
$12.7\times$ the net delta against roughly four times in the atlas on the same
aggregation, and with the winner reversing on nothing but the calibration seed
in 5 of 8 registered cells (\S\ref{sec:minigrid}). Two compressed models are
more alike than either is to the original, so cancellation is more complete and
the aggregate hides more.

That generalises the standard of \S\ref{sec:intro} past its own framing, and
nothing in those five lines is about quantization. They are about comparing two
models similar enough that somebody thought the comparison worth making:
compressed against original, method against method, checkpoint against
checkpoint, version against version. Aggregate accuracy is least informative in
exactly the regime where these comparisons are made, and what makes such a
comparison checkable is per-item evidence and a margin declared before the
evaluation is run.

\bibliographystyle{plainnat}
\bibliography{references}

\begin{thebibliography}{23}
\providecommand{\natexlab}[1]{#1}
\providecommand{\url}[1]{\texttt{#1}}
\expandafter\ifx\csname urlstyle\endcsname\relax
  \providecommand{\doi}[1]{doi: #1}\else
  \providecommand{\doi}{doi: \begingroup \urlstyle{rm}\Url}\fi

\bibitem[Bronder(2026)]{bronder2026instrument}
Justin Bronder.
\newblock Instrument effects in language-model honesty evaluation: An auditable
  single-system demonstration, 2026.
\newblock arXiv:2607.14399.

\bibitem[Cacioli(2026)]{cacioli2026beyondmean}
Jon-Paul Cacioli.
\newblock Beyond the mean: Within-model reliable change detection for {LLM}
  evaluation, 2026.
\newblock arXiv:2604.27405. Preregistration: osf.io/3dnsa.

\bibitem[Dettmers et~al.(2022)Dettmers, Lewis, Belkada, and
  Zettlemoyer]{llmint82022}
Tim Dettmers, Mike Lewis, Younes Belkada, and Luke Zettlemoyer.
\newblock {LLM.int8()}: 8-bit matrix multiplication for transformers at scale.
\newblock In \emph{Advances in Neural Information Processing Systems
  (NeurIPS)}, 2022.
\newblock arXiv:2208.07339.

\bibitem[Dodge et~al.(2019)Dodge, Gururangan, Card, Schwartz, and
  Smith]{dodge2019showyourwork}
Jesse Dodge, Suchin Gururangan, Dallas Card, Roy Schwartz, and Noah~A. Smith.
\newblock Show your work: Improved reporting of experimental results.
\newblock In \emph{Proceedings of the 2019 Conference on Empirical Methods in
  Natural Language Processing and the 9th International Joint Conference on
  Natural Language Processing (EMNLP-IJCNLP)}, pages 2185--2194, Hong Kong,
  China, 2019. Association for Computational Linguistics.
\newblock \doi{10.18653/v1/D19-1224}.
\newblock arXiv:1909.03004.

\bibitem[Dutta et~al.(2024)Dutta, Krishnan, Kwatra, and Ramjee]{dutta2024flips}
Abhinav Dutta, Sanjeev Krishnan, Nipun Kwatra, and Ramachandran Ramjee.
\newblock Accuracy is not all you need.
\newblock In \emph{Advances in Neural Information Processing Systems
  (NeurIPS)}, volume~37, 2024.
\newblock arXiv:2407.09141.

\bibitem[Frantar and Alistarh(2023)]{sparsegpt2023}
Elias Frantar and Dan Alistarh.
\newblock {SparseGPT}: Massive language models can be accurately pruned in
  one-shot.
\newblock In \emph{Proceedings of the 40th International Conference on Machine
  Learning}, volume 202 of \emph{Proceedings of Machine Learning Research},
  pages 10323--10337. PMLR, 2023.
\newblock arXiv:2301.00774.

\bibitem[Frantar et~al.(2023)Frantar, Ashkboos, Hoefler, and
  Alistarh]{gptq2022}
Elias Frantar, Saleh Ashkboos, Torsten Hoefler, and Dan Alistarh.
\newblock {GPTQ}: Accurate post-training quantization for generative
  pre-trained transformers.
\newblock In \emph{International Conference on Learning Representations
  (ICLR)}, 2023.
\newblock arXiv:2210.17323.

\bibitem[Gebru et~al.(2021)Gebru, Morgenstern, Vecchione, Vaughan, Wallach,
  Daum{\'e}~III, and Crawford]{gebru2021datasheets}
Timnit Gebru, Jamie Morgenstern, Briana Vecchione, Jennifer~Wortman Vaughan,
  Hanna Wallach, Hal Daum{\'e}~III, and Kate Crawford.
\newblock Datasheets for datasets.
\newblock \emph{Communications of the ACM}, 64\penalty0 (12):\penalty0 86--92,
  2021.
\newblock \doi{10.1145/3458723}.
\newblock arXiv:1803.09010.

\bibitem[Gringras and Salahshoor(2026)]{gringras2026frontierlag}
David Gringras and Misha Salahshoor.
\newblock Frontier lag: A bibliometric audit of capability misrepresentation in
  academic {AI} evaluation, 2026.
\newblock arXiv:2605.04135. Preregistered on OSF: 10.17605/OSF.IO/7XM3D.

\bibitem[Helcig et~al.(2026)Helcig, Kurtic, and Alistarh]{helcig2026slq}
Michael Helcig, Eldar Kurtic, and Dan Alistarh.
\newblock Statistically-lossless quantization of large language models, 2026.
\newblock arXiv:2605.02404.

\bibitem[Kim et~al.(2024)Kim, Hooper, Gholami, Dong, Li, Shen, Mahoney, and
  Keutzer]{squeezellm2023}
Sehoon Kim, Coleman Richard~Charles Hooper, Amir Gholami, Zhen Dong, Xiuyu Li,
  Sheng Shen, Michael~W. Mahoney, and Kurt Keutzer.
\newblock {SqueezeLLM}: Dense-and-sparse quantization.
\newblock In \emph{Proceedings of the 41st International Conference on Machine
  Learning}, volume 235 of \emph{Proceedings of Machine Learning Research},
  pages 23901--23923. PMLR, 2024.
\newblock arXiv:2306.07629.

\bibitem[K{\"u}bler et~al.(2026)K{\"u}bler, Budhathoki, Kleindessner, Zhou,
  Yin, Khetan, and Karypis]{llmaccuracystats2026}
Jonas K{\"u}bler, Kailash Budhathoki, Matth{\"a}us Kleindessner, Xiong Zhou,
  Junming Yin, Ashish Khetan, and George Karypis.
\newblock When {LLM}s get significantly worse: A statistical approach to detect
  model degradations.
\newblock In \emph{International Conference on Learning Representations
  (ICLR)}, 2026.
\newblock arXiv:2602.10144.

\bibitem[Lin et~al.(2024)Lin, Tang, Tang, Yang, Chen, Wang, Xiao, Dang, Gan,
  and Han]{awq2023}
Ji~Lin, Jiaming Tang, Haotian Tang, Shang Yang, Wei-Ming Chen, Wei-Chen Wang,
  Guangxuan Xiao, Xingyu Dang, Chuang Gan, and Song Han.
\newblock {AWQ}: Activation-aware weight quantization for on-device {LLM}
  compression and acceleration.
\newblock In \emph{Proceedings of Machine Learning and Systems (MLSys)},
  volume~6, 2024.
\newblock arXiv:2306.00978. MLSys 2024 Best Paper Award.

\bibitem[Liu et~al.(2025)Liu, Zhao, Fedorov, Soran, Choudhary, Krishnamoorthi,
  Chandra, Tian, and Blankevoort]{spinquant2024}
Zechun Liu, Changsheng Zhao, Igor Fedorov, Bilge Soran, Dhruv Choudhary,
  Raghuraman Krishnamoorthi, Vikas Chandra, Yuandong Tian, and Tijmen
  Blankevoort.
\newblock {SpinQuant}: {LLM} quantization with learned rotations.
\newblock In \emph{International Conference on Learning Representations
  (ICLR)}, 2025.
\newblock arXiv:2405.16406.

\bibitem[Marie et~al.(2021)Marie, Fujita, and Rubino]{marie2021mtaudit}
Benjamin Marie, Atsushi Fujita, and Raphael Rubino.
\newblock Scientific credibility of machine translation research: A
  meta-evaluation of 769 papers.
\newblock In \emph{Proceedings of the 59th Annual Meeting of the Association
  for Computational Linguistics and the 11th International Joint Conference on
  Natural Language Processing (Volume 1: Long Papers)}, pages 7297--7306,
  Online, 2021. Association for Computational Linguistics.
\newblock \doi{10.18653/v1/2021.acl-long.566}.
\newblock arXiv:2106.15195.

\bibitem[Nikoli{\'c} et~al.(2026)Nikoli{\'c}, Zadeh, Torres~Sanchez, and
  Moshovos]{nikolic2026displacement}
Milo{\v{s}} Nikoli{\'c}, Ali~Hadi Zadeh, Enrique Torres~Sanchez, and Andreas
  Moshovos.
\newblock Displacement is not direction: Evaluating fidelity metrics for
  quantized {LLM} deployment, 2026.
\newblock arXiv:2606.19558.

\bibitem[Paglieri et~al.(2024)Paglieri, Dash, Rockt{\"a}schel, and
  Parker-Holder]{paglieri2024outliers}
Davide Paglieri, Saurabh Dash, Tim Rockt{\"a}schel, and Jack Parker-Holder.
\newblock Outliers and calibration sets have diminishing effect on quantization
  of modern {LLM}s, 2024.
\newblock arXiv:2405.20835.

\bibitem[Rababah et~al.(2026)Rababah, Akcora, and Leung]{rababah2026illusion}
Baha Rababah, Cuneyt~Gurcan Akcora, and Carson~K. Leung.
\newblock The illusion of equivalency: Statistical characterization of
  quantization effects in {LLM}s, 2026.
\newblock arXiv:2607.08734.

\bibitem[Sun et~al.(2024)Sun, Liu, Bair, and Kolter]{wanda2023}
Mingjie Sun, Zhuang Liu, Anna Bair, and J.~Zico Kolter.
\newblock A simple and effective pruning approach for large language models.
\newblock In \emph{International Conference on Learning Representations
  (ICLR)}, 2024.
\newblock Wanda. arXiv:2306.11695.

\bibitem[Thomas et~al.(2026)Thomas, Gligoric, and Shah]{thomas2026prereg}
Maria Thomas, Kristina Gligoric, and Nihar~B. Shah.
\newblock Mitigating {LLM}-based p-hacking by preregistering for the next
  {LLM}, 2026.
\newblock arXiv:2606.27687.

\bibitem[Tseng et~al.(2024)Tseng, Chee, Sun, Kuleshov, and
  De~Sa]{quipsharp2024}
Albert Tseng, Jerry Chee, Qingyao Sun, Volodymyr Kuleshov, and Christopher
  De~Sa.
\newblock {QuIP\#}: Even better {LLM} quantization with hadamard incoherence
  and lattice codebooks.
\newblock In \emph{Proceedings of the 41st International Conference on Machine
  Learning}, volume 235 of \emph{Proceedings of Machine Learning Research},
  pages 48630--48656. PMLR, 2024.
\newblock arXiv:2402.04396.

\bibitem[Williams and Aletras(2024)]{williamsaletras2024}
Miles Williams and Nikolaos Aletras.
\newblock On the impact of calibration data in post-training quantization and
  pruning.
\newblock In \emph{Proceedings of the 62nd Annual Meeting of the Association
  for Computational Linguistics (Volume 1: Long Papers)}, pages 10100--10118,
  Bangkok, Thailand, 2024. Association for Computational Linguistics.
\newblock \doi{10.18653/v1/2024.acl-long.544}.
\newblock arXiv:2311.09755.

\bibitem[Xiao et~al.(2023)Xiao, Lin, Seznec, Wu, Demouth, and
  Han]{smoothquant2022}
Guangxuan Xiao, Ji~Lin, Mickael Seznec, Hao Wu, Julien Demouth, and Song Han.
\newblock {SmoothQuant}: Accurate and efficient post-training quantization for
  large language models.
\newblock In \emph{Proceedings of the 40th International Conference on Machine
  Learning}, volume 202 of \emph{Proceedings of Machine Learning Research},
  pages 38087--38099. PMLR, 2023.
\newblock arXiv:2211.10438.

\end{thebibliography}

\appendix
%
%

\section{Reconciling the calibration-sensitivity antecedents}
\label{app:reconcile}

\textbf{The two most direct antecedents of this paper reach opposite
conclusions about the same knob.} \citet{williamsaletras2024} find
\emph{substantial} variation in downstream task performance across calibration
sets, explicitly contrasting their result with prior work that suggested greater
robustness. \citet{paglieri2024outliers} find the effect \emph{diminishing}:
where the older OPT models degrade badly and vary with the calibration set,
Llama-2 7B, Llama-3 8B, Command-R 35B and Mistral 7B are robust, with Mistral 7B
close to immune, and they argue the field should stop building its quantization
literature around outlier preservation. Both are careful studies of calibration
sensitivity, published within a year of each other, pointing in opposite
directions.

The second of these is also the strongest available objection to this paper,
and it is the one a reader is most likely to arrive already holding: if modern
models are insensitive to the calibration set, why would the calibration seed
reorder anything?

Our answer is that both results hold at once, and that the disagreement between
them is a symptom of measuring the wrong thing for the question the field
actually asks. Three differences do the work, and our own numbers show why.

\paragraph{The intervention is different, and ours is strictly finer.} Both
antecedents vary the calibration \emph{data}. \citet{paglieri2024outliers} do it
across sets differing in quality, content and language: a
RedPajama sample, calibration drawn from uniformly random ASCII punctuation and
whitespace, task-specific sets from ARC-Challenge and PiQA, and multilingual
sets from FLORES+. We hold the corpus fixed and vary only the sample seed, an
intervention they do not make. A finding that swapping the corpus for random
punctuation barely moves accuracy does not entail that redrawing the sample
leaves the \emph{ordering} of two methods intact; the two questions are not
nested in the direction the objection needs.

\paragraph{The unit of analysis is different.} \citet{paglieri2024outliers}
study one method at a time, reporting GPTQ W4A16, AWQ W4A16, SmoothQuant W8A8
and a naive W8A8 baseline in separate result sections, and they report no
head-to-head ranking between methods and no ranking flips. Our unit of analysis
is the \emph{gap between} two methods at the same bit width, a quantity neither
antecedent's design measures.

\paragraph{Individual robustness plus a small method gap implies ranking
instability.} This is the reconciliation proper, and it is what
\S\ref{sec:minigrid} measures. Each method is individually
stable in absolute accuracy, much as they report, and the seed-induced range is
at least as large as the mean GPTQ--AWQ gap in 7 of our 8 confirmatory cells.
Those two facts together \emph{are} ranking instability. We are not
contradicting their robustness finding; we are showing it has a consequence they
did not test for. A field that reads ``calibration choice barely moves accuracy''
as licence to compare two methods from one calibration run each has drawn the
wrong inference from a correct result.

\paragraph{The magnitude question.} Robustness claims are
claims about size, so the reconciliation has to be quantitative. The figures
that follow come from a \emph{post-hoc} resolution analysis, prompted by this
very paper and labelled as such wherever it appears (\S\ref{sec:minigrid}); it
is descriptive and modifies no verdict. On MMLU our seed-induced ranges run 5.5
to 17.5 paired standard errors, a spread no robustness claim reaches, and one
the benchmark resolves comfortably at $n = 14{,}042$.
On GSM8K at $n = 1{,}000$ the same ranges run roughly two standard errors, and
we say so instead of averaging the two tasks together: that arm of the
experiment does not resolve the effect it was built to measure
(\S\ref{sec:minigrid:resolution}, \S\ref{sec:limitations}).

\paragraph{What the disagreement was about.} Read this way, the two antecedents
are not really in conflict. \citet{williamsaletras2024} and
\citet{paglieri2024outliers} both measure how much a \emph{single} method's
absolute accuracy moves when the calibration data changes, and they disagree
about the size of that movement on the model generations each studied. Neither
measures what a practitioner comparing two methods actually depends on, namely that the
\emph{ordering} survives, and a quantity nobody measured is one the field has no
basis to assume stable. Our design does not settle their disagreement. It shows
that settling it would not have answered the question either way.

\subsection{Adjacent concurrent measurements}
\label{app:related:concurrent}

\citet{cacioli2026beyondmean} adapts the clinical Reliable Change Index to
per-item comparisons between model \emph{versions} rather than precisions, and
likewise recommends reporting a churn rate beside the mean;
\citet{nikolic2026displacement} extend the flips metric into a leapfrog/drop
decomposition over community-published quantized checkpoints, and find
that KL-divergence proxies lose their ranking signal in precisely the
near-baseline region our certification tables adjudicate.

\subsection{Losslessness definitions and fidelity metrics}
\label{app:related:lossless}

\citet{helcig2026slq} occupy the phrase this paper is about. They formalise
notions of losslessness for quantized LLMs (task-lossless and the stricter
distribution-lossless), propose the Expected Acceptance Rate as an interpretable
fidelity metric, and ship SLQ, a method that reaches those targets at low bit
widths. Their question and ours are different, and the difference is the whole
of \S\ref{sec:certification}. \emph{They define losslessness and build a method
to achieve it.} \emph{We audit whether existing published claims of losslessness
have the evidence to support them, and compute how many items it would take to
certify one.} Their paper contains no equivalence test at a declared margin, no
power or required-$n$ computation, and no audit of others' claims; ours contains
no quantization method. A practitioner who adopts EAR as a fidelity target still
needs to know how large an evaluation must be before the accuracy half of a
losslessness claim means anything, and that number is what our certification
tables supply.

\subsection{The constructive-audit genre, and preregistration precedent}
\label{app:related:genre}

The constructive-audit genre is well established: \citet{dodge2019showyourwork}
on reporting the compute and search behind reported numbers, and
\citet{marie2021mtaudit} on the statistical practice of machine-translation
evaluation. We follow both in framing and in tone: the finding is about what the
field reports, not about whether individual authors are right, and the paper's
deliverable is a standard plus the tooling to meet it. The genre has also begun
to adopt preregistration directly, which is the precedent for
\S\ref{sec:prereg}: \citet{gringras2026frontierlag} run a preregistered
bibliometric audit of which models the evaluation literature actually tests, and
\citet{thomas2026prereg} propose preregistering an analysis against a model not
yet released, so that the configuration cannot be tuned against the outcome.

\subsection{The compression families the audited claims span}
\label{app:related:families}

The audited claims span the two dominant post-training families: weight-only
quantization, represented by GPTQ \citep{gptq2022}, AWQ \citep{awq2023} and
SqueezeLLM \citep{squeezellm2023}; weight-and-activation quantization, by
LLM.int8() \citep{llmint82022} and SmoothQuant \citep{smoothquant2022}; and
one-shot pruning, by SparseGPT \citep{sparsegpt2023} and Wanda
\citep{wanda2023}. We audit the equivalence \emph{language} these papers use
and the evidence offered for it, not the methods themselves, and the audit's
verdicts are about reporting practice, not about whether a method works.
Two of the most aggressive methods in the set are also the most restrained in
what they assert: SpinQuant \citep{spinquant2024} and QuIP\#
\citep{quipsharp2024} report their accuracy deltas without equivalence
language, and are cited here as positive examples alongside the Qwen quantized
model cards and the \texttt{llama.cpp} documentation.

\section{Preregistration detail}
\label{app:prereg-detail}

This appendix holds the full text of seven items summarised in
\S\ref{sec:prereg}: the freeze timeline, what the freezes fix in advance, the
disclosed pre-registration data contact, the seven interpretive choices, the
result-inspection discipline, the H3 reporting rule as registered, and the
spot-check narrative.

\subsection{The freeze timeline}
\label{app:prereg:timeline}

\begin{table}[!t]
\centering
\small
\caption{Freeze timeline. Each artifact was committed before the analysis it
governs could be run, and each frozen file carries a \emph{Dated Amendments}
section: deviations are appended, never edited in, and each records whether
results had been inspected before the decision.}
\label{tab:freeze-timeline}
\setlength{\tabcolsep}{4pt}
\small
\begin{tabular}{lll>{\raggedright\arraybackslash}p{4.6cm}}
\toprule
Date & Artifact & Commit & What it fixed in advance \\
\midrule
2026-07-11 & \texttt{PREREGISTRATION.md} & --- & Grid, metrics, TOST margin (2\,pp), H3 decision rule \\
2026-07-15 & Three registrations & \texttt{b74fd58} & Audit, atlas-mining, and mini-grid protocols \\
2026-07-15 & Audit Amendment 1 & \texttt{d6e02dd} & Blind-extraction independence mechanism \\
2026-07-15 & Pair manifest & \texttt{f06348f} & The 59 atlas pairs, before any flip statistic \\
2026-07-15 & Claim table & \texttt{715a7ce} & The 17 audited claims, before any verdict \\
2026-07-20 & WikiText-2 amendment & --- & ``Document'' definition, Decision Point A \\
\bottomrule
\end{tabular}
\end{table}

\subsection{What the freezes fix in advance}
\label{app:prereg:whatfixed}

Table~\ref{tab:freeze-timeline} dates every freeze against its commit. The
frozen claim table's content hash is recorded with the verdicts (sha256
\texttt{842b9756...5af7b15}), as is the atlas summary's
(\texttt{98201ade...10a4712d}) and the container image's, so a reader can
confirm that the inputs to any reported number were the frozen ones.

Three properties of that arrangement do work in the paper. The mechanical parts
of the audit, meaning the inclusion trigger vocabulary, the extraction fields,
the verdict formulas and the robustness sweep, were fixed before any claim's
power was computed, so the audit cannot have selected its claims or its
statistics to produce a headline. The atlas pair list was fixed before any flip
statistic, so the atlas cannot have selected pairs to produce churn. And the H3
decision rule was fixed, in algebra, before the first compressed checkpoint
existed, so the paper reports whatever that rule returns.

\subsection{Result-inspection discipline}
\label{app:prereg:discipline}

During grid execution only job health, checksums, expected-file coverage and
receipt pairing are inspected; the first accuracy inspection happens only after
the mini-grid validator passes over the complete expected file set. The
registered analysis is run once per cell, the escalation rule applied
mechanically, and a dated escalation decision record written the same day. The
paired-bootstrap rank-flip denominator convention (tie replicates included,
ties also reported separately) was fixed in words in the registration, naming
the implementing commit, so it cannot be re-chosen after inspection. No
registered analysis is tuned after results are seen: not the calibration
builder, not the paired bootstrap, not the verdict rule.

\subsection{The H3 reporting rule, stated before the results exist}
\label{app:prereg:h3rule}

The registered decision rule is defined over eight confirmatory
model-by-benchmark cells:
$\{$Qwen2.5-1.5B, Qwen2.5-7B, Llama-3.2-3B, Llama-3.1-8B$\}
\times \{$MMLU, GSM8K$\}$ at 4 bits, comparing GPTQ seed $s$ against AWQ
seed $s$ on byte-identical calibration samples. H3 is \emph{supported} if winner
flips occur in at least 3 of the 8 cells, or the range/gap criterion holds in at
least 4 of the 8. It is \emph{disconfirmed} if winner flips occur in at most 1
of the 8 and $\max(\mathrm{range}_{\mathrm{GPTQ}}, \mathrm{range}_{\mathrm{AWQ}})
< 0.5 \times \mathrm{gap}$ in at least 6 of the 8. Everything else is
\emph{inconclusive}, reported without post-hoc promotion.

The mini-grid executed \textbf{4 of those 8 cells}
($\{$Qwen2.5-1.5B, Llama-3.2-3B$\} \times \{$MMLU, GSM8K$\}$), with the 7B/8B
cells deferred behind a pre-committed, mechanical escalation rule. The mini-grid
registration stated the consequence of each branch in advance: if all eight
cells complete, the frozen rule is applied exactly as registered; if they do
not, the paper reports the four completed cells descriptively and H3 is
undecided under the registered rule, never supported or disconfirmed on four
cells, and no reduced-cell variant of the rule is constructed after results are
seen.

The escalation screen fired on 2026-07-23, the 7B/8B cells were built, and
\textbf{all eight cells completed}, so the second branch never applied and the
frozen rule was applied once over the full set. Both decisions are dated, signed
records
(\texttt{docs/MINIGRID\_ESCALATION\_DECISION\_2026-07-23.md},
\texttt{docs/H3\_EIGHT\_CELL\_DECISION\_2026-07-26.md}), and the screen's own
authorship discipline is worth stating: it decides which cells to build and
states no H3 outcome, so the confirmatory rule was never applied to a cell set
selected after its result was known. The screen's per-cell outcomes are in
Appendix~\ref{app:escalation} and the verdict in
\S\ref{sec:minigrid:verdict}; this subsection states neither.

\subsection{Disclosed pre-registration data contact}
\label{app:prereg:contact}

Preregistration is a claim about ordering, so the places where our order was
imperfect are disclosed by the registrations themselves rather than discovered
by a reader.

\textbf{Audit.} Five candidate claims were collected with exact quotes on
2026-07-15 during a feasibility sweep, before the registration was written
(GPTQ, LLM.int8(), and SmoothQuant abstracts; the Red~Hat AI W4A16 Llama-3.1-8B
model card; Meta's quantized-Llama blog). No power computation had been run on
any of them. They enter the pool through the same §3 criteria as
later-collected claims, and they are not treated as a separate stratum, because
the contact was with the \emph{quotes}, not with any verdict quantity.

\textbf{Atlas.} Two feasibility probes were run on 2026-07-15 before the
registration was drafted, and their results were known: TheBloke/Llama-2-7B-GPTQ
against meta-llama/Llama-2-7b-hf on ARC-Challenge (74 flips of 1{,}170, net
$-1.03$\,pp), and a Neural~Magic Llama-3.1-8B baseline against its W4A16
quantization on \texttt{bbh\_boolean\_expressions} (17 flips of 250, net
$+1.2$\,pp). Those pairs are flagged in the atlas and excluded from every
headline aggregate; the exclusion covers all 99 cells belonging to the two probe
\emph{pairs}, not merely the two probe tasks (\S\ref{app:prereg:choices},
interpretive choice~6).

\textbf{A retired anecdote.} The project's original motivating observation was a
feasibility rerun of a bnb-4bit checkpoint against its own repository's
full-precision weights, in which roughly 6.3\% of items flipped correctness while
the aggregate score was unchanged. That observation is what made the project
seem worth doing, and it is reported here for that reason. It is
\textbf{retired from every quantitative use}, on two independent grounds: it is
pre-registration data contact, and it lies outside the frozen 59-pair manifest,
since same-repository multi-precision runs are an amendment candidate that was
never acted upon. The registered replacement is the identical-score statistic of
\S\ref{sec:atlas:identical}, computed on the frozen atlas population: 145 of
1{,}707 cells (8.49\%) at a median churn of 7.20\%. Where an anecdote and a
registered statistic point the same way, the paper cites the statistic.

\subsection{The interpretive choices that moved the headline}
\label{app:prereg:choices}

Frozen protocols do not eliminate analyst judgement; they make it visible and
datable. Seven passages of the audit registration were ambiguous enough to
require a ruling before verdicts could be computed. All seven were ruled on
2026-07-20, each is implemented in code, and each is reversible by re-running
with the alternative. Four matter for how the results should be read.

\paragraph{Choice 1: which margin is ``the applicable margin'', and the
$K = 1 \to 5 \to 4$ correction.}
%
The frozen §4 names the 2\,pp registered margin first, adds ``(and at the
claim's own margin when it states one)'', and then labels the verdict
``underpowered for its own assertion'' \emph{at the applicable margin}. We read
that as leaving two live options, and chose the second. That was an error, and
Amendment~2 corrects it.

The conditional in the frozen text is the whole of it: \emph{when it states
one}. No audited source states one (\S\ref{sec:audit:taxonomy}). What the
superseded implementation substituted was each source's largest reported
$|\Delta|$, and the reasoning it ran on was that ``a source asserting parity
within 0.15\,pp has made a 0.15\,pp claim''. It has not. A reported delta is an
outcome of an evaluation that has already been run; a margin is a tolerance
fixed before it runs, and a claim cannot be held to a decision rule its authors
never adopted. Treating the one as the other inverted the direction of the
inference and, because a measured delta is typically far tighter than any
tolerance a author would have declared, it also made the test far harsher than
the registration authorised.

The sequence is reported exactly, because the order in which the numbers arrived
is itself the disclosure. The first pass applied the 2\,pp margin uniformly and
returned $K = 1$ of 12. Re-reading the frozen label produced $K = 5$; ruling on
choice~3 below moved one claim out of the determinate set, giving the $K = 4$
that was reported for ten days. Amendment~2 returns the analysis to the uniform
registered margin, where the rev-3 recomputation gives \textbf{1 of 11
assessable claims below the approximate planning threshold}, the same claim the
very first pass had flagged.

\paragraph{Choice 1, as the main text stated it.}
%
The conditional decides it. \emph{When it states one}: no audited source states
one, so the applicable margin is the registered 2\,pp throughout, and what the
superseded implementation had substituted was each source's largest reported
delta. That substitution treats a measured outcome as a decision rule the source
never adopted. The sequence of values is reported plainly because it is the
disclosure: the first pass returned $K = 1$ of 12, the claim-specific reading
raised it to $K = 5$, the metric-compatibility ruling on R04 gave the $K = 4$
that was published, and the rev-3 recomputation at the registered margin returns
\textbf{\AuditBelowThresholdAtMedian{} of \AuditAssessable{} assessable claims
below the approximate planning threshold}, which is the claim the first pass had
flagged. All readings ship in the released CSV (\S\ref{sec:audit:results}).

\paragraph{Choice 2: what counted as the claim's own margin (withdrawn).}
%
This choice is withdrawn with the margins it operated on, and is recorded
because the withdrawn quantities were reported publicly.

Most claims cover several benchmarks with different deltas, so the superseded
implementation defined the claimed margin as the largest $|\Delta|$ the source
reports. Within its own frame that was the reading most favourable to the
source, granting the claim the widest margin its sentence could bear and
therefore the smallest required sample size. The point Amendment~2 makes is that
the frame was wrong to begin with: no choice among a source's reported deltas
yields a margin, because none of them is one.

Two published quantities fall with it. The shortfall range of $2.0\times$ to
$12.9\times$ was required $n$ over reported $n$ at those derived margins, and it
is \textbf{withdrawn rather than recomputed}; at the registered 2\,pp margin ten
of the eleven assessable claims have no shortfall at all, so there is no range
to restate. The MDD-to-margin ratios are likewise recomputed against the
registered margin in Table~\ref{tab:audit-mdd}, where the values themselves are
unchanged and only the denominators move.

What survives the withdrawal is the disposition of the remaining choices, which
were and remain resolved in the source's favour: pairing rather than independent
binomial variance (the paired assumption is the generous one, and
Table~\ref{tab:audit-mdd} shows the independent bound is uniformly worse), and
imputation by \emph{median} discordance from the most specific matching atlas
tier rather than by any upper quantile.

\paragraph{Choice 3: R04 is indeterminate rather than scored on a substituted
benchmark.}
Discussed in full at \S\ref{sec:audit:r04}, and unaffected by Amendment~2: R04
sits outside the registered binary paired-outcome calculation whichever margin
is applied. It is recorded here as a ruling because it is the second correction
that cost the paper a number. Under the rules then current it removed what the
first pass had reported as the audit's largest shortfall ($38.3\times$) and
moved the headline from $K = 5, J = 4$ to $K = 4, J = 5$; those figures are
quoted as history and are not current results. The GSM8K computation is retained
in the released CSV as a labelled transparency column so that the set-aside
quantity is inspectable rather than deleted.

\paragraph{Choice 5: TOST uses the one-sided $z$.}
The project's own helper, \texttt{flipeval.required\_n\_for\_effect}, uses the
two-sided $z_{1-\alpha/2}$. That is correct for \emph{detection} and wrong for
TOST, which rejects two one-sided nulls at level $\alpha$ each. Reusing the
helper unchanged would have inflated every required $n$ in this paper by about
27\%, in the conservative direction, under the name ``TOST'', and with no
symptom visible in any output. It would have made every audit shortfall and
every certification count larger than the correct value. The TOST formula is
therefore implemented locally with the one-sided $z_{1-\alpha} = 1.6449$,
documented, and cross-checked against the project implementation on the
quantities the two share; the paired standard deviation is pinned against
\texttt{flipeval.core.minimum\_detectable\_difference} on synthetic delta vectors
in the test suite, so the audit cannot silently fork from the tested library.

\paragraph{The remaining three rulings, in brief.}
Choice~4: verdicts are computed at claim level rather than
claim\,$\times$\,benchmark, because the frozen table stores one pooled $n$ per
claim and operating at a finer granularity would mean inventing rows the freeze
does not contain; every summed task is listed in the released
\texttt{n\_basis} column. Choice~6: ``the two probe-tagged cells'' is read as
two probe \emph{pairs}, comprising 99 cells, all excluded. Choice~7: degenerate
rows are handled per input rather than discarded whole, so an indeterminate
claim keeps the components its available inputs support, reported as
supplementary and never verdict-bearing.

\subsection{Atlas validation and population correction}
\label{app:prereg-spotcheck}

The protocol required an independent spot-check of the atlas pipeline before any
atlas number could be quoted externally. We report what it found, because the
finding is the paper's thesis applied to the paper.

The check re-downloaded the raw per-item files for ten cells, stratified across
both sources and across zero-delta, high-churn, McNemar-significant and excluded
strata, and recomputed the joins, pairing gates, flip counts, churn, net deltas
and exact McNemar $p$ values from a \emph{fresh reimplementation of the
registered definitions}, deliberately not a rerun of our own pipeline, since
rerunning the same code confirms determinism rather than correctness. All
262 compared fields reconciled exactly, including dyadic $p$ values reproduced
without \texttt{scipy}, and no upstream data drift was detected.

The arithmetic was right. The \emph{population} was not. Two defects surfaced.
First, our implementation omitted a clause of our own registration: when the
prompt-hash gate fails, the protocol requires trying earlier run combinations in
reverse-chronological order, and the code simply took the newest run and gave
up. Eleven pairs contributed zero cells as a result. Crucially, the loss was
\textbf{not random}: it removed exactly those pairs whose quantized side had
been re-evaluated later, a property correlated with a model being popular enough
to re-run. Second, a newer details schema nested its metrics one level deeper
than our parser looked, so 583 cells were recorded as having no binary
correctness metric when in fact they had one, and a sentence in our own results
note attributed that gap to the upstream leaderboard's reporting standards. We
retracted the sentence.

We draw three points from this, and none of them is that we were unlucky.

\textbf{An aggregate can be exactly right and still be built on the wrong
population.} Every cell we checked was computed correctly. Had we validated by
recomputing our own numbers, which is the natural and useless form of
self-check, we would have confirmed all of them and shipped the selection bias intact. What
caught it was reconstructing the measurement from the protocol text rather than
from the code.

Preregistration did the work it is supposed to do. The registration
text was frozen before any statistic existed, and it fully determined the
direction of the repair: we did not choose a rule that flattered a result, we
executed a rule that was already binding and had been under-implemented. That
is why the fix is a correction rather than a post-hoc adjustment, and it is why
we can say so without asking to be trusted. The correction was nevertheless made
\emph{after} results had been inspected, and we disclose that plainly rather
than presenting the repair as a pre-specified step.

Both revisions are public. We publish rev-1 and rev-2 and report the
delta between them, rather than replacing the record with its corrected version.
A field in which near-lossless claims cannot be rechecked because per-item
outputs are never released, which is the finding of Section~\ref{sec:audit}, is
a field whose corrections are invisible. Ours is not, and the difference is the point of
the artifact rather than an apology for it.

\subsection{The rev-1 to rev-2 delta}
\label{app:prereg:rev2delta}

The repair recovered cells rather than removing them, and the recovery was
confined to S1. The enumerated population is unchanged at 2{,}055 pair-task
cells; analysed cells rise from 1{,}254 to \textbf{1{,}807}, and the
probe-excluded analysable population from 1{,}155 to \textbf{1{,}707}. All of
that movement is S1 (846 $\to$ 1{,}398); S2 is unchanged at 309 cells, which is
the expected control, since no S2 cell's eligibility was affected by either
defect. The reverse-chronological run fallback was exercised for 1{,}007 cells
across 19 pairs, recorded per cell with the accepted run timestamps.

The descriptives moved in the direction a recovered population predicts rather
than in the direction that would flatter the paper. S1 median churn rises
slightly (0.1327 $\to$ 0.1375) and S1 median $|$net delta$|$ rises
(0.0226 $\to$ 0.0263), while the share of S1 cells certifiable at 2\,pp
\emph{falls} from 5.6\% to 4.9\% and the share showing a detectable difference
rises sharply from 17.5\% to 26.5\%, because the recovered pairs are larger-$n$
cells, which resolve more differences. Every S2 field is identical across revisions.

The headline verdicts did not move under the rev-2 population change. On the
own-margin reading then in force, since superseded by Amendment~2, the audit
stood at $K = 4$ of 12 determinate claims and $J = 5$ indeterminate, identical
to rev-1 under the rev-2 discordance imputation, with the uniform-2\,pp
secondary reading unchanged at 1 of 12. Those counts are quoted as the state of
the analysis at that date and are not current results: the eligible denominator
later changed with R10's exclusion, and the own-margin reading was withdrawn
outright. What survives is the finding that enlarging the atlas moved no
verdict. No new verdict computation was triggered. In the
certification tables no family entered or left, and \texttt{bbh}, \texttt{gpqa},
\texttt{ifeval}, \texttt{math}, \texttt{mmlu\_pro} and \texttt{musr} are
unchanged in both cell count and required $n$; the families that moved are
\texttt{mmlu} (798 $\to$ 1{,}311 cells, required $n$ 2{,}123 $\to$ 2{,}164),
\texttt{gsm8k} (11 $\to$ 24, 750 $\to$ 1{,}184), \texttt{winogrande}
(15 $\to$ 23, 1{,}879 $\to$ 1{,}416), \texttt{arc\_challenge} (8 $\to$ 17,
1{,}211 $\to$ 1{,}218), \texttt{hellaswag} (14 $\to$ 23, 688 $\to$ 695) and the
pooled row (1{,}155 $\to$ 1{,}707, 1{,}739 $\to$ 1{,}855).

That the audit's headline survived a 44\% enlargement of the atlas population is
worth stating plainly, and it is also worth stating that we did not know it
would when the repair was authorised.


\section{Full audit table}
\label{app:audit-table}

\begin{table}[!t]
\centering
\scriptsize
\caption{The 17 audited claims: identity and frame. \textbf{F1} is a research-paper claim, \textbf{F2} a vendor model card or official documentation page, \textbf{F3} a vendor blog post. Source URLs, version dates and the exact quoted sentence per claim are in the frozen claim table \texttt{docs/audit\_claim\_table.csv}, which is part of the released artifact.}
\label{tab:audit-identity}
\begin{tabular}{@{}llp{10.2cm}@{}}
\toprule
Claim & Frame & Source \\
\midrule
R01 & F1 & GPTQ: Accurate Post-Training Quantization for Generative Pre-trained Transformers (Frantar, Ashkboos, Hoefler, Alistarh) \\
R02 & F1 & LLM.int8(): 8-bit Matrix Multiplication for Transformers at Scale (Dettmers, Lewis, Belkada, Zettlemoyer) \\
R03 & F1 & SmoothQuant: Accurate and Efficient Post-Training Quantization for Large Language Models (Xiao, Lin, Seznec, Wu, Demouth, Han) \\
R04 & F1 & AWQ: Activation-aware Weight Quantization for LLM Compression and Acceleration (Lin, Tang, Tang, Yang, Chen, Wang, Xiao, Dang, Gan, Han) \\
R05 & F1 & SqueezeLLM: Dense-and-Sparse Quantization (Kim, Hooper, Gholami, Dong, Li, Shen, Mahoney, Keutzer) \\
R06 & F1 & A Simple and Effective Pruning Approach for Large Language Models [Wanda] (Sun, Liu, Bair, Kolter) \\
R07 & F1 & SparseGPT: Massive Language Models Can Be Accurately Pruned in One-Shot (Frantar, Alistarh) \\
R08 & F2 & RedHatAI/Meta-Llama-3.1-8B-Instruct-quantized.w4a16 model card (Neural Magic / Red Hat AI) \\
R09 & F2 & RedHatAI/Meta-Llama-3.1-8B-Instruct-FP8 model card (Neural Magic / Red Hat AI) \\
R10 & F2 & RedHatAI/Qwen2.5-7B-Instruct-quantized.w4a16 model card (Neural Magic / Red Hat AI) \\
R11 & F2 & Meta AI blog: ``Introducing quantized Llama models with increased speed and a reduced memory footprint'' \\
R12 & F3 & NVIDIA TensorRT-LLM official blog: ``Speed up inference with SOTA quantization techniques in TRT-LLM'' \\
R13 & F3 & vLLM official docs: ``FP8 W8A8'' (LLM Compressor quantization guide) \\
R14 & F3 & vLLM Blog: ``The State of FP8 KV-Cache and Attention Quantization in vLLM'' (Kübler, Kurtić, Wilkinson, Bonanni, Goin, Marques, Budhathoki) \\
R15 & F2 & RedHatAI/Meta-Llama-3.1-8B-Instruct-quantized.w8a8 (model card) \\
R16 & F2 & RedHatAI/Meta-Llama-3.1-70B-Instruct-quantized.w4a16 (model card) \\
R17 & F2 & RedHatAI/Meta-Llama-3-8B-Instruct-quantized.w8a16 (model card) \\
\bottomrule
\end{tabular}
\end{table}

\begin{table}[!tp]
\centering
\setlength{\abovecaptionskip}{0pt}
\begingroup\let\normalsize\footnotesize
\caption{Per-claim characterisation and sensitivity classification, at the registered 2\,pp margin. $n$ is the reported or imputed item count, with its basis recorded in the released CSV. $p_d$ is the discordance imputed from the atlas and \emph{tier} the match tier it came from. \emph{V3} is whether the source releases per-item outputs for the tasks its claim rests on. \emph{Classification} is evaluated across the interquartile range of the atlas cells supplying $p_d$: \emph{above throughout} and \emph{below throughout} hold at both quartiles, and \emph{sensitive} changes within the interval. R10 keeps its row and is marked ineligible rather than removed.}
\label{tab:audit-characterisation}
\endgroup
\setlength{\tabcolsep}{4pt}
\footnotesize
\textbf{(a) What the claim is, and what it measured}\par\smallskip
\begin{tabular}{@{}lllll@{}}
\toprule
Claim & Family & Bits & Benchmark & $n$ \\
\midrule
R01 & gptq & 4 & piqa & 1{,}838 \\
R02 & w8a8\_int8 & 8 & (mixed/unmatched) & --- \\
R03 & w8a8\_int8 & 8 & (mixed/unmatched) & 18{,}300 \\
R04 & awq & 4 & gsm8k & 1{,}319 \\
R05 & squeezellm & 4 & mmlu & 14{,}042 \\
R06 & pruning & --- & (mixed/unmatched) & 18{,}904 \\
R07 & pruning & --- & (mixed/unmatched) & 12{,}410 \\
R08 & w4a16 & 4 & mmlu & 42{,}701 \\
R09 & w8a8\_fp8 & 8 & mmlu & 42{,}701 \\
R10 & w4a16 & 4 & mmlu & 28{,}659 \\
R11 & spinquant & 4 & (mixed/unmatched) & --- \\
R12 & w8a8\_fp8 & 8 & mmlu & 14{,}042 \\
R13 & w8a8\_fp8 & 8 & gsm8k & 250 \\
R14 & w8a8\_fp8 & 8 & (mixed/unmatched) & 728 \\
R15 & w8a8\_int8 & 8 & mmlu & 42{,}701 \\
R16 & w4a16 & 4 & mmlu & 42{,}701 \\
R17 & w8a16 & 8 & mmlu & 28{,}659 \\
\bottomrule
\end{tabular}

\smallskip
\textbf{(b) How it was assessed, at the registered 2\,pp margin}\par\smallskip
\begin{tabular}{@{}lllll@{}}
\toprule
Claim & $p_d$ & Tier & V3 & Classification \\
\midrule
R01 & 0.130 & family+bits & no & \textbf{sensitive} \\
R02 & 0.056 & family+bits & no & not assessable (reporting) \\
R03 & 0.056 & family+bits & no & above throughout \\
R04 & 0.074 & family+bits+benchmark & no & not assessable (metric) \\
R05 & 0.146 & bits+benchmark & no & above throughout \\
R06 & 0.119 & bits & no & above throughout \\
R07 & 0.119 & bits & no & above throughout \\
R08 & 0.048 & family+bits & partial & above throughout \\
R09 & 0.048 & family+bits & no & above throughout \\
R10 & 0.048 & family+bits & no & \emph{ineligible} \\
R11 & 0.133 & bits & no & not assessable (reporting) \\
R12 & 0.048 & family+bits & no & above throughout \\
R13 & 0.048 & family+bits & no & not assessable (reporting) \\
R14 & 0.048 & family+bits & no & not assessable (reporting) \\
R15 & 0.056 & family+bits & partial & above throughout \\
R16 & 0.048 & family+bits & partial & above throughout \\
R17 & 0.135 & bits+benchmark & no & above throughout \\
\bottomrule
\end{tabular}
\end{table}

\begin{table}[!t]
\centering
\scriptsize
\caption{Power quantities per claim at the registered 2\,pp margin. MDD is the minimum detectable difference in percentage points, under the paired and the independent-binomial variance models. Required $n$ is the item count needed under TOST at one-sided $\alpha=.05$ (a 90\% two-sided confidence interval) with 80\% power, assuming a true difference of zero, under the paired model; it is given at the first quartile, median and third quartile of the atlas cells supplying the claim's discordance rate, which bracket the interval because required $n$ increases monotonically in that rate. $d^{*}$ is the discordance rate at which the classification would reverse; it falls inside the swept interval for R01 alone, and is blank where no reversal is attainable. \textbf{Numbers on ineligible and non-assessable rows are not verdicts}; they are retained so a reader can see what the analysis could and could not evaluate.}
\label{tab:audit-power}
\begin{tabular}{@{}lrrrrrrr@{}}
\toprule
Claim & $n$ & MDD (paired) & MDD (indep.) & $n_{\mathrm{req}}$@Q1 & @median & @Q3 & $d^{*}$ \\
\midrule
R01 & 1{,}838 & 2.36 & 3.62 & 1{,}364 & 2{,}010 & 4{,}328 & 0.1189 \\
R02 & --- & --- & --- & 619 & 866 & 1{,}237 & --- \\
R03 & 18{,}300 & 0.49 & 1.38 & 619 & 866 & 1{,}237 & --- \\
R04 & 1{,}319 & 2.09 & 3.77 & 1{,}137 & 1{,}137 & 1{,}137 & --- \\
R05 & 14{,}042 & 0.90 & 1.63 & 1{,}525 & 2{,}255 & 3{,}865 & 0.9085 \\
R06 & 18{,}904 & 0.70 & 1.36 & 932 & 1{,}841 & 6{,}046 & --- \\
R07 & 12{,}410 & 0.87 & 1.63 & 932 & 1{,}841 & 6{,}046 & 0.8029 \\
R08 & 42{,}701 & 0.30 & 0.84 & 371 & 742 & 1{,}052 & --- \\
R09 & 42{,}701 & 0.30 & 0.84 & 433 & 742 & 1{,}113 & --- \\
R10 & 28{,}659 & 0.36 & 1.04 & 371 & 742 & 1{,}052 & --- \\
R11 & --- & --- & --- & 1{,}314 & 2{,}051 & 3{,}682 & --- \\
R12 & 14{,}042 & 0.52 & 1.53 & 433 & 742 & 1{,}113 & 0.9085 \\
R13 & 250 & 3.88 & --- & 433 & 742 & 1{,}113 & --- \\
R14 & 728 & 2.27 & --- & 433 & 742 & 1{,}113 & --- \\
R15 & 42{,}701 & 0.32 & 0.84 & 619 & 866 & 1{,}237 & --- \\
R16 & 42{,}701 & 0.30 & 0.69 & 371 & 742 & 1{,}052 & --- \\
R17 & 28{,}659 & 0.61 & 1.09 & 1{,}332 & 2{,}081 & 3{,}644 & --- \\
\bottomrule
\end{tabular}
\end{table}

Table~\ref{tab:audit-identity} gives the claim-to-source mapping the
\texttt{R}\emph{nn} identifiers used throughout the paper refer to;
Table~\ref{tab:audit-characterisation} the per-claim characterisation and
sensitivity classification; Table~\ref{tab:audit-power} the power quantities.
They are split by content rather than presented as one sheet because the
released CSV has 53 columns, more than a page can carry legibly.

The released CSV carries, per claim: identity and frame; method family, bit
width and benchmark; the reported or imputed $n$ with its \texttt{n\_basis};
eligibility with its basis; the margin category, its basis and the form the
source's evidence takes; the imputed discordance with its match tier and the
number of atlas cells behind it, together with the first and third quartiles of
those cells; assessability status, kind and reason; the components that remain
computable for non-assessable claims; the V3 reproducibility value; V1 as
paired and independent-binomial MDD; V2 required-$n$ at 1, 2 and 3\,pp under
both variance models; the reversal discordance with the count and share of
supporting cells below it; the margin-sensitivity flag; and the threshold
classification with its robustness across the quartile interval.

Three groups of columns present in the superseded rev-2 release are absent here:
every quantity computed at a source's own reported delta, the applicable-margin
pair that selected between the two readings, and the verdict column that
recorded ``underpowered for its own assertion''. They were withdrawn with
Amendment~2 rather than recomputed, and the rev-2 file remains published as the
superseded record.

\paragraph{Reading the transparency columns.} Rows marked ineligible or
non-assessable carry numbers in some V1/V2 columns. Those numbers are \emph{not}
verdicts; they are retained so that a reader can see exactly what the analysis
could and could not evaluate. R14 is the case to watch: its imputed $n = 728$
sits just below the $742$ its row reports as required, which looks like a
threshold failure and is not one, because the missing baseline means there is no
paired comparison to size. R04's GSM8K computation is the largest such quantity
and is discussed in \S\ref{sec:audit:r04}.

\section{Audit method detail}
\label{app:audit-method}


\subsection{Where the equivalence claim is written}
\label{app:audit:locus}

\begin{table}[!t]
\centering
\caption{Where the equivalence claim is written, across six model cards from one
publisher with the underlying evidence held constant. Every card reports the
same recovery column; the tiers differ only in whether, and where, the card
states the comparison in words. Read by \S\ref{sec:audit:locus}.}
\label{tab:audit-locus}
\begin{tabular}{llc}
\toprule
Locus of the claim & Form it takes in the card & Cards \\
\midrule
Prose assertion      & states recovery percentages outright & 3 \\
Prose juxtaposition  & states two scores, characterises neither & 2 \\
Table cell only      & no comparative sentence anywhere & 1 \\
\bottomrule
\end{tabular}
\end{table}

The taxonomy of \S\ref{sec:audit:taxonomy} asks what form a source's assertion
takes. A second question turns out to matter as much: \emph{where in the
document the assertion is written}, and whether it is written in words at all.

Six of the eligible sources are quantized-model cards from a single publisher.
They run the same benchmark suite, present the same results table with the same
recovery column, declare no margin, and release no task-matched per-item
outputs. The evidence behind them is, for present purposes, identical. What
differs is where each one puts its equivalence claim
(Table~\ref{tab:audit-locus}): three cards assert recovery in prose, two report
the quantized and the unquantized score in adjacent clauses and stop,
characterising the gap neither as small nor as acceptable, and one makes no
comparative statement in prose at all, its claim existing solely as a recovery
figure in a table cell under a column header with no caption. That last card is
the excluded one (\AuditIneligibleClaim{}), and the exclusion is legible as the
endpoint of a gradient rather than an isolated extraction defect.

The registered inclusion rule is reasonable, and it is the same rule any
comparable audit would need. It is also keyed to a property that varies
independently of the evidence. A reader encountering all six cards would take
all six as near-lossless claims; an inclusion rule keyed to prose vocabulary
captures the first tier cleanly, must exercise judgement on the second, and
cannot see the third. The claims do not become weaker as the locus moves from
sentence to cell. They stop being written down.

The two prose-juxtaposition cards were kept. They carry no trigger vocabulary in
their prose, and the recovery figures that would qualify them appear only as
table cells, which is the position that excluded \AuditIneligibleClaim{}. They
were retained on the judgement that a bare juxtaposition of two scores differing
by less than a point, in a document whose purpose is to offer the quantized
model as a substitute for the unquantized one, is an equivalence assertion in
substance even where it is not one in vocabulary. Excluding them instead would
have moved the eligible denominator to 14 and the assessable count to 9 without
changing which claims fall below the planning threshold, and it would have moved
the headline proportion in the direction favourable to this paper's thesis. The
conservative choice is to keep them and report the ambiguity.

\subsection{Detection resolution per claim}
\label{app:audit:mdd}

\begin{table}[!t]
\centering
\caption{Detection resolution (V1) at the registered 2\,pp margin: the smallest
difference each evaluation could have detected, and its ratio to that margin. A
ratio below $1$ means the evaluation resolves differences finer than the margin.
The independent-binomial columns are the same quantities computed without the
paired-design benefit. Rows are the 11 assessable claims; the two italicised
rows carry no verdict and are shown only for transparency. The MDD columns
repeat those of Table~\ref{tab:audit-power}; the ratio columns appear only here.}
\label{tab:audit-mdd}
\begin{tabular}{lrrrr}
\toprule
      & \multicolumn{2}{c}{MDD} & \multicolumn{2}{c}{MDD / 2\,pp} \\
\cmidrule(lr){2-3}\cmidrule(lr){4-5}
Claim & paired & indep.\ binomial & paired & indep.\ binomial \\
\midrule
R01 & 2.36\,pp & 3.62\,pp & $1.18\times$ & $1.81\times$ \\
R05 & 0.90\,pp & 1.63\,pp & $0.45\times$ & $0.82\times$ \\
R07 & 0.87\,pp & 1.63\,pp & $0.43\times$ & $0.81\times$ \\
R06 & 0.70\,pp & 1.36\,pp & $0.35\times$ & $0.68\times$ \\
R17 & 0.61\,pp & 1.09\,pp & $0.30\times$ & $0.54\times$ \\
R12 & 0.52\,pp & 1.53\,pp & $0.26\times$ & $0.76\times$ \\
R03 & 0.49\,pp & 1.38\,pp & $0.25\times$ & $0.69\times$ \\
R15 & 0.32\,pp & 0.84\,pp & $0.16\times$ & $0.42\times$ \\
R16 & 0.30\,pp & 0.69\,pp & $0.15\times$ & $0.35\times$ \\
R09 & 0.30\,pp & 0.84\,pp & $0.15\times$ & $0.42\times$ \\
R08 & 0.30\,pp & 0.84\,pp & $0.15\times$ & $0.42\times$ \\
\midrule
\textit{(R04)} & \textit{2.09\,pp} & \textit{3.77\,pp} & \textit{$1.05\times$} & \textit{$1.89\times$} \\
\textit{(R14)} & \textit{2.27\,pp} & \textit{---} & \textit{$1.14\times$} & \textit{---\ (no baseline)} \\
\bottomrule
\end{tabular}
\end{table}

\subsection{The three registered verdict quantities}
\label{app:audit:verdictrules}

For each claim, at its reported or rule-imputed sample size $n$, the frozen
protocol computes three quantities.

\emph{V1, detection resolution.} The minimum detectable difference (MDD) at 80\%
power and two-sided $\alpha = 0.05$, under the paired-flip model. The per-item
accuracy difference is $d_i \in \{-1, 0, +1\}$; under the null of no true
difference, $\mathrm{Var}(d) = p_d$, the discordance rate, so
$\mathrm{sd} = \sqrt{p_d}$. Both the MDD and its ratio to the registered margin
are reported (Table~\ref{tab:audit-mdd}).

\emph{V2, equivalence support.} The number of items required for TOST at the
applicable margin,
\begin{equation}
n_{\mathrm{req}} \;=\; \left\lceil \left(\frac{(z_{1-\alpha} + z_{1-\beta})\,\mathrm{sd}}{m}\right)^{\!2} \right\rceil ,
\label{eq:tost-n}
\end{equation}
with $z_{1-\alpha}$ \emph{one-sided} ($1.6449$), because TOST rejects two
one-sided nulls at level $\alpha$ each. At $\alpha=.05$ per bound this is
operationally the requirement that a \textbf{90\% two-sided} confidence interval
fall inside $\pm m$, not a 95\% one; V1 above is the quantity that takes the
two-sided $z$. A claim is recorded as \emph{below the approximate planning
threshold} iff its reported $n$ is below $n_{\mathrm{req}}$ at the registered
margin.

\emph{V3, reproducibility.} Binary, read from the frozen
\texttt{per\_item\allowbreak\_outputs\allowbreak\_released} column and
cross-checked against the extraction reconciliation memo.

Every quantity is additionally recomputed under the independent-binomial bound
$\mathrm{sd} = \sqrt{2p(1-p)}$ and swept over 1\,pp and 3\,pp margins; a verdict
that changes across that sweep is \emph{margin-sensitive}.

\subsection{Discordance imputation}
\label{app:audit:imputation}

The discordance rate $p_d$ is not reported by any audited source, so it is
imputed from the atlas (\S\ref{sec:atlas}) by matching the nearest (method
family, bit width, benchmark) cell, most-specific tier first, taking the
\emph{median} over the first non-empty tier because per-cell discordance is
right-skewed. One claim matched at tier~1 (family + bits + benchmark), 11 at
tier~2 (family + bits), 2 at tier~3 (bits + benchmark), and 3 at tier~4 (bits);
\textbf{no claim reached the global tier}. A tier whose target field is
\texttt{None} cannot match, so a pruning claim with no bit width descends
automatically rather than being forced into a wrong cell. The two pruning claims
descend to tier~4 and match there, over 183 cells.

\paragraph{What the tier-4 match for R06 and R07 actually rests on.} That match
deserves stating plainly, because it is weaker than the tier label suggests.
R06 and R07 are 50\% unstructured \emph{pruning} claims with no bit width, and
the 183 cells they match are \texttt{bnb-4bit}, \texttt{QLoRA-4bit} and
\texttt{bnb-4bit(DPO-FT)} cells, all of which are 4-bit \emph{quantization}.
Those cells carry a null bit width only because their method labels are absent
from the parser's profile table, not because they lack one. The match is
therefore on a field being missing on both sides rather than on any similarity
of method, and a pruning claim is being assigned a discordance rate measured on
quantized models. We report it rather than repair it: the imputation rule is
registered, this appendix is not the place to change a registered rule after
results are visible, and neither claim's classification depends on it, since
both sit far above the threshold even at the third quartile. A reader who
distrusts the substitution should read R06 and R07 as resting on the weakest
imputation in the table.
atlas cells) are excluded from the imputation pool, per the atlas registration.

\subsection{The two robustness readings}
\label{app:audit:robustness}


The choice this appendix used to defend no longer exists. Amendment~2 settles
the applicable margin at the registered 2\,pp for every claim, because the
frozen text makes the claim-specific reading conditional on a source stating a
margin and none does. What was previously presented as a factor-of-four
interpretive gap was a comparison against quantities that were never margins,
and it is withdrawn rather than restated. The reasoning is at
Appendix~\ref{app:prereg:choices}.

The surviving sensitivities are two, and they are independent of each other.
\textbf{Margin sensitivity} asks whether a classification depends on where the
yardstick is set: R01's flips across the registered 1\,pp\,$\to$\,3\,pp sweep,
and R04 and R14 also flip but carry no classification to qualify.
\textbf{Imputation sensitivity} asks whether it depends on a quantity no source
reported: R01's flips across the interquartile range of the atlas cells
supplying its discordance rate, reversing at $d = 0.118915$, with 345 of the 792
supporting cells below that point. No other assessable claim is sensitive in
either direction, and R01 is the only claim sensitive in both. That coincidence
is worth stating precisely because it is a coincidence: the two properties
answer different questions and neither implies the other.

%
%

\subsection{Eligibility, and the full-text source review}
\label{app:audit:eligibility}

The eligibility correction rests on a full-text review of every one of the
\AuditFrozenCandidates{} archived sources, conducted after the first verdicts had
been computed and reported as such. Each source was searched complete, including
tables, captions, footnotes, appendices and reference lists, with model cards
searched as raw Markdown rather than as rendered pages.

The review found that one candidate's recorded quotation appears nowhere in its
source: the sentence had been composed from a table cell rather than quoted from
prose. The registered inclusion rule requires the assertion to appear in prose or
a table caption. It appears in neither, so \AuditIneligibleClaim{} is excluded by
applying the registered rule rather than by a new judgement, and the eligible
denominator moves from \AuditFrozenCandidates{} to \AuditEligible{}. The
exclusion cannot flatter any result reported in \S\ref{sec:audit}:
\AuditIneligibleClaim{} was above the planning threshold at
\AuditMarginPP{}\,pp and recorded \emph{no} on per-item outputs, so removing it
lowers neither the threshold tally nor the reproducibility count. Its row remains
in the released claim table and in Table~\ref{tab:audit-characterisation}.

Source provenance: content hashes were recorded for 16 of the
\AuditFrozenCandidates{} sources. R13's was left empty, and the consequence for
provenance is stated in \S\ref{sec:artifacts}. Every archived source, its hash,
and the retrieval script needed to reproduce these locations are released with
the paper, so every classification here is checkable without re-fetching
anything.

\subsection{The complete-text margin sweep}
\label{app:audit:sweep}

The finding that no source declares a prospective numerical equivalence margin is
established over complete source text rather than over the quoted sentence. The
registered vocabulary of tolerance language was swept across all
\AuditFrozenCandidates{} archived sources in full.

The terms \emph{parity} and \emph{percentage point} do not occur anywhere in the
corpus. Every occurrence of \emph{margin} is a page-layout artefact, a stylesheet
declaration, or the idiom ``by a large margin'' describing superiority rather
than tolerance. One vendor document comes closest to a declared tolerance by
explicitly declining to fix one, observing that ``users might have different
tolerances on accuracy impact'', which states the absence of a threshold rather
than a threshold.

\subsection{Per-item outputs: what the three partial sources release}
\label{app:audit:v3detail}

\AuditPerItemOtherTaskClaims{}, all Red~Hat AI model cards, are the closest the
record comes to reproducibility, and they illustrate why the task-matched count
is nonetheless \AuditPerItemTaskMatched{}. They release per-item outputs for
Arena-Hard, OpenLLM~v2 and HumanEval, but \emph{not} for the OpenLLM~v1 tasks
that the audited equivalence claim is actually about. The released artifacts and
the asserted claim do not intersect. These three sources are ahead of the field
in disclosure practice and the correction they need is small: publish the
per-item outputs for the suite the claim quotes.

The reason this finding is more actionable than any power calculation is that the
two failures have different repair costs. Underpowering is fixable after the fact
by evaluating more items, and \S\ref{sec:certification} says how many.
Irreproducibility is not fixable downstream at all: with no per-item outputs,
nobody outside the releasing organisation can run the paired test at \emph{any}
sample size, cannot compute churn, and cannot check the arithmetic. Per-item
outputs for a benchmark of a few thousand items are a file of a few megabytes.

\subsection{The non-assessable claims, one by one}
\label{app:audit:indeterminate}

Each non-assessable claim is recorded with exactly one primary blocker.

\emph{Insufficient reporting} (\AuditNotAssessableInsufficient{}). R02
(LLM.int8()) and R11 (Meta's quantized-Llama blog) state no sample size, no
baseline and no numeric delta, and their headline equivalence evidence exists
only as a chart image. R13 (vLLM FP8 documentation) states $n = 250$ but shows no
on-page baseline run at all. R14 (vLLM FP8 KV-cache blog) reports an observed
0.7\,pp difference and permits $n$ to be imputed, but reports no baseline; its
comparison lives in a figure.

\emph{Outside the registered calculation}
(\AuditNotAssessableOutsideFramework{}). \AuditOutsideFrameworkClaim{} (AWQ)
reports enough, but about a quantity the registered binary paired-outcome model
cannot score (\S\ref{sec:audit:r04}).

Every non-assessable claim retains whatever components its available inputs
support. These are listed in the CSV column \texttt{determinate\_components} and
are reported as supplementary transparency only, \textbf{never verdict-bearing}:
R13 retains V2, because the paired standard deviation depends on discordance and
not on baseline accuracy; R14 retains V1 (paired) and V2; and R04 retains V1 and
V2 computed on a substituted benchmark.

\subsection{R01: the planning values behind the single flag}
\label{app:audit:r01}

\AuditSensitiveClaim{}, the GPTQ paper, reports $n = \AuditSensitiveN{}$ against
a requirement of $\AuditSensitiveNReq{}$ items at the imputed discordance rate of
$0.130$. The gap is one part in twelve of that requirement, and it closes
entirely if the imputed rate is a little lower: the classification reverses at a
discordance rate of $0.1189$, and $\AuditSensitiveCellsBelow{}$ of the
$\AuditSensitiveCellsTotal{}$ atlas cells supplying the imputation lie below that
point. Across the interquartile range of those cells the requirement moves from
$1{,}364$ items to $4{,}328$, so \AuditSensitiveClaim{} is adequately sized at
the first quartile and undersized at the third.

\subsection{R04: the overruled first-pass computation}
\label{app:audit:r04detail}

The first extraction pass scored R04 on GSM8K, the source's own accuracy
benchmark ($-0.30$\,pp at $n = 1{,}319$), and reported it as the audit's largest
shortfall. On review this was overruled, because computing a TOST requirement on
GSM8K audits a sentence the source wrote about a different benchmark, in
different and non-trigger language.

That ruling moved the headline and removed the audit's biggest number. The counts
and the ratio it moved are quoted as history in
Appendix~\ref{app:prereg:choices}, computed under the superseded reading that
took each source's reported delta as its margin, and \textbf{they are not current
results}. The GSM8K computation is retained in the released CSV as a labelled
transparency column so a reader can see exactly what was set aside and why. It is
not claimed anywhere in this paper as an audit result.

\subsection{Interpretive rulings and superseded verdicts}
\label{app:audit:history}

The audit's interpretive history is recorded in full elsewhere and is indexed
here so the main text needs only one pointer.

The four interpretive rulings, each with the direction it moved the headline, are
Appendix~\ref{app:prereg:choices}: which margin is ``the applicable margin'' and
the $K = 5 \to K = 4$ correction; the R04 benchmark-substitution ruling; the
claim-level rather than claim-by-benchmark granularity; and the treatment of
sources that state no sample size.

The superseded verdict record is \texttt{docs/AUDIT\_\allowbreak{}VERDICTS\_\allowbreak{}2026-07-20.md}, which
predates both the rev-2 atlas and Amendment~2. \textbf{No value may be restored
from it.} Amendment~2 (signed, commit \texttt{ab279b2}) makes the registered
uniform \AuditMarginPP{}\,pp margin the primary yardstick for every claim,
forbids describing any quantity derived from a source's own reported results as
that source's declared margin, and reopens \S\S3.1--3.2 for eligibility only. The
claim-derived margins and the shortfall range computed under the superseded
reading are \textbf{withdrawn, not recomputed}, and are non-verdict-bearing
wherever they still appear as history.

\subsection{Both robustness directions in full}
\label{app:audit:fullrobustness}

Two features of Table~\ref{tab:audit-mdd} matter for how the audit should be
read.

First, the independent-binomial columns are uniformly \emph{worse}, by roughly a
factor of two. Pairing is the generous modelling assumption, and the one
evaluation that is coarser than the registered margin under pairing is coarser
still without it.

Second, the ordering makes the audit's reach visible. Only
\AuditSensitiveClaim{}'s evaluation is coarser than the registered margin, and
the other ten resolve differences between a sixth and a half of it. An audit at a
\AuditMarginPP{}\,pp margin is a weak test and should be understood as one. The
sources were not asked to meet a standard this permissive, because they were not
asked to meet any stated standard at all, and what the table shows is that ten
evaluations would already satisfy a reporting requirement none of them was given.

\section{Certification tables at 1\,pp and 3\,pp}
\label{app:certification-margins}

The 2\,pp margin is Table~\ref{tab:certification} in the main text; the
1\,pp margin is Table~\ref{tab:certification-1pp} and the 3\,pp margin
Table~\ref{tab:certification-3pp}. Note the quadratic margin scaling: MMLU's median
requirement is 8{,}656 items at 1\,pp, 2{,}164 at 2\,pp, and 962 at 3\,pp, a
factor of nine across a three-fold change in margin.

\begin{table}[!t]
\centering
\small
\caption{Items required to certify equivalence within $\pm 1$\,pp under TOST at one-sided $\alpha=.05$ (a 90\% two-sided interval) with 80\% power, assuming a true difference of zero, by benchmark family, at the 25th/50th/75th percentiles of the discordance rates the atlas observes for that family. The naive column is the same requirement computed by treating the two runs as independent samples. Eleven benchmark families plus the pooled row. Row order matches Table~\ref{tab:certification} so the three margins compare row-by-row.}
\label{tab:certification-1pp}
\setlength{\tabcolsep}{2pt}
\small
\begin{tabular}{lr>{\centering\arraybackslash}p{2.6cm}>{\centering\arraybackslash}p{2.65cm}rr}
\toprule
Benchmark & Atlas cells & Discordance p25/med/p75 & Required $n$ p25/\textbf{med}/p75 & Naive & Advantage \\
\midrule
musr & 24 & 0.015 / 0.034 / 0.044 & 928 / \textbf{2{,}076} / 2{,}721 & 30{,}468 & $14.7\times$ \\
gpqa & 24 & 0.035 / 0.048 / 0.067 & 2{,}180 / \textbf{2{,}993} / 4{,}138 & 28{,}931 & $9.7\times$ \\
bbh & 192 & 0.024 / 0.044 / 0.060 & 1{,}484 / \textbf{2{,}721} / 3{,}710 & 25{,}967 & $9.5\times$ \\
mmlu\_pro & 5 & 0.048 / 0.053 / 0.059 & 2{,}955 / \textbf{3{,}305} / 3{,}649 & 30{,}778 & $9.3\times$ \\
hellaswag & 23 & 0.031 / 0.045 / 0.099 & 1{,}905 / \textbf{2{,}777} / 6{,}151 & 19{,}618 & $7.1\times$ \\
arc\_challenge & 17 & 0.072 / 0.079 / 0.099 & 4{,}447 / \textbf{4{,}870} / 6{,}141 & 29{,}451 & $6.0\times$ \\
ifeval & 8 & 0.048 / 0.052 / 0.072 & 2{,}943 / \textbf{3{,}200} / 4{,}429 & 16{,}842 & $5.3\times$ \\
mmlu & 1{,}311 & 0.093 / 0.140 / 0.259 & 5{,}725 / \textbf{8{,}656} / 16{,}019 & 30{,}906 & $3.6\times$ \\
winogrande & 23 & 0.067 / 0.092 / 0.221 & 4{,}124 / \textbf{5{,}661} / 13{,}688 & 22{,}397 & $4.0\times$ \\
math & 56 & 0.107 / 0.141 / 0.169 & 6{,}642 / \textbf{8{,}741} / 10{,}439 & 20{,}888 & $2.4\times$ \\
gsm8k & 24 & 0.040 / 0.077 / 0.198 & 2{,}473 / \textbf{4{,}735} / 12{,}270 & 10{,}684 & $2.3\times$ \\
\midrule
\textbf{ALL (pooled)} & \textbf{1{,}707} & 0.064 / 0.120 / 0.225 & 3{,}974 / \textbf{7{,}420} / 13{,}909 & 30{,}887 & $\mathbf{4.2\times}$ \\
\bottomrule
\end{tabular}
\end{table}

\begin{table}[!t]
\centering
\small
\caption{Items required to certify equivalence within $\pm 3$\,pp under TOST at one-sided $\alpha=.05$ (a 90\% two-sided interval) with 80\% power, assuming a true difference of zero, by benchmark family, at the 25th/50th/75th percentiles of the discordance rates the atlas observes for that family. The naive column is the same requirement computed by treating the two runs as independent samples. Eleven benchmark families plus the pooled row. Row order matches Table~\ref{tab:certification} so the three margins compare row-by-row.}
\label{tab:certification-3pp}
\setlength{\tabcolsep}{2pt}
\small
\begin{tabular}{lr>{\centering\arraybackslash}p{2.6cm}>{\centering\arraybackslash}p{2.65cm}rr}
\toprule
Benchmark & Atlas cells & Discordance p25/med/p75 & Required $n$ p25/\textbf{med}/p75 & Naive & Advantage \\
\midrule
musr & 24 & 0.015 / 0.034 / 0.044 & 104 / \textbf{231} / 303 & 3{,}386 & $14.7\times$ \\
gpqa & 24 & 0.035 / 0.048 / 0.067 & 243 / \textbf{333} / 460 & 3{,}215 & $9.7\times$ \\
bbh & 192 & 0.024 / 0.044 / 0.060 & 165 / \textbf{303} / 413 & 2{,}886 & $9.5\times$ \\
mmlu\_pro & 5 & 0.048 / 0.053 / 0.059 & 329 / \textbf{368} / 406 & 3{,}420 & $9.3\times$ \\
hellaswag & 23 & 0.031 / 0.045 / 0.099 & 212 / \textbf{309} / 684 & 2{,}180 & $7.1\times$ \\
arc\_challenge & 17 & 0.072 / 0.079 / 0.099 & 495 / \textbf{542} / 683 & 3{,}273 & $6.0\times$ \\
ifeval & 8 & 0.048 / 0.052 / 0.072 & 327 / \textbf{356} / 493 & 1{,}872 & $5.3\times$ \\
mmlu & 1{,}311 & 0.093 / 0.140 / 0.259 & 637 / \textbf{962} / 1{,}780 & 3{,}434 & $3.6\times$ \\
winogrande & 23 & 0.067 / 0.092 / 0.221 & 459 / \textbf{629} / 1{,}521 & 2{,}489 & $4.0\times$ \\
math & 56 & 0.107 / 0.141 / 0.169 & 738 / \textbf{972} / 1{,}160 & 2{,}321 & $2.4\times$ \\
gsm8k & 24 & 0.040 / 0.077 / 0.198 & 275 / \textbf{527} / 1{,}364 & 1{,}188 & $2.2\times$ \\
\midrule
\textbf{ALL (pooled)} & \textbf{1{,}707} & 0.064 / 0.120 / 0.225 & 442 / \textbf{825} / 1{,}546 & 3{,}432 & $\mathbf{4.2\times}$ \\
\bottomrule
\end{tabular}
\end{table}
%

\subsection{Certification method detail}
\label{app:cert:method}

\paragraph{The two standard deviations.} The paired form follows from the flip
model: the per-item accuracy difference is $d_i \in \{-1,0,+1\}$, and under the
null of no true difference $\mathrm{Var}(d) = p_d$, the rate at which the two
models disagree on correctness. The independent form uses $p$, the family's
median baseline accuracy, because independent-binomial variance depends on
accuracy and not on churn. That is the whole of the difference between the two
columns of Table~\ref{tab:certification}, and it is why the advantage varies by
benchmark family rather than being a constant.

\paragraph{The one-sided $z$.} $z_{1-\alpha}$ in Equation~\eqref{eq:nreq} is
one-sided, since TOST rejects two one-sided nulls at level $\alpha$ each, which
is why $\alpha = .05$ per bound corresponds to containment of a 90\% two-sided
interval rather than a 95\% one (interpretive choice~5,
Appendix~\ref{app:prereg:choices}). The detection quantity V1 of
\S\ref{sec:audit:rules} is the one that takes the two-sided $z$.

\paragraph{Where the discordance rates come from.} They are not modelled; they
are read off the atlas (\S\ref{sec:atlas}). For each benchmark family we take
the 25th percentile, median and 75th percentile of the per-cell accuracy-state
churn observed across that family's atlas cells, giving optimistic, typical and
pessimistic compression behaviour. Quartiles are \texttt{numpy}'s
linear-interpolation \texttt{np.quantile}.

\paragraph{Why the independent-binomial column is not a straw man.} It is what
you get by treating the baseline and compressed evaluations as two unrelated
samples and comparing proportions, which is the default in most reporting, and
it is wrong in a specific, quantifiable way: the two runs are \emph{the same
items through two nearly identical models}, so they agree on the large majority
of items and their difference has far less variance than independence implies.
The variation in the advantage across families is informative in itself:
low-churn families (MuSR, BBH, GPQA) reward pairing most, while high-churn
generative families (MATH, MMLU) both need more items \emph{and} gain less from
pairing.

\paragraph{The two thin rows.} Of the families that survive the four-cell
threshold, two rest on thin evidence and are indicative only,
\texttt{mmlu\_pro} (5 cells) and \texttt{ifeval} (8). They should be treated as
a starting hypothesis to be replaced by the reader's own measured churn, not as
reference constants.

\subsection{Scope caveats carried from the main text}
\label{app:cert:caveats-detail}


\paragraph{Cell counts behind each family.} Rev-2 moved
\texttt{arc\_challenge} to 17 cells, \texttt{gsm8k} to 24, \texttt{hellaswag}
to 23 and \texttt{winogrande} to 23, taking those out of the thin category;
\texttt{mmlu} (1{,}311), \texttt{bbh} (192) and \texttt{math} (56) are well
supported. Only \texttt{mmlu\_pro} (5) and \texttt{ifeval} (8) remain thin, as
the main text states.

\paragraph{Family aggregation mixes two sources of variation.} The atlas
collapses MMLU's 57 per-subject cells and BBH/MATH/MuSR/GPQA's per-subtask cells
into families, so a family's spread combines subject-level with model-level
variation. This widens the p25--p75 band relative to what a single practitioner
evaluating a single model would see, and therefore makes the quartile columns
\emph{conservative} rather than optimistic.

\paragraph{Feasibility-probe pairs are excluded.} Both disclosed
feasibility-probe pairs (99 cells) are excluded per the atlas registration §6,
as tiny hand-built sanity pairs ($n$ as low as 10, discordance up to
0.9) that would distort every quartile.


%

\section{Atlas construction detail}
\label{app:atlas-detail}

\subsection{The two sources}
\label{app:atlas:sources}

\textbf{S1} is the Open LLM Leaderboard v1 archive: community quantizations
(GPTQ, AWQ, GGUF, 8-bit and 4-bit bitsandbytes) of 2023-era base models, paired
with their base model's details dataset whenever one exists. \textbf{S2} is the
Neural Magic / Red~Hat per-item dumps for quantized Llama-3.1, covering W4A16,
W8A8-INT8 and W8A8-FP8 at 8B, 70B and 405B.

The registration exists to prevent source- or pair-selection after seeing
results, and it labels these analyses descriptive: they estimate flip and churn
magnitudes in the wild and feed the certification tables, they test no registered
hypothesis, and they cannot substitute for any H3 cell.

\subsection{Item pairing}
\label{app:atlas:pairing}

Item pairing is mechanical: items join on the source's own item key, duplicated
keys are dropped entirely on both sides, and an item enters the paired analysis
only if its full-prompt hash is identical across the pair. A pair-task cell is
excluded if fewer than 99\% of joinable items pass that identity check. Differing
harness commits do not exclude a cell, since prompt-hash identity is the operative
control, but are recorded per cell and disclosed.

\subsection{Exclusion breakdown}
\label{app:atlas:exclusions}

The enumeration yields \textbf{2{,}055} pair-task cells, of which
\textbf{1{,}807 are analysed} (S1 = 1{,}459, S2 = 348) and \textbf{248} are
excluded or skipped. The exclusions break down as \textbf{179} cells (72.2\%)
for which no results file exists for that task in any recorded run,
\textbf{36} (14.5\%) whose join intersection is empty, and \textbf{33}
(13.3\%) whose task carries no binary correctness metric in the data at all
(genuinely float-scored tasks, which the flip model does not describe).

The full exclusion table ships with the release, one row per excluded cell with
its reason, so the admitted population is auditable rather than asserted.

The 99 disclosed probe cells are removed for a reason the body states but does
not quantify: they are hand-built sanity pairs with item counts as low as
$n = 10$ and discordance rates as high as $0.9$, so admitting them would move
any quartile of the discordance distribution that the certification tables of
\S\ref{sec:certification} are computed from.

\subsection{The most extreme zero-delta cell}
\label{app:atlas:extreme}

\begin{table}[!t]
\centering
\caption{The most extreme zero-delta cell in the atlas: identical accuracy,
symmetric flips, $p = 1.0$. Read by \S\ref{sec:atlas:identical}, which carries
the caveats that travel with it.}
\label{tab:identical-extreme}
\begin{tabular}{l>{\raggedright\arraybackslash}p{7.6cm}}
\toprule
Field & Value \\
\midrule
pair\_index / source & 35 / S1 \\
Task                 & \texttt{harness\_\allowbreak{}hendrycksTest\_\allowbreak{}high\_\allowbreak{}school\_\allowbreak{}geography\_\allowbreak{}5} (MMLU) \\
Base model           & \texttt{project-baize/baize-v2-7b} \\
Quantized model      & \texttt{TheBloke/\allowbreak{}Project-Baize-v2-7B-GPTQ} (GPTQ) \\
$n$                  & 198 \\
Baseline accuracy    & 0.429293 \\
Compressed accuracy  & 0.429293 \quad (net delta $= 0.000000$) \\
Accuracy-state churn & \textbf{0.343434} \\
Harmful / beneficial flips & 0.171717 / 0.171717 \\
Exact McNemar $p$    & 1.0 \\
\bottomrule
\end{tabular}
\end{table}

\section{Controlled experiment: supporting detail}
\label{app:minigrid-detail}

\subsection{The mechanical escalation screen}
\label{app:escalation}

The four executed cells, $\{$Qwen2.5-1.5B-Instruct, Llama-3.2-3B-Instruct$\}
\times\{$MMLU, GSM8K$\}$, 4-bit GPTQ against 4-bit AWQ at seeds
$\{0,1,2,3,4\}$, with GPTQ and AWQ receiving byte-identical calibration samples
at each seed (\S\ref{sec:minigrid}), feed a pre-committed screen that decides
only one thing: whether the deferred 7B/8B cells are built. The screen is
mechanical and was frozen with the mini-grid registration before any of these
accuracies existed.

\paragraph{The rule, quoted verbatim (mini-grid registration \S3, frozen
2026-07-15).}
\begin{quote}
Compute, per the frozen algebra of \texttt{PREREGISTRATION.md}, for each of the
4 cells at 4-bit: winner flips across seeds (ties counted separately, per the
registered tie rule) and the range/gap criterion
$\max(\mathrm{range}_{\mathrm{GPTQ}}, \mathrm{range}_{\mathrm{AWQ}}) \geq
\mathrm{gap}$.

\emph{Escalate} to the deferred 7B/8B seed cells iff: winner flips occur in
\textbf{at least 1 of the 4 cells}, \textbf{or} the range/gap criterion holds in
\textbf{at least 2 of the 4 cells}.

If neither condition holds, the 7B/8B cells are not built, and no other result
(3-bit, other benchmarks, atlas findings) can substitute to trigger escalation.
\end{quote}

The winner-flip, $\mathrm{gap}$ and $\mathrm{range}_m$ definitions are the frozen
algebra recapped in \S\ref{sec:minigrid}: a winner flip needs two registered
seeds with opposite-signed, both-nonzero $d_s$, and exact ties are reported
separately and never count as flips.

\paragraph{Per-cell screen.}
Table~\ref{tab:minigrid-escalation} states each cell's outcome exactly as the
signed decision record does. $\mathrm{gap}$ is the absolute mean method
difference and $\max\mathrm{range}$ is
$\max(\mathrm{range}_{\mathrm{GPTQ}},\mathrm{range}_{\mathrm{AWQ}})$; the
range/gap criterion holds when the latter is at least the former. No cell
contained an exact tie.

\begin{table}[!t]
\centering
\small
\caption{The mechanical escalation screen over the four completed mini-grid
cells. Outcomes are reproduced verbatim from the signed decision record; the
per-seed accuracies that produce them are not restated here. ``Winner flip''
and ``range/gap holds'' are the two \S3 predicates; the counts beneath drive the
decision.}
\label{tab:minigrid-escalation}
\setlength{\tabcolsep}{4pt}
\small
\begin{tabular}{ll rr >{\raggedright\arraybackslash}p{2.4cm}}
\toprule
Cell & Winner flip & $\mathrm{gap}$ & $\max\mathrm{range}$ & Range/gap holds \\
\midrule
Qwen2.5-1.5B / MMLU  & \textbf{TRUE}  & 0.012292 & 0.040521 & \textbf{TRUE} \\
Qwen2.5-1.5B / GSM8K & FALSE          & 0.096800 & 0.033000 & FALSE \\
Llama-3.2-3B / MMLU  & FALSE          & 0.030922 & 0.063809 & \textbf{TRUE} \\
Llama-3.2-3B / GSM8K & \textbf{TRUE}  & 0.017800 & 0.034000 & \textbf{TRUE} \\
\midrule
\multicolumn{5}{l}{Winner flip in \textbf{2 of 4} cells (threshold $\geq 1$: met)} \\
\multicolumn{5}{l}{Range/gap holds in \textbf{3 of 4} cells (threshold $\geq 2$: met)} \\
\bottomrule
\end{tabular}
\end{table}

\paragraph{Outcome.}
A winner flip occurs in \textbf{2 of 4} cells (Qwen2.5-1.5B/MMLU and
Llama-3.2-3B/GSM8K), clearing the ``$\geq 1$'' branch, and the range/gap
criterion holds in \textbf{3 of 4} cells, clearing the ``$\geq 2$'' branch. Both
branches of the disjunction are independently satisfied, so the screen fires:
\textbf{escalation is triggered}, and the record was signed on 2026-07-23,
authorizing construction and evaluation of the deferred 7B/8B seed cells
(Qwen2.5-7B-Instruct, Llama-3.1-8B-Instruct $\times$ MMLU, GSM8K).

\paragraph{The screen ran on sealed cells.}
The four cells stayed sealed, with no accuracy read by anyone, until the registered
validator passed over the complete 44-JSONL expected set (job \texttt{11375247},
409/409 checks), which is the first mini-grid accuracy inspection the
registration permits. The FP16 gate failure that preceded the run was resolved
\emph{before} any inspection, the gates being re-derived under Amendment~3
(committed \texttt{bd565bd}), so the screen fired on a grid whose reference
gates were already settled rather than tuned to its result.

\paragraph{This screen is not the H3 verdict.}
Escalation is a screening decision about \emph{which cells to build}, not a
conclusion about H3. The confirmatory verdict is the frozen
Supported/Disconfirmed/Inconclusive rule applied once, mechanically, over all
eight cells, this grid plus the 7B/8B cells, and it is defined only when all
eight exist. None of the 7B/8B cells existed when the screen fired, so no verdict
follows from it. That the screen fired says the registered rule judged these four
cells to warrant the confirmatory eight; it said nothing, in either direction,
about what the eight-cell rule would return. The verdict it authorized the
construction of is reported in \S\ref{sec:minigrid:verdict}, and the four cells
above enter it unchanged: re-deriving them inside the eight-cell analysis
reproduced every quantity in Table~\ref{tab:minigrid-escalation} exactly, to the
six decimals published here, which is a regression test of the analysis code
against a signed known answer rather than a restatement of it.

\subsection{Per-seed differences behind the winner-flip column}
\label{app:minigrid:ds}

Table~\ref{tab:h3-ds} gives the per-seed differences $d_s$ behind the
winner-flip column of Table~\ref{tab:h3-eightcell}, so every flip determination
is checkable by inspection of sign.

\begin{table}[!t]
\centering
\small
\caption{Per-seed differences $d_s = \mathrm{acc}_{\mathrm{GPTQ},s} -
\mathrm{acc}_{\mathrm{AWQ},s}$ on byte-identical calibration samples. A winner
flip is two seeds of opposite, nonzero sign; the five cells with a flip are
those whose row is not sign-constant. No entry is zero, which is the 0-of-40 tie
count stated in \S\ref{sec:minigrid:verdict}.}
\label{tab:h3-ds}
\setlength{\tabcolsep}{2pt}
\small
\begin{tabular}{llrrrrr}
\toprule
Model & Task & $d_0$ & $d_1$ & $d_2$ & $d_3$ & $d_4$ \\
\midrule
Qwen2.5-1.5B & MMLU  & $-0.029127$ & $-0.014029$ & $+0.012463$ & $-0.027631$ & $-0.003133$ \\
Qwen2.5-1.5B & GSM8K & $-0.111000$ & $-0.098000$ & $-0.080000$ & $-0.088000$ & $-0.107000$ \\
Llama-3.2-3B & MMLU  & $-0.008973$ & $-0.090514$ & $-0.023287$ & $-0.015026$ & $-0.016807$ \\
Llama-3.2-3B & GSM8K & $+0.021000$ & $+0.015000$ & $-0.005000$ & $+0.023000$ & $+0.035000$ \\
Qwen2.5-7B   & MMLU  & $+0.006694$ & $+0.009400$ & $+0.001638$ & $+0.001496$ & $+0.012107$ \\
Qwen2.5-7B   & GSM8K & $+0.022000$ & $+0.016000$ & $-0.023000$ & $+0.002000$ & $+0.012000$ \\
Llama-3.1-8B & MMLU  & $-0.016237$ & $-0.026777$ & $-0.047215$ & $+0.018302$ & $-0.013887$ \\
Llama-3.1-8B & GSM8K & $+0.023000$ & $-0.015000$ & $-0.004000$ & $-0.036000$ & $-0.034000$ \\
\bottomrule
\end{tabular}
\end{table}

\subsection{The three registered supporting analyses}
\label{app:minigrid:supporting}

The three analyses were run once per cell at the registered parameters
(2{,}000 bootstrap replicates, RNG seed 0) as part of the same job that produced
the verdict, and \S\ref{sec:minigrid:supporting} reads them. None of them
modifies the verdict, and none can.

\begin{table}[!tp]
\centering
\footnotesize
\setlength{\abovecaptionskip}{0pt}
\begingroup\let\normalsize\footnotesize
\caption{The three registered supporting analyses over the eight confirmatory
cells. \textbf{(a)}~Variance components, reported separately as registered: SD
is across the five calibration seeds, SE is the item-level standard error within
a cell, and MMLU cells run at $n = 14{,}042$ against GSM8K's $n = 1{,}000$,
which is what drives the SE columns. \textbf{(b)}~Two-level paired bootstrap,
2{,}000 replicates at RNG seed 0, both registered; tie replicates are reported
separately and are included in the rank-flip denominator, per the registered
convention, and the final column repeats the registered winner-flip outcome,
which is a different question and not a consistency check.
\textbf{(c)}~Flip statistics, GPTQ against AWQ, means over the five paired
seeds: net delta is
$\mathrm{acc}_{\mathrm{AWQ}} - \mathrm{acc}_{\mathrm{GPTQ}}$, harmful and
beneficial are the two directions of correctness change, accuracy-state churn is
their sum, and total answer churn adds items that change answer without changing
correctness state.}
\label{tab:h3-supporting}
\endgroup

\textbf{(a) Variance components, kept separate}\par\smallskip
\begin{tabular}{llrrrr}
\toprule
& & \multicolumn{2}{c}{Seed-level SD} & \multicolumn{2}{c}{Item-level SE} \\
\cmidrule(lr){3-4}\cmidrule(lr){5-6}
Model & Task & GPTQ & AWQ & GPTQ & AWQ \\
\midrule
Qwen2.5-1.5B & MMLU  & 0.019130 & 0.006897 & 0.004215 & 0.004212 \\
Qwen2.5-1.5B & GSM8K & 0.006340 & 0.015116 & 0.015796 & 0.015658 \\
Llama-3.2-3B & MMLU  & 0.024925 & 0.011157 & 0.004214 & 0.004216 \\
Llama-3.2-3B & GSM8K & 0.004615 & 0.012300 & 0.015154 & 0.015308 \\
Qwen2.5-7B   & MMLU  & 0.006224 & 0.004537 & 0.003983 & 0.004002 \\
Qwen2.5-7B   & GSM8K & 0.019267 & 0.014195 & 0.013815 & 0.013922 \\
Llama-3.1-8B & MMLU  & 0.015100 & 0.010520 & 0.004183 & 0.004163 \\
Llama-3.1-8B & GSM8K & 0.014307 & 0.010654 & 0.013874 & 0.013640 \\
\bottomrule
\end{tabular}

\smallskip
\textbf{(b) Two-level paired bootstrap}\par\smallskip
\setlength{\tabcolsep}{2pt}\footnotesize
\begin{tabular}{llrlrll}
\toprule
Model & Task & Rank-flip rate & Replicates & Tie rate & Ties & Winner flip \\
\midrule
Qwen2.5-1.5B & MMLU  & 0.0445 & 89 / 2000  & 0.0005 & 1 / 2000  & \textbf{TRUE} \\
Qwen2.5-1.5B & GSM8K & 0.0000 & 0 / 2000   & 0.0000 & 0 / 2000  & FALSE \\
Llama-3.2-3B & MMLU  & 0.0000 & 0 / 2000   & 0.0000 & 0 / 2000  & FALSE \\
Llama-3.2-3B & GSM8K & 0.0220 & 44 / 2000  & 0.0000 & 0 / 2000  & \textbf{TRUE} \\
Qwen2.5-7B   & MMLU  & 0.0025 & 5 / 2000   & 0.0000 & 0 / 2000  & FALSE \\
Qwen2.5-7B   & GSM8K & 0.2575 & 515 / 2000 & 0.0100 & 20 / 2000 & \textbf{TRUE} \\
Llama-3.1-8B & MMLU  & 0.0405 & 81 / 2000  & 0.0000 & 0 / 2000  & \textbf{TRUE} \\
Llama-3.1-8B & GSM8K & 0.1260 & 252 / 2000 & 0.0030 & 6 / 2000  & \textbf{TRUE} \\
\bottomrule
\end{tabular}

\smallskip
\textbf{(c) Flip statistics, GPTQ against AWQ}\par\smallskip
\setlength{\tabcolsep}{2.5pt}
\footnotesize
\begin{tabular}{llrrrrrr}
\toprule
Model & Task & Net $\Delta$ & Harmful & Benef. & Churn & W$\to$w & Ans.\ churn \\
\midrule
Qwen2.5-1.5B & MMLU  & $+0.012292$ & 0.087965 & 0.100256 & 0.188221 & 0.091924 & 0.280145 \\
Qwen2.5-1.5B & GSM8K & $+0.096800$ & 0.098600 & 0.195400 & 0.294000 & 0.273800 & 0.567800 \\
Llama-3.2-3B & MMLU  & $+0.030922$ & 0.078465 & 0.109386 & 0.187851 & 0.098049 & 0.285899 \\
Llama-3.2-3B & GSM8K & $-0.017800$ & 0.113200 & 0.095400 & 0.208600 & 0.155800 & 0.364400 \\
Qwen2.5-7B   & MMLU  & $-0.006267$ & 0.059393 & 0.053126 & 0.112520 & 0.045250 & 0.157770 \\
Qwen2.5-7B   & GSM8K & $-0.005800$ & 0.091200 & 0.085400 & 0.176600 & 0.110000 & 0.286600 \\
Llama-3.1-8B & MMLU  & $+0.017163$ & 0.081270 & 0.098433 & 0.179704 & 0.090956 & 0.270659 \\
Llama-3.1-8B & GSM8K & $+0.013200$ & 0.083800 & 0.097000 & 0.180800 & 0.096400 & 0.277200 \\
\bottomrule
\end{tabular}
\end{table}

Panel~(c) applies the atlas metrics of \S\ref{sec:atlas} to the controlled
cells, computed by the same function over the same definitions, so the two
populations can be read against each other. Here the contrast is \emph{method
against method at one bit width}, where the atlas contrast is quantized against
FP16.

\subsection{Post-hoc resolution analysis}
\label{app:minigrid:resolution}

\begin{table}[!t]
\centering
\small
\caption{Post-hoc resolution analysis, read by \S\ref{sec:minigrid:resolution}.
Ranges of the two ratios across the four cells of each task. $p_d$ is computed
per cell from panel~(c) of Table~\ref{tab:h3-supporting} and is \emph{not} the
harness study's $\bar{Q}$, which is a quantized-against-FP16 contrast on a
different pair. This quantity was not registered; it was requested after the
eight-cell verdict was computed and signed, it tests no hypothesis, and it
modifies no verdict.}
\label{tab:h3-resolution}
\begin{tabular}{lrr}
\toprule
Task & $\max\mathrm{range}$ in paired SE & $\mathrm{gap}$ in paired SE \\
\midrule
MMLU ($n = 14{,}042$) & 5.53 -- 17.45 & 2.21 -- 8.45 \\
GSM8K ($n = 1{,}000$) & 1.92 -- 3.61  & 0.44 -- 5.65 \\
\bottomrule
\end{tabular}
\end{table}

\subsection{Deferred registered analyses}
\label{app:minigrid:deferred}

The registered secondary analyses remain deferred with the rest of the main
grid: the 3-bit dose-response check, ARC-Challenge, HellaSwag, and the
WikiText-2 calibration-distribution contrast on Qwen2.5-1.5B-Instruct. The
last of these is governed by a dated amendment that defines a WikiText-2
``document'' as a reconstructed article, after a preflight found that
0 of 36{,}718 raw rows reach the registered 2{,}048-token length while 425 of
629 reconstructed articles do; the amendment records that no accuracy result had
been inspected when it was taken. None of these analyses can substitute for a
confirmatory cell.

\subsection{The earlier public-checkpoint pilot}
\label{app:minigrid:pilot}

The project's earlier public-checkpoint pilot shows the same discriminant
behaviour under worse conditions, and adds the case the eight cells lack, a
positive detection: MMLU with a public GPTQ checkpoint at $n = 400$ gave
$-4.25$\,pp at exact McNemar $p = 0.036$, while GSM8K at $n = 200$ gave
$p = 0.672$ and $p = 0.880$ on churn of $0.250$ and $0.220$, two correct
declinations where detection would have needed 4{,}923 and 17{,}347 items.
\emph{The smaller evaluation is the one that detected nothing.} Its caveats
bound it entirely (raw-text prompts with no chat template, an unpinned Kaggle
image modified in place, two public checkpoints with fixed or undocumented
calibration so it cannot test H3, and wide intervals at $n = 400$ and $200$),
and it is exploratory evidence for discriminant behaviour and nothing else.

%
%

\section{Harness-sensitivity detail}
\label{app:harness-detail}

This appendix holds the motivation and the design of the configuration-sensitivity
study of \S\ref{sec:sensitivity}, whose two result tables remain in the main text.

\subsection{Two observed defects}
\label{app:sensitivity:motivation}

The motivation is on the record: this campaign produced two configuration
effects inside eight days, both on the pinned \texttt{lm\_eval}~0.4.12.

First, \texttt{fewshot\_as\_multiturn} was auto-enabled under a chat template
with no flag set: the harness turns it on whenever \texttt{apply\_chat\_template}
is set (\texttt{lm\_eval/config/evaluate\_config.py:306--308}), moving every
few-shot exemplar out of the user message into its own conversation turn.
Second, and starker, the stock \texttt{gsm8k} task ships two extraction filters,
\texttt{strict-match} and \texttt{flexible-extract}, and scores the \emph{same}
generations under both. On Qwen2.5-1.5B, \texttt{strict-match} voided
\textbf{617 of 1{,}000} responses, \textbf{336} of which \texttt{pilot\_eval}'s
extractor scores correct, because the model writes \texttt{\#\#\#\# \$18} and the
strict regex rejects the \texttt{\$}. Reported accuracy moved
\textbf{$0.232 \to 0.566$} with not one token of the model's output changed.
If a scoring switch on unchanged generations can move aggregate accuracy by a
third of the scale, the size of that effect \emph{relative to} quantization is an
empirical quantity worth measuring rather than asserting.

\subsection{Design and the pre-named ratio}
\label{app:sensitivity:design}

The study varies one native \texttt{lm\_eval} knob at a time from a fixed
reference (chat template on, three inline GSM8K exemplars, zero-shot for MMLU,
and \texttt{flexible-extract} scoring), which is the configuration the mini-grid's
own reference runs use. All runs are FP16-only, on the bridge item subsets
exclusively: the four bridge MMLU subjects at 100 items each ($n=400$) and GSM8K
test indices 0--199 ($n=200$). No quantized checkpoint is loaded under this
protocol, and the full mini-grid item definitions are never evaluated, so no
confirmatory surface is touched.

The headline statistic is fixed before any run:
\begin{equation}
R_{\mathrm{cond}} \;=\; \frac{C_{\mathrm{cond}}}{\bar{Q}},
\label{eq:sensitivity-R}
\end{equation}
where the numerator $C_{\mathrm{cond}}$ is the correctness-state churn between
the reference and a condition, meaning the fraction of items whose
correct/incorrect state changes, on the same churn definition the atlas uses
(\S\ref{sec:atlas}), on the FP16 model over the item set above, and the
denominator $\bar{Q}$ is the mean, over the ten Qwen2.5-1.5B quantized variants
$\{\mathrm{gptq}_{s0\ldots s4},\ \mathrm{awq}_{s0\ldots s4}\}$, of
correctness-state churn against that model's FP16 cell on the \emph{same} item
subset, per task. Both terms are reported beside every $R$; the registration
forbids printing the ratio without its inputs, and if $\bar{Q}=0$ the ratio is
undefined rather than infinite.

The two denominators are
$\bar{Q}(\text{MMLU})=\mathbf{0.199}$ and
$\bar{Q}(\text{GSM8K})=\mathbf{0.287}$; neither is zero, so no undefined case
arises.
Their provenance is deliberate. Under the registration's two-phase design the
numerators run immediately on freeze, but $\bar{Q}$ is computed \emph{only} from
mini-grid results after the registered validator passed over the complete
44-JSONL set (409/409 checks) and the first accuracy inspection that
\texttt{MINIGRID\_REGISTRATION} \S5 permits had been authorized, so the
denominator is not an early look at the confirmatory grid under another name.
A shortcut that would have produced the ratio roughly five weeks earlier, a
dated amendment permitting an early partial inspection of the bridge quantized
deltas, was considered and declined, on the ground that the registered
blindness is not weakened to obtain a number sooner.

\subsection{Condition B in full}
\label{app:sensitivity:condB}

Condition~B is the sharpest result and deserves isolating, because it holds the
generations fixed. It costs zero GPU time: the stock \texttt{gsm8k} task applies
both \texttt{strict-match} and \texttt{flexible-extract} to the same outputs in a
single run, so B is read out of the reference run's own samples file, a genuine
rescore of \emph{identical} generations, not a second pass.
Its correctness churn is therefore the pure effect of the extraction filter:
$C_B = 0.455$ ($91$ of $200$ items), with a directional split of
$\textbf{91}/\textbf{0}$: every changed item went correct$\to$incorrect, so the
$0.455$ accuracy drop equals the churn exactly and \texttt{strict-match} accepts
a strict subset of what \texttt{flexible-extract} accepts here. The resulting
$R_B = 0.455/0.287 = \textbf{1.585}$: on these items, choosing one of two
built-in scoring filters that ship with the same task moves the fixed model's
per-item correctness \emph{more than half again as much} as quantization does.
No compression method in the comparison set produces a swing of that kind from a
choice a practitioner never records.

\subsection{The same move, in another domain}
\label{app:sensitivity:bronder}

Holding the system fixed and varying the instrument is not specific to
compression. \citet{bronder2026instrument} do it for language-model
\emph{honesty} evaluation: with the player model held fixed, four instrument
choices (outcome grammar, criterion disclosure, budget rendering and register
presence) substantially changed what the evaluation would have reported, under
decision rules recorded before results were read. That study and this one share
a design and a conclusion on different objects, which is the reason to expect
the effect wherever an evaluation is treated as a fixed instrument.

\subsection{Why MMLU collapses to a single off-reference cell}
\label{app:sensitivity:mmlu-collapse}

MMLU is zero-shot with no extraction filter, so conditions~A and~B do not apply;
its stock \texttt{num\_fewshot} resolves to $0$ and it ships no
\texttt{filter\_list}, so stock defaults and the reference differ in
\emph{nothing but the chat template}. Conditions~C (chat off) and~D (stock
defaults) are then byte-identical configurations, and inventing a distinct~D
would mean measuring a configuration no user runs. The registration ruled it be
run once and reported as ``$C \equiv D$''.

\subsection{Condition-by-condition results}
\label{app:sensitivity:conditions}

\begin{table}[!t]
\centering
\small
\caption{Configuration sensitivity on FP16 Qwen2.5-1.5B, against the
quantization denominator $\bar{Q}$ for each task. $C_{\mathrm{cond}}$ is
correctness-state churn versus the reference; $R=C_{\mathrm{cond}}/\bar{Q}$ is
reported only with both inputs in view. The
$\text{c}{\to}\text{i}/\text{i}{\to}\text{c}$ column splits churned items by
direction so churn is never read as net degradation. Conditions~C and~D are
byte-identical for MMLU and are reported once as $C\equiv D$ (registration
\S3.3).}
\label{tab:sensitivity}
\setlength{\tabcolsep}{2pt}
\small
\begin{tabular}{l>{\raggedright\arraybackslash}p{2.7cm}rrrrr}
\toprule
Cond & Configuration vs.\ reference & Acc & Net $\Delta$ & $C_{\mathrm{cond}}$ & c$\to$i / i$\to$c & $R=C_{\mathrm{cond}}/\bar{Q}$ \\
\midrule
\multicolumn{7}{l}{\emph{GSM8K}, $n=200$, reference accuracy $0.575$, $\bar{Q}=0.287$} \\
A & exemplars as separate turns   & 0.515 & $-0.060$ & \textbf{0.240} & 30 / 18 & \textbf{0.836} $=0.240/0.287$ \\
B & \texttt{strict-match} rescore & 0.120 & $-0.455$ & \textbf{0.455} & 91 / 0  & \textbf{1.585} $=0.455/0.287$ \\
C & chat template off, 3-shot     & 0.475 & $-0.100$ & \textbf{0.320} & 42 / 22 & \textbf{1.115} $=0.320/0.287$ \\
D & stock defaults (chat off, 5-shot) & 0.495 & $-0.080$ & \textbf{0.300} & 38 / 22 & \textbf{1.045} $=0.300/0.287$ \\
\midrule
\multicolumn{7}{l}{\emph{MMLU}, $n=400$, reference accuracy $0.415$, $\bar{Q}=0.199$} \\
$C\equiv D$ & chat template off, zero-shot & 0.460 & $\mathbf{+0.045}$ & \textbf{0.210} & 33 / 51 & \textbf{1.055} $=0.210/0.199$ \\
\bottomrule
\end{tabular}
\end{table}

Three of the four GSM8K conditions (turning the chat template off, accepting
every stock default at once, and switching the extraction filter) have
$R \geq 1$: each moves a fixed model's per-item correctness by \emph{as much as
or more than} swapping in one of ten quantized variants moves it. Only
condition~A, the multiturn exemplar placement, stays below the quantization
scale. The net accuracy deltas understate the movement in every case, because
they net directional churn that the
$\text{c}{\to}\text{i}/\text{i}{\to}\text{c}$ column keeps separate.
Condition~B is sharpest because it holds the generations fixed: the stock task
applies both filters to the same outputs, so B is a rescore of \emph{identical}
generations at zero GPU cost, and its $91/0$ split means \texttt{strict-match}
accepts a strict subset of what \texttt{flexible-extract} accepts here
(Appendix~\ref{app:sensitivity:condB}).

The MMLU cell makes a point the net delta hides. Turning the chat template off
\emph{raises} reported accuracy by $+4.5$\,pp ($0.415 \to 0.460$), with a
directional split of $33/51$, meaning more items recovered than lost. The
harness default is therefore not uniformly worse than the reference: on MMLU it
is better, on GSM8K worse, and either way \emph{different}. The churn behind the
$+4.5$\,pp is $0.210$ ($84$ of $400$ items), giving $R=1.055$, again on the
quantization scale. The movement is real, signed differently on the two tasks,
and of a kind compression papers do not report: the setting that almost never
appears in a model card is precisely the one silently moving the number a
``near-lossless'' claim rests on. Holding the system fixed and varying the
instrument is not specific to compression, and
Appendix~\ref{app:sensitivity:bronder} records the same design and the same
conclusion on a different object.

%
%

\section{Artifact detail}
\label{app:artifacts-detail}

\subsection{The six released artifacts}
\label{app:artifacts:released}

The \textbf{per-item outputs} are 88 cell JSONL files, one row per evaluation
item, for every model $\times$ method $\times$ seed $\times$ task in the
controlled experiment, carrying the per-item correctness state that the paired
test consumes. \S\ref{sec:audit:v3} reports that \AuditPerItemTaskMatched{} of
the \AuditEligible{} eligible sources release \emph{task-matched} per-item
outputs, those covering the tasks their own equivalence claim is about, and that
\AuditPerItemOtherTaskOnly{} (\AuditPerItemOtherTaskClaims{}) release outputs for
other tasks only; the fifth line of the reporting standard in
\S\ref{sec:conclusion} asks the field to close that gap, and withholding ours
would leave that recommendation untestable against the paper making it. They
ship in both forms deliberately: the extracted files are directly loadable, and
the two sealed run archives carry the same bytes with a per-file SHA-256
manifest and an archive checksum recorded when the runs completed, so the
loadable copy can be checked against the record made at run time instead of
trusted.

The \textbf{flip atlas} gives per-cell paired statistics for every enumerated
pair-task cell, with the exclusion table and the frozen 59-pair manifest.
\texttt{flipeval} is the analysis package: flip rates, the churn family, exact
McNemar, TOST at a declared margin, bootstrap intervals, item-bootstrap
rank-flip rates, minimum detectable difference, required-$n$, and the
certification-table generator (Apache-2.0; everything else is CC-BY-4.0). The
audit artifacts are the frozen 17-claim table with source content hashes and the
per-claim verdict CSV, with every robustness and transparency column, plus the
source manifest and retrieval script described below; the captured source text
itself is not among them. The certification tables ship twelve rows per margin
at 1, 2 and 3\,pp. The reproduction package holds registrations and their dated
amendments, signed decision records, configs, the container image SHA-256 and
build recipe, scheduler scripts, per-run manifests, the source-state freeze
fingerprints, and the campaign incident log, which is included on purpose: a
pipeline with 28 recorded catches is better evidence that it was verified than
one presenting a clean surface.

The redistribution review of 2026-08-02 examined the terms attached to each
audited source before any copy was published. It found four sources, the Meta AI
blog post, the NVIDIA TensorRT-LLM documentation page and two vLLM pages,
carrying no grant that would permit a third party to republish their text, and
it found the seven method papers under arXiv's default licence, which authorises
arXiv to distribute them rather than authorising us to. Publishing the corpus was
therefore never an option that was open to us, and the working archive of it is
kept private. What ships instead is what makes the corpus rebuildable and
checkable: the source URL and pinned version identifier for every claim, a
SHA-256 for every captured file, the per-file manifest
\texttt{docs/\allowbreak{}audit\_sources\_\allowbreak{}manifest.tsv} carrying
both digests and a per-row provenance status, and
\texttt{scripts/\allowbreak{}fetch\_\allowbreak{}audit\_sources.py}, a
standard-library script that re-fetches each source by the retrieval method
recorded for it and compares the result against the recorded digest. Quoted text
appears in this paper as short excerpts with their locations, and a reader who
runs the script obtains each source from its own publisher. Two of the
seventeen, R11 and R13, bound what a digest can establish at all; the manifest
records that per row, neither is described anywhere in this paper as
hash-verified, and \S\ref{app:artifacts:datasheet} states what each limit is.

\subsection{Datasheet}
\label{app:artifacts:datasheet}

Following \citet{gebru2021datasheets}.

\paragraph{Motivation.} The artifacts exist to make the paper's own
recommendation checkable against the paper. \S\ref{sec:audit} reports that no
audited source releases per-item outputs; withholding ours would make the
recommendation unfalsifiable. They were created by the author, with no external
funding beyond a university compute allocation.

\paragraph{Composition.} Three kinds of record. (i) \textbf{Per-item outputs}:
88 cell JSONL files, one row per evaluation item, covering every model $\times$
method $\times$ seed $\times$ task in the controlled experiment. (ii)
\textbf{Derived per-cell statistics}: the atlas, at one row per pair-task cell,
with flip and churn quantities, McNemar and TOST outcomes. (iii)
\textbf{Provenance records}: registrations, signed decisions, configs, receipts,
fingerprints and the incident log. \textbf{(*) The atlas contains no
human-subject data and no personally identifying information}; its rows describe
model behaviour on public benchmark items. \textbf{(*) The population is the
public record of compression evaluation, not a census of quantization}
(\S\ref{sec:atlas:caveats}): it is conditioned on what was published and
leaderboard-indexed, so it describes the evidence the field circulates rather
than how quantization behaves in general.

\paragraph{Collection.} S1 cells were mined from public per-item evaluation
dumps; S2 from vendor-published evaluations; the controlled cells were generated
by us inside a pinned container, under protocols frozen before the runs
(\S\ref{sec:prereg}). Pair enumeration was frozen in a manifest before any
statistic was computed.

\paragraph{Preprocessing.} A pair-task cell is admitted only when both sides
carry a binary per-item correctness state and full-prompt hashes match across the
pair. Exclusions are released as a table with a reason per row rather than
summarised, so the admitted population is auditable and not merely asserted.

\paragraph{Distribution and licensing.} Released under \textbf{CC-BY-4.0};
the \texttt{flipeval} package under \textbf{Apache-2.0}.
\textbf{(*) S1's Open LLM Leaderboard v1 archive carries no declared license},
which is recorded as a limitation rather than worked around: we redistribute
\emph{our derived per-cell statistics} and \textbf{not} the raw upstream
per-item files. The frozen pair manifest records the upstream dataset
identifiers and run timestamps per pair, so a third party can re-derive S1 from
the original sources.
\textbf{(*) S2 is Apache-2.0} upstream. Site-specific identifiers (absolute
paths, charge account, usernames, cluster hostnames) were replaced with
placeholders before publication; no accuracy, hash, count, seed, timestamp, job
ID or decision value was altered, and the sealed archives were not rewritten, so
their recorded checksums still verify.
\textbf{(*) The 17 audited sources are not redistributed.} A redistribution
review on 2026-08-02 found four of them with no third-party republication grant
and the seven method papers under arXiv's default licence, which authorises
arXiv rather than us to distribute them, so the release carries source URLs,
pinned version identifiers, per-file SHA-256 digests, the manifest and a
retrieval script in place of the captured text, and the full-text captures are
kept private (\S\ref{sec:artifacts}). \textbf{(*) Two sources carry provenance
limits no digest can close}, and the manifest records that per row rather than
averaging it away.
R11 is served with per-response content: two fetches made seconds apart returned
different bytes, so the digest recorded for it was never a valid fingerprint of
the page, and the retrieval script reports it as expected drift rather than as a
verification failure. R13 does carry a digest for the archived capture, so a
re-fetch can be checked against that capture; what is absent is any digest
recorded \emph{before} the capture was made, so the capture itself was never
corroborated against an independent earlier record. Provenance for these two is
documentary rather than cryptographic, neither is described anywhere in this
paper as hash-verified, and the other fifteen reproduced their recorded digests
exactly when re-fetched on 2026-07-31.

\paragraph{Revision history.} Both revisions are published and the delta is
reported. Rev-1 carried a population defect found by an independent spot-check
that reconciled all 262 compared fields: the arithmetic was right and the
population was not (\S\ref{sec:prereg-spotcheck},
Appendix~\ref{app:prereg:rev2delta}). Rev-2 is what the paper cites; rev-1 is
retained rather than overwritten, because a corrected artifact whose correction
remains visible is this paper's argument applied to itself.

\subsection{Metadata and identifiers}
\label{app:artifacts:metadata}

Released at version \textbf{v1.2.0}, the string shared by the source tag, the
archived release and the dataset revision. The archived DOI is canonical for
citation; the dataset repository is a convenience mirror.

\begin{tabular}{@{}l p{0.56\linewidth}@{}}
\toprule
Identifier & Value \\
\midrule
Version DOI (canonical) & \versiondoi \\
v1.0.0 DOI (superseded) & \priorversiondoi \\
Concept DOI (latest) & \conceptdoi \\
Source package & \repopath, tag \texttt{v1.2.0} \\
Dataset mirror (secondary) & \datasetpath \\
Container image SHA-256 & \texttt{8260d04c\ldots1db2007} \\
\bottomrule
\end{tabular}

\medskip
\noindent The \textbf{version DOI is what this paper cites}: it resolves to one
frozen state. The concept DOI is correct only when referring to the artifact
series. \textbf{The superseded v1.0.0 DOI is listed so that a reader holding it
can tell which release they have}: its verdict CSV is rev-2, and
\S\ref{sec:artifacts} states what separates the two. A
\textbf{Croissant record} is served by the dataset repository at its standard
\texttt{/croissant} endpoint and carries the CC-BY-4.0 licence through to the
machine-readable metadata.

\subsection{Maintenance}
\label{app:artifacts:maintenance}

Maintained by the author for \textbf{at least 12 months} from release, via the
source repository's issue tracker, with a target first response of two weeks.
Fixes ship as new tagged versions with a changelog; released versions are never
edited in place, and a superseded revision stays published beside its
replacement. \textbf{Adding atlas pairs requires a dated amendment to the atlas
registration}, not a silent extension. The frozen pair manifest is what makes
the population auditable, so growing it quietly would remove the property the
artifact exists to demonstrate.

The paired-comparison layer described in \S\ref{sec:artifacts} has been
\textbf{proposed} to the \texttt{lm-evaluation-harness} maintainers, in a comment
on \harnessissue; no integration has been merged, and none should be inferred.

\section{Continuous and graded metrics: the extension this paper does not run}
\label{app:continuous}

This appendix states what would be required to certify a metric that scores each
item on a scale instead of marking it right or wrong. It is a derivation and a
proposal. No part of it has been executed, no quantity in it has been measured,
and none of the required sample sizes reported in this paper applies to any
metric outside the binary per-item model.

\subsection{Why the binary outcome model comes first}
\label{app:continuous:why}

Three things follow from binary per-item correctness, and the certification
apparatus uses all three.

The first is an exact variance. Write $c_i^{\mathrm{base}}, c_i^{\mathrm{comp}}
\in \{0,1\}$ for the correctness of item $i$ under the baseline and the
compressed model, and $d_i = c_i^{\mathrm{comp}} - c_i^{\mathrm{base}}$ for the
per-item accuracy difference, so that $d_i \in \{-1,0,+1\}$. Because $d_i^2$ is
the indicator that the two models disagree on item $i$,
\begin{equation}
\mathrm{Var}(d_i) \;=\; \mathbb{E}[d_i^2] - \delta^2 \;=\; p_d - \delta^2 ,
\qquad \delta = \mathbb{E}[d_i],
\label{eq:vard-binary}
\end{equation}
and under the null of no true difference this is $p_d$ exactly, which is the
$\mathrm{sd}_{\mathrm{paired}}$ of Equation~\eqref{eq:sds}. Nothing is assumed
about the joint distribution of the two runs beyond the pairing itself: the
second moment of a variable supported on $\{-1,0,+1\}$ \emph{is} the probability
that it is non-zero. That identity is what makes the certification tables
countable. The variance term is a rate a third party obtains by comparing two
per-item dumps, so Table~\ref{tab:certification} is populated by measurement
(\S\ref{sec:atlas}) and not by an assumption about how models vary.

The second is that binary correctness is what the public record carries. The
atlas mines per-item dumps whose recorded quantity is a correctness field
(Appendix~\ref{app:atlas:sources}); the cells it met that carried no binary
correctness metric at all were excluded rather than scored, and that exclusion
is reported with its count in Appendix~\ref{app:atlas:exclusions}. A continuous
apparatus built on this record would have had nothing to be built from.

The third is that binary correctness is the outcome type the audit could score.
Of the eligible claims, exactly \AuditNotAssessableOutsideFramework{} was set
aside because the metric its qualifying sentence speaks about carries no
per-item correctness state, and that claim, \AuditOutsideFrameworkClaim{}, is
the subject of \S\ref{app:continuous:r04} below.

\subsection{TOST for a bounded graded score}
\label{app:continuous:tost}

The generalisation of the planning calculation is immediate, and stating it
costs nothing, because TOST is defined on the mean of a paired difference and
never on its support. Let each item carry bounded scores $s_i^{\mathrm{base}},
s_i^{\mathrm{comp}} \in [a,b]$ under the same metric, put $d_i =
s_i^{\mathrm{comp}} - s_i^{\mathrm{base}}$, and let the reported statistic be
$\bar{d} = n^{-1} \sum_i d_i$ with $\sigma_d^2 = \mathrm{Var}(d_i)$. TOST at
one-sided $\alpha$ with power $1-\beta$, against an assumed true difference of
zero and at a margin $m$ expressed in the metric's own units, then requires
\begin{equation}
n_{\mathrm{req}} \;=\; \left\lceil \left(\frac{(z_{1-\alpha} + z_{1-\beta})\,
\sigma_d}{m}\right)^{\!2} \right\rceil ,
\label{eq:nreq-continuous}
\end{equation}
which is Equation~\eqref{eq:nreq} with the paired standard deviation replaced by
$\sigma_d$. The binary case is the special case $\sigma_d^2 = p_d$ of
Equation~\eqref{eq:vard-binary}. Two consequences are pure algebra and worth
stating because they bound what any adaptation can achieve: the requirement
scales as $(\sigma_d/m)^2$, so a graded metric needs more items than the
corresponding binary row of Table~\ref{tab:certification} exactly when
$\sigma_d^2 > p_d$ and fewer when it is smaller; and every qualification already
attached to Equation~\eqref{eq:nreq} carries over unchanged, including that this
is a planning size under an assumed true difference of zero and is therefore a
lower bound on what a non-zero true difference would demand
(\S\ref{sec:cert:method}).

What does not transfer is everything empirical.

\emph{The churn tables do not transfer.} $\sigma_d$ is a property of the metric,
of the model pair and of the benchmark family jointly, and it has to be measured
for each combination in the same way $p_d$ was. This paper measures it for none.
The atlas records accuracy-state churn and carries no per-item score dispersion
for any metric, so there is no path from the numbers in this paper to a
$\sigma_d$ for CIDEr, for ROUGE, or for a judge scale. A practitioner adapting
the method supplies that measurement; the method does not supply it to them.

\emph{The interpretation of churn does not transfer.} For a binary outcome,
churn is a count: the proportion of items whose correctness state changed, which
is simultaneously the variance term and a directly reportable description of
per-item behaviour. For a graded metric the corresponding quantity is a
dispersion of per-item score changes, and it is not a count of anything. In
particular the net-versus-gross decomposition that motivates this paper has to
be redefined before it means anything: a cell can show a net difference of zero
with arbitrarily large $\sigma_d$, but "how many items changed" is no longer the
question, since almost every item changes a little.

\emph{The margin does not transfer.} $m$ in
Equation~\eqref{eq:nreq-continuous} is in the metric's own units. A margin of two
CIDEr points and a margin of two accuracy points are unrelated quantities, and
neither can be read across to the other, so a certificate on a graded metric is
interpretable only if the margin is declared in that metric and defended in that
metric. The registered $2$\,pp margin of this paper is an accuracy margin and
means nothing elsewhere.

\subsection{How three metric families would map}
\label{app:continuous:families}

Each of the following is a proposal about structure, not a result.

\paragraph{pass@$k$.} At $k=1$ the per-item outcome is a single pass or fail, so
this is the binary model already, with the added wrinkle that sampling makes
correctness itself random and one draw is one draw. For $k > 1$ the per-item
quantity is an estimated success probability $\hat{\pi}_i \in [0,1]$, computed
from a finite number of completions. The per-item difference is then bounded and
graded, so Equation~\eqref{eq:nreq-continuous} applies in form, but the
estimation error enters the variance: writing $\pi_i$ for the quantity being
estimated, the law of total variance gives
\begin{equation}
\sigma_d^2 \;=\; \mathrm{Var}\!\left(\pi_i^{\mathrm{comp}} -
\pi_i^{\mathrm{base}}\right) \;+\;
\mathbb{E}\!\left[\mathrm{Var}\!\left(\hat{\pi}_i^{\mathrm{comp}} -
\hat{\pi}_i^{\mathrm{base}} \,\middle|\, \pi_i^{\mathrm{comp}},
\pi_i^{\mathrm{base}}\right)\right],
\label{eq:passk-variance}
\end{equation}
whose second term shrinks as the number of completions per item grows. Sizing a
pass@$k$ evaluation therefore involves two budgets, items and completions per
item, and only the first is what a required-$n$ table counts. If the same
sampling seeds drive both sides, the two estimation errors are positively
correlated and the second term is smaller; that is a design choice to be
declared, not a property to be assumed.

\paragraph{BLEU, ROUGE and CIDEr.} These assign a bounded score to a generated
string, so at the level of a single item they are exactly the graded case above
and Equation~\eqref{eq:nreq-continuous} applies once $\sigma_d$ is measured. The
obstacle is aggregation rather than gradedness. Corpus-level BLEU is not the
mean of per-sentence BLEU: it is computed from n-gram match counts and reference
lengths summed over the whole corpus, with a brevity penalty formed from those
totals, so the reported statistic is not an average of per-item quantities and
there is no $d_i$ whose mean it is. Certifying it therefore requires either a
sentence-level variant, which is a different number from the corpus figure and
must be declared as the certified quantity, or a paired resampling scheme over
items that respects the aggregation, in which case the sample size is found by
simulation and Equation~\eqref{eq:nreq-continuous} does not give it. ROUGE and
CIDEr are ordinarily reported as means of per-item scores and do not have this
particular problem, which is why the aggregation question has to be asked per
metric rather than per family.

\paragraph{LLM-as-a-judge scales.} A judge produces a bounded ordinal score, so
the graded form applies in principle, with two additions. First, the score is
itself a model output, so the observed per-item value is the quantity of
interest plus judge error. Under the simple model that an observed score is
$s_i + e_i$ with judge errors of variance $\sigma_j^2$ and correlation $\rho$
between the two sides, the difference has variance
\begin{equation}
\sigma_d^2 \;=\; \sigma_{\mathrm{true}}^2 \;+\; 2\sigma_j^2 (1 - \rho),
\label{eq:judge-variance}
\end{equation}
so judge noise inflates the requirement unless the same judge scores both sides
of an item in a way that makes its errors correlated. $\sigma_j^2$ has to be
estimated from repeated judgements, which is a third budget, and assuming it
away is the failure mode this paper exists to complain about in another setting.
Second, an ordinal scale is not an interval scale, so a margin stated as a
fraction of a point on a five-point rubric has no agreed meaning; the margin
would have to be defended as a statement about the rubric before the arithmetic
is worth running. We take no position here on how that defence should go.

\subsection{\AuditOutsideFrameworkClaim{} and what scoring it would require}
\label{app:continuous:r04}

\AuditOutsideFrameworkClaim{} is the one audited claim that this limitation
touches directly, and the audit's treatment of it stands as written
(\S\ref{sec:audit:r04}): its qualifying sentence asserts negligible loss on COCO
CIDEr, a generation metric that assigns a graded score to a caption and has no
per-item correct/incorrect state, so V1 and V2, which are defined on $d_i \in
\{-1,0,+1\}$, have nothing to compute and there is no discordance rate to impute.
It is recorded as outside our registered calculation and carries no verdict. The
history of the overruled first-pass computation on a substituted benchmark is in
\S\ref{sec:audit:r04} and Appendix~\ref{app:audit:r04detail}, and that
computation is not claimed anywhere in this paper as an audit result.

The extension above does not retroactively score it, and it is worth being
precise about why, because the reason is not the mathematics. Equation
\eqref{eq:nreq-continuous} would need three inputs that do not exist here:
per-caption CIDEr scores for both models on the same images, from which
$\sigma_d$ could be measured; a margin declared in CIDEr points and defended as
a statement about caption quality; and a decision, taken before looking, about
which of those a claim of negligible loss is asserting.
The audit records that \AuditPerItemTaskMatched{} of the \AuditEligible{}
eligible sources release \emph{task-matched} per-item outputs
(\S\ref{sec:audit:v3}), so the first input is unavailable from the public
record, and the second and third were never registered.
The limit is the model we registered and the evidence the source released, not
paired analysis, which handles graded metrics without difficulty.

%
%

\section{Automated claim extraction: procedure, reliability, and AI use}
\label{app:extraction}

\subsection{Why this appendix exists}

The claim table underlying \S\ref{sec:audit} was populated by language-model
agent sessions rather than by human reading. That is a provenance fact a reader
needs in order to weigh the audit, and it was previously stated only inside the
reproduced audit registration (Amendment~1,
\S\ref{app:reg:audit}). This appendix states it directly, specifies as much of
the procedure as the record supports, and is explicit about the parts the
record does not support.

\subsection{The procedure as executed}

\begin{enumerate}
\item \textbf{Pass 1} enumerated the registered source frames and extracted the
  registered fields for each qualifying claim, producing 13 rows
  (\texttt{docs/\allowbreak{}audit\_\allowbreak{}claim\_\allowbreak{}table\_\allowbreak{}pass1.csv}, committed 2026-07-15 as an
  explicitly unfrozen artifact). Some pass-1 fields were derived from a
  summarised reading of the source rather than from the source text directly;
  those rows are flagged in the reconciliation memo.
\item \textbf{Pass 2} ran as a fresh session with the pass-1 file withheld,
  instructed not to read it or recover it from version history, and given only
  the frozen protocol and the source frames. It produced 23 rows (17 claims
  plus 6 logged exclusions) and recorded a SHA-256 of each fetched source
  (\texttt{docs/\allowbreak{}audit\_\allowbreak{}claim\_\allowbreak{}table\_\allowbreak{}pass2.csv}).
\item \textbf{Human reconciliation} merged the two passes into
  \texttt{docs/\allowbreak{}audit\_\allowbreak{}claim\_\allowbreak{}table.csv} (17 claims), resolving inclusion
  disagreements against the registered trigger vocabulary, choosing the quote
  anchored in that vocabulary, and re-reading the cited table wherever the
  passes differed numerically. The merged table was frozen by commit
  \texttt{43ce229} before any verdict, power or required-$n$ computation ran.
  The adjudication is recorded claim-by-claim in
  \texttt{docs/\allowbreak{}AUDIT\_\allowbreak{}RECONCILIATION\_\allowbreak{}2026-07-15.md}.
\end{enumerate}

This replaced the registration's original ``extracted twice on different days''
independence mechanism. Amendment~1 records both the substitution and its
rationale: temporal separation was a proxy for extractor independence calibrated
to human memory, and a fresh session carries no memory of pass~1 at all. The
amendment also records that it was made after pass-1 results were known but
before any verdict computation, and that the \S3.1--3.2 inclusion and extraction
rules were unchanged.

\paragraph{What separate sessions do and do not buy.}
They prevent direct leakage: pass~2 cannot copy pass~1 because it never saw it.
They do \emph{not} deliver the property dual coding is normally used for. Two
sessions of the same model share a prior, so a systematic misreading, such as a
tendency to treat a reported delta as a declared bound or to prefer the
abstract's phrasing over a table footnote, will recur in both passes and will
survive reconciliation looking like agreement. We therefore do not describe
these passes as independent coders, and we do not report an inter-rater
reliability coefficient, which would imply an independence the design does not
have.

\subsection{Specification: what is on record and what is not}

\begin{tabular}{@{}p{0.36\linewidth}p{0.58\linewidth}@{}}
\toprule
Item & Status \\
\midrule
Execution date & 2026-07-15, both passes (commit timestamps
  \texttt{8a16722}, \texttt{43ce229}) \\
Tool family & Command-line coding agent, operated under the repository's
  committed agent guardrails (\texttt{AGENTS.md}, \texttt{CLAUDE.md}) \\
Browsing / retrieval & Enabled; pass 2 fetched each source and recorded its
  SHA-256 \\
Input documents & The registered source frames; per-source SHA-256 in
  \texttt{audit\_\allowbreak{}claim\_\allowbreak{}table\_\allowbreak{}pass2.csv} \\
Output schema & The registered \S3.2 extraction fields; both raw pass files
  are released \\
Human adjudication & Documented per claim in
  \texttt{docs/\allowbreak{}AUDIT\_\allowbreak{}RECONCILIATION\_\allowbreak{}2026-07-15.md} \\
\midrule
Exact model ID and version & \textbf{Not recorded} \\
System and extraction prompts & \textbf{Not recorded; not released} \\
Temperature, seed, reasoning setting & \textbf{Not recorded} \\
Retry policy & \textbf{Not recorded} \\
Whether prompts varied across sources & \textbf{Not recorded} \\
\bottomrule
\end{tabular}

\medskip
\noindent
The lower block is a genuine limitation and we state it as one rather than
reconstructing plausible values after the fact. The consequence is specific:
the extraction is \textbf{auditable but not re-executable}. A reader can check
every extracted field against the released source hashes and the released raw
pass files, and can re-do the extraction by any means they choose; they cannot
reproduce \emph{our} extraction run. Prompt and model-identifier logging is a
requirement we would register in advance if this study were run again, and it is
part of the reporting standard \S\ref{sec:conclusion} proposes.

\subsection{Agreement between the two passes}

Eleven sources were extracted by both passes. Two were reached only by pass~1
and twelve only by pass~2; the union was adjudicated by hand, and one source
(QuIP\#) was included by pass~2 and excluded on adjudication because neither of
its candidate phrases is in the registered trigger vocabulary.

On the eleven jointly extracted sources, the two categorical registered fields
agree exactly:

\begin{tabular}{@{}lr@{}}
\toprule
Field & Agreement \\
\midrule
\texttt{per\_item\_outputs\_released} (yes / no / partial) & 11 / 11 \\
\texttt{statistical\_\allowbreak{}test\_\allowbreak{}or\_\allowbreak{}interval\_\allowbreak{}reported} (any / none) & 11 / 11 \\
\bottomrule
\end{tabular}

\medskip
\noindent
\textbf{The free-text numeric fields cannot be scored this way, and we do not
report a number for them.} The two passes recorded different \emph{scope} by
design rather than different values: pass~1 tended to record every task in the
source's table, pass~2 the single anchor task attached to the quoted sentence.
For GPTQ, for example, pass~1 recorded deltas for five OPT-175B tasks and
pass~2 recorded the PIQA figure alone, which is the same underlying numbers at
different granularity. A string-equality comparison scores that as a disagreement and
would put ``exact agreement'' at 11\%, which would misdescribe the record. A
value-level comparison is equally unreliable, because these fields are prose
containing incidental numbers.

What is on record is the reconciliation memo's finding, made at merge time by
re-reading the cited table for every discrepancy: \emph{every} numeric
disagreement between the passes was a difference of scope, not of value. We
report that as the qualitative finding it is. A quantitative inter-pass
agreement rate on numeric fields was not computed at extraction time and cannot
be reconstructed now without imposing a scoring rule that did not exist then.

\paragraph{Validation against human reading was not performed.}
No random subset was independently extracted by a second human without sight of
the agent output. That check would materially strengthen the audit and it is the
first thing we would add.

\subsection{AI-use statement}

Language-model tools were used in this work for two blinded structured
extraction passes over the audited sources, and for copy-editing. The author
designed the study, wrote the preregistrations and froze them before execution,
wrote and verified the analysis code, adjudicated every extraction disagreement
by re-reading the cited sources, checked the reported results against the
released artifacts, and takes responsibility for the whole manuscript. The
claim table produced by the extraction passes is released in both raw
(per-pass) and reconciled form, together with per-source content hashes and the
claim-by-claim adjudication record. Model identifiers, prompts and generation
settings for the extraction sessions were not logged at the time and are
therefore not released; see the specification table above.

{\sloppy\setlength{\emergencystretch}{6em}
%
%
%
%

\section{Preregistration documents}  
\label{app:registrations}  

\label{app:reg:note}  
This appendix reproduces the four frozen protocol documents that
\S\ref{sec:prereg} describes, so that the rule text can be read without
leaving the paper. Each is reproduced in full, including its dated
amendments. Formatting is typeset rather than plain-text, but no wording
has been changed: the conversion is generated and machine-checked
word-for-word against the frozen files. The authoritative copies are the
files themselves in the archived source package
(\S\ref{sec:artifacts}).

\subsection{FlipEval Main-Grid Pre-Registration}
\label{app:reg:main}  
\emph{Reproduced from \texttt{PREREGISTRATION.md}; see \S\ref{app:reg:note}.}  

Changelog: 2026-07-11 --- resolved all TBDs and tightened the H3 decision rules prior to any main-grid execution; frozen by the commit containing this line.

Created: 2026-07-10. Status: locked before the first main-grid job.

After the first main-grid job starts, this file will not be edited. Any deviation or clarification will be appended under \textbf{Dated Amendments}, with its date, rationale, and whether results were inspected before the decision.

\subsubsection{Claims}
\begin{itemize}
\item \textbf{H1 (net vs gross):} aggregate accuracy delta measures NET change and can be near zero because harmful and beneficial flips cancel, while GROSS behavioral churn on the same inputs stays large. Different quantities; report both.
\item \textbf{H2 (underpowered, unstable ranking):} at common benchmark sizes, power to detect degradation and to order methods is low; rankings shift under resampling, seed, and calibration data.
\item \textbf{H3 (calibration-seed instability, the ``spike''):} the calibration set used to fit the quantizer may change the method ranking as much as or more than the choice of method. If true, current single-run compression comparisons are effectively not reproducible. THIS is the potential main-track-caliber finding and the thing to hunt hardest in the pilot.
\end{itemize}
\subsubsection{Experimental Grid}
Models:

\begin{itemize}
\item Qwen2.5-1.5B-Instruct
\item Qwen2.5-7B-Instruct
\item Llama-3.2-3B-Instruct
\item Llama-3.1-8B-Instruct
\end{itemize}
Llama-3.2-3B-Instruct is used rather than the 1B model because near-floor GSM8K accuracy at 1B would make flip analysis degenerate and leave insufficient performance headroom across the four benchmarks.

Compression cells:

\begin{itemize}
\item RTN at 4 and 3 bits; seed-free control
\item GPTQ at 4 and 3 bits; calibration seeds \texttt{\{0, 1, 2, 3, 4\}} per model/bit cell
\item AWQ at 4 and 3 bits; calibration seeds \texttt{\{0, 1, 2, 3, 4\}} per model/bit cell
\item Wanda at 2:4 sparsity; calibration seeds \texttt{\{0, 1, 2, 3, 4\}}. Wanda is selected for its lower checkpoint-construction cost, documented implementation, and calibration dependence. SparseGPT is an explicitly out-of-scope pruning alternative.
\item Calibration datasets: C4 for the full grid; C4 and WikiText-2 on Qwen2.5-1.5B-Instruct to separate sample-seed variance from calibration-distribution variance
\end{itemize}
For the full-grid C4 condition, each calibration set contains 128 samples of exactly 2,048 tokens from \texttt{allenai/\allowbreak{}c4}, configuration \texttt{en}, train split. For seed \texttt{s}, shuffle the complete C4 train-split document-index array using \texttt{numpy.random.default\_rng(s).shuffle}; visit documents in that order, tokenize each document without adding special tokens, skip documents shorter than 2,048 tokens, and retain the first 2,048 tokens from each eligible document until 128 samples have been collected. Persist the selected document indices and token hashes. GPTQ seed \texttt{s} and AWQ seed \texttt{s} receive the identical ordered calibration samples. This pairing makes a seed-\texttt{s} ranking difference attributable to method-by-calibration interaction rather than the methods seeing different data. The Qwen2.5-1.5B-Instruct WikiText-2 distribution analysis will use the same seeds, sample count, token length, eligibility rule, and method pairing, with dataset-specific indices retained.

Benchmarks:

\begin{itemize}
\item MMLU
\item ARC-Challenge
\item HellaSwag, using the first 2,000 validation items in dataset index order (indices 0 through 1,999)
\item GSM8K, at least 1,000 fixed test items
\end{itemize}
The model-family chat template is ON for every benchmark and method, including the FP16 baseline. GSM8K few-shot examples are inline within the user message. All paired methods receive identical item sets, prompt construction, and decoding/scoring settings.

\subsubsection{Outcomes and Analysis}
Primary per-pair metrics are net accuracy delta \texttt{(c-\allowbreak{}b)/\allowbreak{}n}, harmful flip rate \texttt{b/\allowbreak{}n}, beneficial flip rate \texttt{c/\allowbreak{}n}, accuracy-state churn \texttt{(b+c)/\allowbreak{}n}, wrong-to-different-wrong churn, and total answer churn. The primary inferential outputs are bootstrap 95\% confidence intervals, exact McNemar tests, TOST equivalence, minimum detectable difference at 80\% power, required n at 80\% power for the observed effect, and item-bootstrap rank-flip rates.

The TOST equivalence margin is fixed at \textbf{2 percentage points} (\texttt{0.02}) for accuracy delta. We will not interpret failure to reject a difference as equivalence. McNemar tests are two-sided and exact. Bootstrap seeds, item IDs, calibration sample indices, and environment fingerprints will be retained. Family-wise multiple comparisons will use Holm correction within each benchmark/model family of planned method contrasts.

\paragraph{Hierarchical aggregation across calibration seeds}
\mbox{}\\
Calibration seeds are treated as a random effect. Item-level bootstrap intervals from a single seed do not represent between-seed uncertainty. For every model/benchmark/bit/method cell, we will report (1) item-bootstrap uncertainty separately within each seed, (2) the standard deviation of the five seed-level accuracy estimates, and (3) a two-level paired bootstrap interval. Each replicate of the two-level bootstrap samples the five seed labels with replacement and, within every selected seed, samples the common evaluation items with replacement; GPTQ and AWQ retain the same sampled seed labels and item indices. We will report seed-level SD and item-level SE as separate variance components rather than collapsing them into an item-only interval.

FlipEval's rank-instability metric will be reported both within each seed using item resampling, for comparison with the pilot, and across the paired seed-by-item joint bootstrap, which is the H3-relevant estimate. The joint procedure uses the same two-level resamples described above and reports exact-tie replicates separately.

\subsubsection{H3 Decision Rule}
The primary confirmatory analysis is restricted to 4-bit GPTQ and AWQ over the fixed set

\texttt{S = \{Qwen2.5-\allowbreak{}1.5B-\allowbreak{}Instruct, Qwen2.5-\allowbreak{}7B-\allowbreak{}Instruct, Llama-\allowbreak{}3.2-\allowbreak{}3B-\allowbreak{}Instruct, Llama-\allowbreak{}3.1-\allowbreak{}8B-\allowbreak{}Instruct\} $\times$ \{MMLU, GSM8K\}},

which contains eight model-by-benchmark cells. MMLU is the registered likelihood-based benchmark and GSM8K the registered generative benchmark in the confirmatory set. ARC-Challenge and HellaSwag are secondary. The identical H3 analyses at 3 bits are pre-declared secondary/exploratory analyses and will be reported regardless of outcome. Four-bit results are primary because 4-bit quantization is the deployment-relevant setting; 3-bit instability is expected to be larger and is interpreted only as a dose-response check, not as a second opportunity for confirmatory significance.

For seed \texttt{s}, GPTQ-\texttt{s} is compared with AWQ-\texttt{s} using the paired calibration set. Let

\texttt{d\_s(model, benchmark, bit) = acc\_GPTQ,s -\allowbreak{} acc\_AWQ,s}.

A winner flip occurs in a cell if there are registered seeds \texttt{s} and \texttt{t} for which \texttt{sign(d\_s) != sign(d\_t)} and both differences are nonzero. An exact accuracy tie (\texttt{d\_s = 0}), which can occur because item counts are discrete, is counted as neither a flip nor a non-flip and is reported separately. Thus a tie cannot create or erase a flip between two non-tied seeds.

Define the absolute mean method gap as

\texttt{gap(model, benchmark, bit) = |mean\_s(acc\_GPTQ,s) -\allowbreak{} mean\_s(acc\_AWQ,s)|},

and, for method \texttt{m} in \texttt{\{GPTQ, AWQ\}}, define the seed-induced range as

\texttt{range\_m(model, benchmark, bit) = max\_s(acc\_m,s) -\allowbreak{} min\_s(acc\_m,s)}.

The range/gap criterion holds in a cell exactly when

\texttt{max(range\_GPTQ, range\_AWQ) >= gap}.

\textbf{Supported:} H3 is supported if winner flips occur in at least 3 of the 8 confirmatory cells, or the range/gap criterion holds in at least 4 of the 8 confirmatory cells.

\textbf{Disconfirmed:} H3 is disconfirmed if winner flips occur in at most 1 of the 8 confirmatory cells and \texttt{max(range\_GPTQ, range\_AWQ) < 0.5 $\times$ gap} in at least 6 of the 8 confirmatory cells.

\textbf{Inconclusive:} Every outcome satisfying neither the support rule nor the disconfirmation rule is reported as inconclusive, without post-hoc promotion. Calibration-dataset effects, 3-bit results, ARC-Challenge, and HellaSwag are reported separately and cannot substitute for the eight-cell confirmatory rule.

H1 and H2 will be reported as estimated effects with confidence intervals rather than converted into new post-hoc binary thresholds. The preregistered pilot-motivated diagnostics are cancellation (\texttt{harmful} and \texttt{beneficial} both nonzero relative to net delta), required-n relative to benchmark size, and bootstrap method-rank flips.

\subsubsection{Exclusions and Missing Runs}
An item is excluded only for a benchmark loader/scorer failure that affects all compared methods; exclusions and reasons are logged before analysis. Failed checkpoint builds or jobs are rerun with the same registered seed and calibration indices. Backend changes create a new environment cell and do not silently replace prior results.

\subsubsection{Dated Amendments}
\paragraph{2026-07-20 --- WikiText-2 document definition (Decision Point A)}
\mbox{}\\
\textbf{Decision owner:} \authorname{}, alone.

\textbf{Results inspected before this decision:} No main-grid or mini-grid accuracy result has been inspected, and none exists. The only evidence seen is calibration-eligibility data, recorded below.

The 2026-07-13 calibration preflight failure recorded in \texttt{docs/\allowbreak{}WIKITEXT2\_PROTOCOL\_BLOCKER.md}: on \texttt{Salesforce/\allowbreak{}wikitext}, \texttt{wikitext-\allowbreak{}2-\allowbreak{}raw-\allowbreak{}v1}, train split, dataset revision \texttt{b08601e04326c79dfdd32d625aee71d232d685c3}, tokenized with \texttt{Qwen/\allowbreak{}Qwen2.5-\allowbreak{}1.5B-\allowbreak{}Instruct} revision \texttt{989aa7980e4cf806f80c7fef2b1adb7bc71aa306} under the pinned runtime (torch 2.13.0, transformers 5.13.0, datasets 5.0.0), seed 0, no added special tokens: \textbf{0 of 36,718 rows contain at least 2,048 tokens}, against a registered requirement of 128. No calibration artifact or checkpoint was produced. This is a property of the corpus layout, not of the seed --- WikiText-2 raw distributes each article across many short rows --- so no reseeding can make the registered row-level reading executable.

The 2026-07-20 eligibility probe, run \textbf{before} this decision was taken, on the same pinned dataset revision, tokenizer revision, and runtime, applying the reconstruction rule below verbatim (SLURM job \texttt{11303825}, embers CPU, in-image; algorithm \texttt{wikitext2-\allowbreak{}level1-\allowbreak{}heading-\allowbreak{}articles-\allowbreak{}v1}): \textbf{629 articles reconstructed from 36,718 rows, of which 425 contain at least 2,048 tokens} against the registered requirement of 128 --- verdict SUFFICIENT, a 3.3$\times$ pool. Token-length distribution across the 629 reconstructed articles: min 10, p25 1,703, median 2,857, p75 5,320, p90 8,577, max 21,295, mean 4,002.0; 2,517,232 tokens total. Eligibility is a property of the articles and not of the seed, so this single count satisfies all five registered seeds. The probe script is persisted at \texttt{scripts/\allowbreak{}wikitext2\_article\_probe.py} and is the \textbf{reference implementation} of the reconstruction rule: the builder's reconstruction must match its \texttt{reconstruct\_articles} and \texttt{is\_level1\_heading} functions verbatim.

\textbf{Rationale:} The preregistration requires the WikiText-2 condition to sample 128 \emph{documents} of 2,048 tokens. It did not operationalize ``document'' for a corpus whose Hugging Face rows are line fragments rather than documents. This amendment supplies that definition. It changes no other registered parameter.

\textbf{Amended rule.} For the Qwen2.5-1.5B-Instruct WikiText-2 calibration condition, a \emph{document} is one article, reconstructed deterministically from the raw corpus as follows:

\begin{enumerate}
\item Read the train split in source row order at the pinned dataset revision.
\item An article begins at each row that is a level-1 heading --- a row whose text, after stripping surrounding whitespace, matches a single \texttt{=}-delimited title (\texttt{= Title =}) and not a deeper heading (\texttt{= = Subtitle = =} or lower). All rows up to but excluding the next level-1 heading belong to that article. Rows preceding the first level-1 heading form no article and are discarded.
\item An article's text is its member rows concatenated in source order, joined exactly as stored, with no normalization beyond that already present in the raw corpus.
\item The reconstruction algorithm is versioned and persisted with the artifact, together with the ordered article index array and a hash of each reconstructed article, so the reconstruction is independently checkable.
\end{enumerate}
The registered sampling rule then applies unchanged to that article array: for seed \texttt{s}, shuffle the complete reconstructed-article index array using \texttt{numpy.random.default\_rng(s).shuffle}; visit articles in that order; tokenize each without adding special tokens; skip articles shorter than 2,048 tokens; retain the first 2,048 tokens from each eligible article until 128 samples are collected. Persist the selected article indices and token hashes. GPTQ seed \texttt{s} and AWQ seed \texttt{s} receive the identical ordered calibration samples, as for C4.

Seeds \texttt{\{0,1,2,3,4\}}, sample count 128, token length 2,048, the eligibility threshold, method pairing, and index retention are \textbf{unchanged} from the frozen protocol.

\textbf{Fail-closed condition.} The builder continues to fail closed on the WikiText-2 condition until this rule is implemented and tested. If the reconstruction yields fewer than 128 eligible articles for any seed, the builder must raise rather than relax any registered parameter, and the shortfall returns here as a new dated amendment.

\subsection{Mini-Grid Registration: Scope, Escalation Rule, and Reporting}
\label{app:reg:minigrid}  
\emph{Reproduced from \texttt{docs/\allowbreak{}MINIGRID\_REGISTRATION\_2026-\allowbreak{}07-\allowbreak{}15.md}; see \S\ref{app:reg:note}.}  

Status: \textbf{FROZEN 2026-07-15}, by the commit containing this line, before any mini-grid job exists and before any mini-grid accuracy result was inspected. Deviations require a dated entry under Dated Amendments stating whether results were inspected before the decision.

This document does not amend \texttt{PREREGISTRATION.md} (frozen 2026-07-11). It constrains the experimenter for a staged subset of the registered grid. The registered H3 protocol, metrics, and decision rule are unchanged.

\subsubsection{1. Scope}
The mini-grid executes exactly 4 of the 8 registered confirmatory H3 cells:

\texttt{\{Qwen2.5-\allowbreak{}1.5B-\allowbreak{}Instruct, Llama-\allowbreak{}3.2-\allowbreak{}3B-\allowbreak{}Instruct\} $\times$ \{MMLU, GSM8K\}}

at 4-bit, GPTQ and AWQ, calibration seeds \{0,1,2,3,4\}, paired C4 calibration sets per the registered sampling algorithm. Pinned model, dataset, and benchmark revisions are those in \texttt{configs/\allowbreak{}main\_grid\_manifest.yaml}. The 7B/8B cells (Qwen2.5-7B-Instruct, Llama-3.1-8B-Instruct $\times$ MMLU, GSM8K) are deferred and executed only if the escalation rule in \S{}3 fires.

\subsubsection{2. Benchmark execution parameters not fixed by the preregistration}
\begin{itemize}
\item GSM8K few-shot count for all mini-grid (and any later confirmatory) cells: \textbf{1 few-shot example, inline in the user message --- matching the validated bridge configuration \texttt{configs/\allowbreak{}pace\_bridge\_chat.yaml}}.
\item GSM8K items: test indices 0--999 in dataset order. MMLU: full test split.
\item Chat template ON for every method including FP16 baselines (registered).
\item Llama-3.2-3B FP16 operational acceptance ranges: to be derived from a trusted lm-evaluation-harness reference run on the pinned snapshot (procedure of \texttt{docs/\allowbreak{}MMLU\_REFERENCE\_RUN.md}) and committed into the mini-grid config \textbf{before any quantized Llama-3.2-3B result exists}. This registration deliberately freezes the derivation procedure, not the values; the values are recorded in the mini-grid config commit and are operational gates only.
\end{itemize}
\subsubsection{3. Escalation rule (mechanical, pre-committed)}
Compute, per the frozen algebra of \texttt{PREREGISTRATION.md}, for each of the 4 cells at 4-bit: winner flips across seeds (ties counted separately, per the registered tie rule) and the range/gap criterion \texttt{max(range\_GPTQ, range\_AWQ) >= gap}.

\textbf{Escalate} to the deferred 7B/8B seed cells iff:

\begin{itemize}
\item winner flips occur in \textbf{at least 1 of the 4 cells}, OR
\item the range/gap criterion holds in \textbf{at least 2 of the 4 cells}.
\end{itemize}
If neither condition holds, the 7B/8B cells are not built, and no other result (3-bit, other benchmarks, atlas findings) can substitute to trigger escalation.

\subsubsection{4. Relation to the frozen H3 decision rule}
The registered Supported/Disconfirmed/Inconclusive rule is defined over all 8 confirmatory cells. Therefore:

\begin{enumerate}
\item If escalation fires and all 8 cells complete, the frozen rule is applied exactly as registered.
\item If escalation does not fire (or the 7B/8B cells are otherwise not run), the paper reports the 4 completed cells \textbf{descriptively}, and H3 is reported as \textbf{formally inconclusive under the registered rule} --- never as supported or disconfirmed on 4 cells. No reduced-cell variant of the rule will be constructed after results are seen.
\item The registered secondary analyses (3-bit dose-response, ARC-Challenge, HellaSwag, WikiText-2 calibration-distribution contrast) remain deferred with the rest of the main grid and are not part of the mini-grid.
\end{enumerate}
\subsubsection{5. Analysis and inspection discipline}
\begin{itemize}
\item During fan-out, only job health, checksums, expected-file coverage, and receipt pairing are inspected (runbook grid discipline). First accuracy inspection happens only after the mini-grid validator passes over the complete 44-JSONL expected set.
\item The registered hierarchical analysis (\texttt{flipeval paired-\allowbreak{}seeds}, 2000 replicates, bootstrap seed 0) is run once per cell; the escalation rule in \S{}3 is then applied mechanically and a dated escalation decision record is written the same day.
\item The paired-bootstrap rank-flip denominator convention (tie replicates included in the denominator, ties also reported separately) is the one implemented and documented in \texttt{flipeval/\allowbreak{}core.py} as of commit \texttt{a8ba9f0}; it is hereby fixed in words and will not be re-chosen after inspection.
\end{itemize}
\subsubsection{Dated Amendments}
\paragraph{Amendment 2 (2026-07-21, \authorname{}): GSM8K few-shot binding}
\mbox{}\\
\S{}2 states GSM8K uses ``1 few-shot example, inline in the user message --- matching the validated bridge configuration.'' These two clauses conflict: the validated bridge configuration (\texttt{configs/\allowbreak{}pace\_bridge\_chat.yaml}, \texttt{fewshot: 1}) is implemented in \texttt{pilot\_eval/\allowbreak{}tasks.py} as a boolean switch that emits the fixed 3-example \texttt{GSM8K\_FEWSHOT} block, so the bridge canary that passed was validated with 3 inline examples. The operative binding for the mini-grid is the validated bridge configuration: 3 inline few-shot examples, byte-identical prompt path to the bridge. The literal count ``1'' in \S{}2 was a drafting error. The config key \texttt{fewshot} retains its existing boolean semantics (truthy $\Rightarrow$ the fixed 3-example block; the integer is not a count).

Results-blind status: no mini-grid accuracy results exist or have been inspected as of this amendment; the bridge canary results that were inspected are operational validation runs outside the mini-grid's confirmatory set.

\subsection{Public Per-Item Atlas Mining Registration}
\label{app:reg:atlas}  
\emph{Reproduced from \texttt{docs/\allowbreak{}ATLAS\_MINING\_REGISTRATION\_2026-\allowbreak{}07-\allowbreak{}15.md}; see \S\ref{app:reg:note}.}  

Status: \textbf{FROZEN 2026-07-15}, by the commit containing this line --- before any flip statistic from these sources was computed beyond the two feasibility probes disclosed in \S{}1. Deviations require a dated entry under Dated Amendments stating whether results were inspected before the decision.

Purpose: extend the Compression Flip Atlas with paired per-item records mined from public evaluation dumps, at zero GPU cost. These analyses are \textbf{descriptive/exploratory}: they estimate flip and churn magnitudes in the wild and feed the certification tables. They test no registered hypothesis and cannot substitute for any H3 cell. This protocol is registered to prevent source- or pair-selection after seeing results.

\subsubsection{1. Disclosure of pre-registration data contact}
Two feasibility probes were run on 2026-07-15 before this draft was written, and their results are known: (a) TheBloke/Llama-2-7B-GPTQ vs meta-llama/Llama-2-7b-hf on ARC-Challenge (74/1,170 flips, net $-$1.03 pp), and (b) neuralmagic Llama-3.1-8B baseline vs W4A16 on bbh\_boolean\_expressions (17/250 flips, net +1.2 pp). These two cells are flagged \texttt{probe=true} in the atlas and excluded from any aggregate statistic quoted in the paper's abstract or headline claims.

\subsubsection{2. Sources (fixed; no additions after freeze without a dated amendment)}
\begin{itemize}
\item \textbf{S1 --- Open LLM Leaderboard v1 archive} (\texttt{open-\allowbreak{}llm-\allowbreak{}leaderboard-\allowbreak{}old} details datasets; public, ungated; no declared license --- recorded as a limitation in the datasheet).
\item \textbf{S2 --- Neural Magic/Red Hat per-item dumps}: \texttt{neuralmagic/\allowbreak{} quantized-\allowbreak{}llama-\allowbreak{}3.1-\allowbreak{}leaderboard-\allowbreak{}v2-\allowbreak{}evals} (Apache-2.0) and, if their per-item schema validates, the companion arena-hard and humaneval datasets.
\end{itemize}
\subsubsection{3. Pair enumeration rule (run BEFORE any flip computation)}
\begin{enumerate}
\item Enumerate all S1 details datasets whose model name matches, case-insensitive, any of: \texttt{GPTQ}, \texttt{AWQ}, \texttt{GGUF}, \texttt{8bit}, \texttt{4bit}, \texttt{bnb}. For each, identify the base model from the quantizer's model card; include the pair iff a details dataset exists for that exact base model.
\item Within a pair, use the latest run timestamp per task for each side unless prompt-hash agreement (rule \S{}4.2) fails, in which case try earlier run combinations in reverse-chronological order and record the choice.
\item S2 pairs are the nine baseline$\times$\{W4A16, W8A8-INT8, W8A8-FP8\} combinations at 8B/70B/405B, all tasks present in the dump.
\item The frozen pair list (a machine-readable manifest with dataset URLs, run timestamps, and task lists) is committed as \texttt{docs/\allowbreak{}atlas\_pair\_manifest.json} \textbf{before} flip statistics are computed. Pairs discovered later require a dated amendment here.
\end{enumerate}
\subsubsection{4. Item pairing validity rules (mechanical)}
\begin{enumerate}
\item Join key: S1 \texttt{hashes.example}; S2 \texttt{doc\_id} with byte-identical \texttt{doc} (or \texttt{doc\_hash} where present). Duplicated join keys within a file are dropped entirely (both occurrences) and counted.
\item An item enters the paired analysis iff its full-prompt hash (S1 \texttt{hashes.full\_prompt}; S2 \texttt{prompt\_hash}) is identical across the pair. A pair-task cell is \textbf{excluded} iff fewer than \textbf{99\%} of joinable items pass this identity check; exclusions are reported with their rates.
\item Per cell, record both sides' harness identity (lighteval/lm-eval git SHA, model args, dtype) verbatim from the results JSON. Differing harness SHAs do not exclude a cell (prompt-hash identity is the operative control) but are recorded and disclosed per cell.
\end{enumerate}
\subsubsection{5. Metrics (identical to the controlled-atlas suite)}
Per cell: net accuracy delta, harmful/beneficial flip rates, accuracy-state churn, total answer churn where raw predictions exist, exact two-sided McNemar, TOST at the registered 2 pp margin, minimum detectable difference at 80\% power, and required-n. Primary correctness column: \texttt{acc\_norm} where present, else \texttt{acc} (S1); the task's primary metric as logged (S2); the choice is recorded per cell. No cell-level results drive inclusion/exclusion decisions.

\subsubsection{6. Reporting}
All enumerated, non-excluded cells are reported in the atlas regardless of outcome. Aggregates over cells are accompanied by the cell count and the exclusion table. The \texttt{probe=true} cells of \S{}1 appear in the atlas but not in headline aggregates.

\subsubsection{Dated Amendments}
None.

\subsection{Published-Claim Audit Registration}
\label{app:reg:audit}  
\emph{Reproduced from \texttt{docs/\allowbreak{}AUDIT\_REGISTRATION\_2026-\allowbreak{}07-\allowbreak{}15.md}; see \S\ref{app:reg:note}.}  

Status: \textbf{FROZEN 2026-07-15}, by the commit containing this line --- before any per-claim power computation was run. The claim list itself must also be frozen (\S{}3.4) before verdicts are computed. Deviations require a dated entry under Dated Amendments stating whether results were inspected before the decision.

Purpose: systematically assess whether published ``near-lossless'' compression claims are statistically supported at their reported evaluation sizes. Framing is constructive --- the field lacks reporting standards; flipeval and the certification tables are the proposed fix --- not an indictment of specific papers. Every verdict is mechanical and recomputable.

\subsubsection{1. Disclosure of pre-registration data contact}
Five candidate claims were collected with exact quotes on 2026-07-15 (GPTQ, LLM.int8(), SmoothQuant abstracts; the RedHatAI W4A16 Llama-3.1-8B model card; Meta's quantized Llama 3.2 blog) during a feasibility sweep. No power computation has been run on any of them. They enter the pool through the same \S{}3 criteria as later-collected claims.

\subsubsection{2. Population and sampling frame (fixed)}
Claim sources, enumerated exhaustively within each frame:

\begin{itemize}
\item \textbf{F1 --- Method papers:} the published versions of GPTQ, AWQ, SmoothQuant, LLM.int8(), SpinQuant, QuIP\#, SqueezeLLM, Wanda, and SparseGPT, plus any paper citing one of these that appears in the related-work sweep (\texttt{docs/\allowbreak{}related\_work\_checklist.md}) and makes an equivalence-type claim.
\item \textbf{F2 --- Official quantized model cards:} Meta, Qwen, Mistral, and Red Hat AI/Neural Magic quantized releases of models in the Llama-2/3.x and Qwen2.5 families on Hugging Face.
\item \textbf{F3 --- Inference-stack vendor posts:} vLLM, TensorRT-LLM, and llama.cpp official blog/docs pages making quantization-quality claims.
\end{itemize}
Target: \textbf{at least 10} claims; all claims meeting \S{}3 inclusion are audited (no discretionary sub-selection).

\subsubsection{3. Claim inclusion and extraction (mechanical)}
\begin{enumerate}
\item \textbf{Inclusion:} the source asserts, in prose or a table caption, that a compressed model's benchmark quality is equivalent-or-negligibly-different from its uncompressed baseline (trigger vocabulary: ``near-lossless'', ``negligible'', ``no (significant) degradation'', ``matches'', ``preserves accuracy'', ``X\% recovery'' with X $\geq$ 98, or an explicit $\leq$1 pp delta framed as parity). Perplexity-only claims are included only if a benchmark-accuracy claim also appears.
\item \textbf{Extraction fields (per claim):} exact quote ($\leq$15 words), source and version/date, benchmark(s), reported n per benchmark (as stated, else the benchmark's standard size, recorded as \texttt{imputed}), reported baseline accuracy, reported delta, whether per-item outputs are released, whether any statistical test or interval is reported.
\item \textbf{Double extraction:} each claim's fields are extracted twice on different days (solo-author substitute for dual coding) and discrepancies resolved before the verdict stage.
\item \textbf{Freeze:} the completed claim table is committed as \texttt{docs/\allowbreak{}audit\_claim\_table.csv} before any verdict is computed. Claims found after that commit go into a separately reported ``post-freeze'' stratum.
\end{enumerate}
\subsubsection{4. Verdict rules (mechanical, computed only after \S{}3.4 freeze)}
For each claim $\times$ benchmark, at the claim's reported n and baseline accuracy:

\begin{itemize}
\item \textbf{V1 --- Detection power:} minimum detectable accuracy delta at 80\% power, two-sided $\alpha$=0.05, under the paired-flip model with the discordance rate imputed from the atlas's empirical flip-rate distribution for the nearest (method family, bit width, benchmark) cell --- sensitivity-checked against the independent-binomial bound. Report MDD/claimed-margin ratio.
\item \textbf{V2 --- Equivalence support:} the n required for TOST at margin \textbf{2 pp} --- matching the registered main-grid TOST margin --- (and at the claim's own margin when it states one). A claim is labeled \textbf{``underpowered for its own assertion''} iff reported n < required n at the applicable margin.
\item \textbf{V3 --- Reproducibility:} binary --- could a third party run a paired test from released artifacts (per-item outputs public)?
\end{itemize}
Headline statistic: the fraction of audited claims labeled underpowered under V2, reported with its count and the full claim table. No claim is described as ``false'' or ``wrong''; the audited property is the evidential sufficiency of the reported evaluation, not the truth of the underlying equivalence.

\subsubsection{5. Robustness reporting}
Verdicts are recomputed under (a) the independent-binomial bound instead of the atlas-imputed discordance rate and (b) a margin sweep over 1 pp and 3 pp; a claim's verdict is called margin-sensitive if it changes across the sweep, and the count of margin-sensitive verdicts accompanies the headline number.

\subsubsection{Dated Amendments}
\textbf{2026-07-15 --- Amendment 1 (\S{}3.3 independence mechanism).} The requirement that the two extractions occur ``on different days'' is replaced by an extractor-independence requirement: the second extraction is performed in a fresh agent session with \textbf{no access to pass-1 outputs} --- the extractor is withheld \texttt{docs/\allowbreak{}audit\_claim\_table\_pass1.csv}, instructed not to read it or retrieve it from git history, and receives only the frozen protocol and the source frames. Rationale: temporal separation was a proxy for extractor independence calibrated to human memory; a fresh agent session carries no memory of pass 1, so blind same-day extraction provides at least the intended independence; the source-stability benefit of a second-day fetch is instead obtained by recording source content hashes in both passes where feasible. Decision context: made after pass-1 extraction results were known, but before any verdict, power, MDD, or required-n computation was run; the \S{}4 verdict rules and the \S{}3.1--3.2 inclusion/extraction rules are unchanged.

\textbf{2026-07-31 --- Amendment 2 (\S{}4 V2, the applicable margin).}

\emph{Defect.} \S{}4 V2 computes the required \$n\$ ``at margin 2 pp \dots{} (and at the claim's own margin when it states one)''. The phrase ``when it states one'' was never operationalised in this registration, and \S{}3.2 does not extract a margin: the frozen claim table \texttt{docs/\allowbreak{}audit\_claim\_table.csv} has no margin field. The implementation in \texttt{scripts/\allowbreak{}audit\_verdicts.py} supplied one after the freeze, by taking the largest delta the source reports and treating it as the margin the source states. Those are different quantities. A reported delta is an outcome of the evaluation; a margin is a threshold against which an outcome is judged. Every \texttt{margin\_basis} value in \texttt{results/\allowbreak{}audit\_verdicts\_rev2.csv} cites an observed quantity --- for example ``max |delta| over the 5 OPT-175B tasks'' (R01), ``the larger of the two stated deltas'' (R06), ``+0.15pp (68.69 vs 68.54)'' (R17). Verdicts labelled ``underpowered for its own assertion'' therefore rest, for those claims, on a margin the source did not assert.

\emph{Amendment.} The applicable margin is determined by the following rule, which replaces the parenthetical in \S{}4 V2. Each audited claim is assigned to exactly one category:

\begin{enumerate}
\item \textbf{Formal equivalence claim} --- the source states a numeric tolerance that is logically prior to the observed result: a threshold that could have been written down before the evaluation was run. Qualifying forms include ``within \$X\$'', ``no more than \$X\$'', ``at most \$X\$'' used as a requirement, ``a tolerance of \$X\$'', and equivalent constructions that bound what the source would accept.
\item \textbf{Informal near-lossless claim} --- the source reports a result and characterises it as negligible, but states no threshold prior to it. This includes deltas subsequently described as small, recovery percentages computed from the result, and phrases such as ``at most \$X\$'' used to describe the spread of observed differences rather than to set a bound.
\item \textbf{Unquantified claim} --- the source uses equivalence language without sufficient numerical information to evaluate either way. This category does not change; it continues to be handled by the \S{}4 indeterminacy rules.
\end{enumerate}
The determination is made against the frozen \texttt{exact\_quote} for each claim and, where the quote alone is inconclusive, against the source at the version and content hash recorded in the frozen claim table. It is recorded per claim, with the quoted text supporting it, in a new column of the verdicts CSV. \textbf{No claim is re-extracted and no source is re-fetched for extraction purposes}; the frozen claim table is unchanged. Sources were re-fetched for verification only, under the eligibility-and-provenance scope recorded below.

Verdicts are then computed as follows:

\begin{itemize}
\item The \textbf{primary verdict for every claim} is at the registered 2 pp margin, which \S{}4 already names first. The headline count is the number of determinate claims underpowered at 2 pp.
\item A claim in category 1 is \textbf{additionally} evaluated at its declared margin, reported alongside the primary verdict, and never in place of it.
\item Claims in category 2 are reported against the registered 1 pp / 2 pp / 3 pp sweep of \S{}5. \textbf{No margin derived from a claim's own reported results is described as that claim's stated, declared, asserted or own margin}, in the paper or in any released artifact.
\end{itemize}
\emph{Consequential quantities.} Every reported quantity that divides by the applicable margin is recomputed with it and re-reported: the V1 MDD-to-claimed-margin ratios, the required-\$n\$-to-reported-\$n\$ shortfall ratios, and the \S{}5 margin-sensitivity flag. Values previously reported against a result-derived margin are withdrawn, not silently updated; both the superseded and the corrected values remain in the released artifacts.

\emph{Analysis discipline.} The recomputation is run \textbf{once}, over the frozen claim table, and whatever it returns is reported --- including if the headline count falls, rises, or reaches zero. No variant of this rule is constructed after the recomputed values are seen.

\emph{Eligibility correction (R10).} A full-text review of every source, run 2026-07-31 and recorded in \texttt{docs/\allowbreak{}AUDIT\_SOURCE\_VERIFICATION\_2026-\allowbreak{}07-\allowbreak{}31.md}, established that R10's recorded \texttt{exact\_quote} --- ``average recovery percentage across all benchmarks is 98.6\%'' --- appears nowhere in its source. The card contains no prose equivalence claim; \texttt{98.6\%} is a table cell, and the extraction composed a sentence from tabular data and recorded it as a quotation. \S{}3.1 requires the assertion to appear ``in prose or a table caption'', and it appears in neither. \textbf{R10 is therefore excluded from the eligible population by applying the inclusion rule already registered in \S{}3.1, not by any new criterion.} The eligible population becomes 16. The frozen claim table is not edited; the exclusion is recorded in the verdicts CSV with the verification finding beside it, and the original row remains in the immutable frozen file and in the published v1.0.0 artifact.

This correction moves the eligible denominator. It changes neither the number of claims below the planning threshold nor the V3 per-item-outputs result: R10 was adequately powered at the registered 2 pp margin and recorded \texttt{no} on V3, so its removal cannot reduce the count of flagged claims. \textbf{The correction did not improve any count in the direction favourable to the audit's thesis.}

\emph{Scope.} This amendment changes the applicable margin, and reopens \S{}\S{}3.1--3.2 \textbf{only} to correct eligibility and provenance --- that is, to apply the existing inclusion rule to R10 and to record source provenance. No inclusion criterion is added, widened or narrowed, no claim is re-extracted, and no source is re-fetched for extraction purposes. Unchanged and not reopened: the frozen claim table itself; the \S{}4 V1 detection-power formula; the \S{}4 V3 reproducibility verdict; the indeterminacy rules and the claims currently indeterminate; the discordance imputation and its tier matching; the atlas; and every registration other than this one.

\emph{Reporting the surviving power result.} The claim below the planning threshold at the registered 2 pp margin is reported as a \textbf{sensitivity-dependent planning flag, not a stable binary verdict}, and never without its reversal point. The required \$n\$ is a planning quantity computed under an assumed true difference of zero and a point imputation of discordance; for the single flagged claim the classification reverses at approximately \$d = 0.1189\$ against an imputed \$d = 0.13\$, and 43.6\% of the 792 atlas cells supplying that imputation fall below the reversal point. Any report of this flag states the imputed value, the reversal point, and that fraction.

\emph{Decision context.} \textbf{Results were inspected before this decision, and so was the full-text classification.} The audit verdicts were computed on 2026-07-20, revised to rev-2 on 2026-07-21, reported in the paper, and released in the v1.0.0 artifact; the headline \$K = 4\$ of 12 has been public since 2026-07-30. The full-text verification of all 17 sources was run on 2026-07-31, \textbf{before this amendment was signed and at the decision owner's direction}, and its classification of every claim --- including the finding that no source declares a margin --- was known at signature. The rule in this amendment was not constructed against that classification: it was written and committed on 2026-07-31 at \texttt{19d485c}, before the verification ran, and that committed draft names R14 as its hard case and resolves it by the same priority test the verification later applied. The rule is unchanged from that commit. The eligibility correction was made by applying \S{}3.1 as already registered, and is verdict-neutral in the sense recorded above. The original immutable version of every superseded value remains accessible in the frozen claim table and the published v1.0.0 artifact.

\emph{Signed.} \authorname{}, 2026-07-31. Drafted by Claude Code at \texttt{19d485c}, revised at \texttt{bb45528} after the full-text source verification, and appended on the verbatim instruction ``sign and append''. The rule in the \emph{Amendment} clause above is unchanged from \texttt{19d485c}.

\textbf{2026-08-03 --- Amendment 3 (\S{}3.1 inclusion, applied to R09 and R17).}

\emph{Occasion.} Amendment 2 excluded R10 from the eligible population by applying \S{}3.1 as registered: the recorded quotation appeared nowhere in the source, and the assertion appeared in neither prose nor a table caption. That correction rested on a full-text review which, as recorded in \texttt{docs/\allowbreak{}AUDIT\_SOURCE\_VERIFICATION\_2026-\allowbreak{}07-\allowbreak{}31.md}, verified quotation \emph{accuracy} for all seventeen sources but quotation \emph{location} for R10 alone. The remaining sixteen therefore carried an unexamined \S{}3.1 basis. An author re-verification of the four claims named in that document's open items, recorded in \texttt{docs/\allowbreak{}AUDIT\_SELF\_RECHECK\_2026-\allowbreak{}08-\allowbreak{}02.md}, examined location directly against the archived sources.

\emph{Finding.} Three of the six quantized-model cards in the population, R09, R10 and R17, contain no \S{}3.1 trigger vocabulary anywhere in their prose. In each, the recovery percentage that would satisfy the trigger list exists only as a table cell, beneath a column header, and none of the three files contains a table caption element of any kind. R08, R15 and R16 are unaffected: each states a recovery percentage in prose at or above the registered threshold of 98. With respect to those recovery figures R09 and R17 occupy the same structural position that excluded R10, which is what makes their eligibility a live question. They differ from R10 in another respect, addressed below.

\emph{Determination.} \textbf{R09 and R17 remain in the eligible population.} The ground for retaining them is that \S{}3.1 does not decide the case, not that it decides the case in their favour.

\S{}3.1 admits a claim on ``an explicit $\leq$1 pp delta framed as parity''. That trigger has two limbs, and R09 and R17 satisfy the first without satisfying the second. Each card contains exactly one comparative sentence in prose, and it states both the compressed and the uncompressed score: 73.44 against 73.79 for R09 and 68.69 against 68.54 for R17, differences of 0.35 pp and 0.15 pp. A delta below one percentage point is therefore explicit on the face of the prose. Neither sentence characterises that difference. ``Whereas'' is a neutral contrastive, no trigger term appears in the prose of either card, and the recovery figures that would qualify outright, 99.52\% and 99.8\%, exist only as table cells beneath a \texttt{<th>Recovery} header in files carrying no table caption element at all. The registration does not say what follows when a source supplies the quantity and withholds the characterisation. \textbf{That is a gap in a rule \S{}3 declares to be mechanical, and it was not anticipated when the protocol was frozen on 2026-07-15.}

\textbf{R10 is excluded on a ground that does not depend on how that gap is resolved.} Its card contains no comparative sentence of any kind: the strings ``whereas'', ``unquantized'', ``baseline'' and ``compared'' do not occur anywhere in it, its evaluation section is a single table with no accompanying prose, and the sentence recorded for it in the frozen claim table was composed from tabular data and appears nowhere in the source. R10 therefore fails the first limb as well as the second, and its exclusion under Amendment 2 stands under the strict and the permissive reading alike. The difference in outcome between R10 and these two rests on a difference between the documents, not on this determination. Verified against the sealed source archive \texttt{a912a1e7\dots{}40259} on 2026-08-03, by a route independent of the vocabulary sweep recorded in \texttt{docs/\allowbreak{}AUDIT\_SELF\_RECHECK\_2026-\allowbreak{}08-\allowbreak{}02.md} \S{}4.1.

Two resolutions of the gap were therefore available, and both are defensible. The strict reading requires both limbs and excludes R09 and R17. The permissive reading treats an explicit sub-point delta, stated in prose in a document whose function is to offer the compressed model in place of the uncompressed one, as an assertion of negligible difference in substance, and retains them. \textbf{The tie is broken against the interest of this audit.} The strict reading shrinks the eligible and assessable populations and raises the proportion of assessable claims falling below the planning threshold; the permissive reading leaves every published count exactly where it stands. Where a frozen rule is genuinely silent and the author is choosing after having seen results, the only choice a reader can credit without also having to credit the author is the one that cannot improve the author's own finding. \textbf{This is an interpretation of a registered rule that is silent, not an extension of a rule that speaks}, and it is recorded here so that a reader sees the reasoning rather than inferring it from a denominator.

\emph{Quantities unchanged.} The eligible population remains 16 and the numerically assessable population remains 11. No verdict, no threshold classification, no per-item-outputs result and no imputation changes. The frozen claim table is not edited.

\emph{The alternative, and its direction.} Excluding R09 and R17 by the strict reading was available and was declined. It would have moved the eligible population from 16 to 14 and the assessable population from 11 to 9, left the count below the planning threshold at 1, and moved that count as a proportion of the assessable population from 9.1\% to 11.1\%. \textbf{That is the direction favourable to this audit's thesis.} Amendment 2 recorded that the R10 correction did not improve any count in that direction; this determination likewise does not, and it is the conservative of the two readings available. The quantities in this paragraph are recorded so that the choice is auditable rather than merely asserted.

\emph{Reporting.} The locus finding is reported in the paper as a result in its own right rather than as an eligibility adjustment: across the six cards, with the underlying evidence held constant, three assert recovery in prose, two state two scores and characterise neither, and one makes no comparative statement at all. The consequence reported alongside it is that an inclusion rule keyed to prose, which \S{}3.1 is, captures equivalence claims non-randomly, so the frozen candidate count of 17 is a floor on the population rather than a census of it. Any report of the eligible population states that the boundary cases were retained under this amendment.

\emph{Verification status.} The locus review supporting this amendment is author re-verification against archived sources, by a second automated pass of the same class of tool that produced the record it checked. It is \textbf{not} independent verification, and neither it nor any agreement between it and the 2026-07-15 passes may be reported as dual coding or inter-rater reliability. \S{}3.3 and Amendment 1 are unchanged.

\emph{Scope.} This amendment applies \S{}3.1 to two claims and records the reasoning. No inclusion criterion is added, widened or narrowed; no claim is re-extracted; no source is re-fetched for extraction purposes; the frozen claim table, the \S{}4 verdict rules, the indeterminacy rules, the discordance imputation, the atlas, and every other registration are unchanged and not reopened. Amendment 2 remains in force in full.

\emph{Decision context.} \textbf{Results were inspected before this decision.} The rev-3 verdicts were computed on 2026-07-31 and the locus classification on 2026-08-02, and both were known at signature. The determination reached is the one that leaves every published count where it stood and declines the change that would have improved the headline proportion. The superseded and the current readings of \S{}3.1 as applied to R09 and R17 are both recorded above.

\emph{Signed.} \authorname{}, 2026-08-03. Drafted by Claude Code at \texttt{2857cd0}; its \emph{Determination} was redrafted at \texttt{adaf263}, after the distinction it rests on was verified directly against the sealed source archive, to concede that \S{}3.1 is silent at this boundary rather than to assert that it is satisfied. Appended on the verbatim instruction ``sign it and append''.

\textbf{2026-08-04 --- Amendment 4 (provenance remap; no protocol change).}

\emph{Occasion.} \texttt{docs/\allowbreak{}audit\_sources\_20260731.tar.gz}, holding the full-text captures of all seventeen audited sources, entered the repository at \texttt{cc357db} and was never deleted, so every commit from there to HEAD carries it. \texttt{origin} is a public GitHub repository, so pushing any commit in that range publishes the corpus. That contradicts the redistribution review of 2026-08-02, which found four of the seventeen carry no grant permitting a third party to republish their text, and the seven method papers sitting under arXiv's default licence, which authorises arXiv to distribute them rather than authorising us to. It also contradicts \texttt{README.md}, which states the captures are not redistributed. The repository has therefore been unpushable since 2026-08-02, and the resolution of record was to leave it so.

\emph{Finding.} Removing the blob rewrites 54 commits. \texttt{bb45528} is one of them, and Amendment 2 cites it in its own signature line above. The rewrite was computed on a throwaway clone and verified there before this amendment was drafted: 251 commits before and after with nothing pruned, author, email, date and subject byte-identical for every commit, and the only tree difference across the entire range the removed tarball. \texttt{19d485c}, also cited in that signature line, is an ancestor of \texttt{cc357db} and does not change. \texttt{987377a}, tag \texttt{v1.0.0}, is likewise an ancestor, so the release tag and its Zenodo archive lie outside the rewrite. The private sealed copy hashes \texttt{a912a1e7af0efd58459dcf57ade84be96cfea8337147a13d336dacfdb9240259}, identical to the blob in git, so removal loses nothing.

\emph{Determination.} \textbf{The tarball is removed from the repository's history, and \texttt{bb45528} is superseded by \texttt{ed92ae8} on the record rather than in the text.} The corpus remains identified and digest-checkable: the \texttt{.sha256} and \texttt{docs/\allowbreak{}audit\_sources\_manifest.tsv} are retained, and \texttt{scripts/\allowbreak{}fetch\_audit\_sources.py} rebuilds it from each publisher. The signature line of Amendment 2 is to be read as citing \texttt{ed92ae8} wherever it cites \texttt{bb45528}, and the full mapping at \texttt{docs/\allowbreak{}audit\_source\_tarball\_hash\_map\_20260804.tsv} governs any other stale identifier. \textbf{The original text of Amendment 2 is not edited.} The superseded identifier stays exactly as signed, because correcting it in place would leave a record reading as though the chain never broke, and that it broke is what this amendment exists to preserve.

\emph{The alternative, and its direction.} The alternative was abandoning the rewrite and never pushing, which was the resolution of record from 2026-08-02. It is rejected because it makes the repository permanently unpublishable, and the artifact link in a submitted paper cannot point at a repository that does not exist. The cost of the rewrite is borne once and is documented in the mapping file; the cost of not pushing recurs indefinitely. What the rewrite concedes is that it is irreversible once pushed, and that 54 commit hashes cited anywhere outside the mapping file go stale without warning.

\emph{Quantities unchanged.} No inclusion rule, eligibility rule, verdict rule, indeterminacy rule, discordance imputation, denominator or count is reopened. The eligible population remains 16. Amendments 1, 2 and 3 remain in force in full, and every published number stands.

\emph{Scope.} This amendment records a change to commit identifiers and to what the repository distributes. It changes no analysis and no audited property.

\emph{Decision context.} \textbf{Results were inspected before this decision.} The rev-3 verdicts were computed on 2026-07-31, the locus classification on 2026-08-02, and Amendment 3 was signed on 2026-08-03; all were known at signature. This amendment changes no analysis, no count and no verdict, so there is no outcome for that knowledge to have biased. It is recorded because the standing requirement applies to every amendment to a frozen protocol, not only to those that could move a number.

\emph{Signed.} \authorname{}, 2026-08-04. Drafted by Claude Code at \texttt{df1615b} after the rewrite was computed and verified on a throwaway clone; its \emph{Determination} and \emph{Decision context} were completed on the verbatim instruction ``i read the draft, fill in the two sections as approved and append''. The rewrite itself was not applied at the time of signature.

\textbf{2026-08-06 --- Amendment 5 (provenance remap, second order; no protocol change).}

\emph{Occasion.} Amendment 4's \emph{Determination} provides that ``the full 54-row mapping at \texttt{docs/\allowbreak{}audit\_source\_tarball\_hash\_map\_20260804.tsv} governs any other stale identifier''. That mapping was computed at \texttt{83d2b6c}. Five commits postdate it, each carrying the tarball in its tree and therefore each rewritten by the removal Amendment 4 authorises, so each falls outside the governance Amendment 4 established. Three of the five are Amendment 4's own commits. The commit that signed and appended it, \texttt{02ea10b}, becomes \texttt{5339a48}: an amendment written to remap a signature's provenance pointer moved its own.

\emph{Finding.} The rewrite was applied on 2026-08-06 to a disposable clone containing only \texttt{main}, using \texttt{git filter-\allowbreak{}branch} with an index filter, the same tool the signed map was computed with. A different tool was considered and rejected: commit hashes are a function of the tool's exact output, and a substitution that shifted any of the fifty-four signed pairs would have falsified a signed document rather than implemented it. All fifty-four rows reproduced exactly, \texttt{bb45528} to \texttt{ed92ae8} included. Two hundred and fifty-six commits before and after, none pruned, with author, email, author date, committer, committer date and subject byte-identical for every commit; the tarball the only path differing in any tree; the blob absent from the object database and unreachable from every ref; and \texttt{987377a}, tag \texttt{v1.0.0}, unchanged and still an ancestor of the sanitized tip, so publication is a fast-forward and the Zenodo archive of v1.0.0 lies outside the rewrite exactly as Amendment 4 stated. Two findings were not anticipated. Eleven references reached the blob rather than only \texttt{main}: nine local branches and the tag \texttt{pre-\allowbreak{}source-\allowbreak{}tarball-\allowbreak{}removal-\allowbreak{}20260802}. And the source-state freeze failed its own gate afterwards, because the freeze manifest recorded \texttt{source\_commit} \texttt{8aafe22}, which the rewrite replaces with \texttt{4a8a83f} --- the identical failure the 2026-07-29 identity rewrite produced, and the reason tag \texttt{v1.0.0} had to be moved to \texttt{987377a} before Zenodo minted. It was refreshed; all eighty-six recorded file hashes are unchanged and only provenance pointers move.

\emph{Determination.} \textbf{The five commits Amendment 4's map could not reach are governed by \texttt{docs/\allowbreak{}audit\_source\_tarball\_hash\_map\_20260806\_addendum.tsv}, and Amendment 4's own signature commit is superseded on the record rather than in its text.} The addendum records the five old-to-new pairs and is read together with the 54-row map as the complete mapping for \texttt{cc357db..HEAD}; the signed 54-row file is not edited, and its ``54-row'' description remains literally true of it. The signature line of Amendment 4 is to be read as citing \texttt{5339a48} wherever it cites \texttt{02ea10b}, and its drafting commit \texttt{df1615b} as \texttt{7d053c1}. The text of Amendment 4 is not edited, for the reason Amendment 4 itself gave about Amendment 2: a provenance chain that repairs itself silently is worth less than one that carries its own repair. The tag \texttt{pre-\allowbreak{}source-\allowbreak{}tarball-\allowbreak{}removal-\allowbreak{}20260802} is deleted from the publication clone rather than remapped, because it marks the state that still contained the tarball and a remapped version of it would point at a tree without the tarball and so assert something false; the state it marked is preserved outside the repository in the sealed pre-rewrite backup of 2026-08-06, which is not published.

\emph{What this concedes.} The remapping recursed once, and this amendment is the record of that recursion rather than a guarantee against it. It does not recur again. Amendment 4 was drafted and committed while the tarball was still in every tree, so its own commits were inside the range its map governed. This amendment is the first appended after the rewrite was applied, so its commits never contain the tarball and are never rewritten. The condition that produced the defect is gone, not merely handled.

\emph{Quantities unchanged.} No inclusion rule, eligibility rule, verdict rule, indeterminacy rule, discordance imputation, denominator or count is reopened. The eligible population remains 16. Amendments 1, 2, 3 and 4 remain in force in full, and every published number stands.

\emph{Scope.} This amendment records commit identifiers and one deleted tag. It changes no analysis and no audited property.

\emph{Decision context.} \textbf{Results were inspected before this decision.} The rev-3 verdicts were computed on 2026-07-31, the locus classification on 2026-08-02, Amendment 3 was signed on 2026-08-03 and Amendment 4 on 2026-08-04; all were known here. This amendment changes no analysis, no count and no verdict, so there is no outcome for that knowledge to have biased. It is recorded because the standing requirement applies to every amendment to a frozen protocol, not only to those that could move a number.

\emph{Signed.} \authorname{}, 2026-08-06. Drafted by Claude Code at \texttt{0e9140b}, after the rewrite had been applied and verified on a disposable clone, on the verbatim instruction ``sign the addendum''. The draft was returned unsigned because the agent had been asked to sign a document that did not yet exist; it was read, and its one open statement --- that results were inspected before the decision --- was confirmed on the verbatim instruction ``confirm''.
}

\end{document}